%% file: 00_paper.tex
\documentclass{article}
\PassOptionsToPackage{numbers}{natbib}
\usepackage{microtype}
\usepackage{graphicx}
\usepackage{booktabs} 

\usepackage{hyperref}

\usepackage[accepted]{icml2022}

\usepackage{amsmath}
\usepackage{amssymb}
\usepackage{mathtools}
\usepackage{amsthm}

\usepackage[capitalize,noabbrev]{cleveref}

\usepackage{rotating}
\usepackage{subcaption} 
\usepackage{adjustbox}
\usepackage{float}
\usepackage{multirow}
\usepackage[compact]{titlesec}
\usepackage[inline]{enumitem}

\usepackage{xcolor}         

\theoremstyle{plain}

\theoremstyle{definition}

\theoremstyle{remark}

\usepackage[textsize=tiny]{todonotes}

\titlespacing{\section}{0pt}{*0.3}{*0.3}
 \titlespacing{\subsection}{0pt}{*0.3}{*0.3}
\icmltitlerunning{PSP-HDRI$+$: A Synthetic Dataset Generator for Pre-Training of Human-Centric Computer Vision Models}

\begin{document}

\twocolumn[
\icmltitle{PSP-HDRI+: A Synthetic Dataset Generator for Pre-Training of \\
Human-Centric Computer Vision Models}




\begin{icmlauthorlist}
\icmlauthor{Salehe Erfanian Ebadi}{uwu}
\icmlauthor{Saurav Dhakad}{uwu}
\icmlauthor{Sanjay Kumar Vishwakarma}{uwu}
\icmlauthor{Chunpu Wang}{uwu}
\icmlauthor{You-Cyuan Jhang}{uwu}
\icmlauthor{Maciek Chociej}{uwu}
\icmlauthor{Adam Crespi}{uwu}
\icmlauthor{Alex Thaman}{uwu}
\icmlauthor{Sujoy Ganguly}{uwu}
\end{icmlauthorlist}

\icmlaffiliation{uwu}{Unity Technologies}

\icmlcorrespondingauthor{Applied Machine Learning Research}{applied-ml-research@unity3d.com}

\icmlkeywords{Machine Learning, Synthetic Data, Computer Vision, Human-Centric Computer Vision, Pre-training}

\vskip 0.3in
]



\printAffiliationsAndNotice{}  

\begin{abstract}
\input{01_Abstract}
\end{abstract}
\input{02_Introduction}
\input{03_RelatedWork}
\input{04_PSP_HDRI}
\input{05_Experiments}
\input{07_Conclusions}

\bibliography{bib}
\bibliographystyle{icml2022}
\newpage
\renewcommand\thefigure{\thesection.\arabic{figure}}  
\renewcommand\thetable{\thesection.\arabic{table}}  
\appendix
\onecolumn
\input{08_Appendix}

\end{document}

%% file: 01_Abstract.tex
%
We introduce a new synthetic data generator PSP-HDRI$+$ that proves to be a superior pre-training alternative to ImageNet and other large-scale synthetic data counterparts. We demonstrate that pre-training with our synthetic data will yield a more general model that performs better than alternatives even when tested on out-of-distribution (OOD) sets. Furthermore, using ablation studies guided by person keypoint estimation metrics with an off-the-shelf model architecture, we show how to manipulate our synthetic data generator to further improve model performance.

%% file: 02_Introduction.tex
\section{Introduction}
Supervised pre-training has accelerated success in computer vision applications. 
There remain questions about which type of data is best suited for pre-training models that are specialized to solve one task. For human-centric computer vision, researchers have established large-scale human-labeled datasets~\citep{lin2014microsoftcoco, andriluka14cvpr, li2019crowdpose, milan2016mot16, Johnson10lspet, zhang2019pose2seg}. 
These datasets are hard to create and label and are increasingly scrutinized for labeling and data bias, ethics, legality, and safety issues. 
Recently, researchers have started considering synthetic data alternatives to mitigate those issues~\citep{wood2021fake, fabbri2018jta, hu2019sailvos, hu2021sailvos3d, fabbri2021motsynth, bak2018people, pishchulin2011people, varol2017humans, kviatkovsky2020people, hassan2021populating, ros2016synthia, gaidon2016virtual, dosovitskiy2017carla, richter2017benchmarks, wrenninge2018synscapes, roberts2020hypersim, li2021openrooms, morrical2021nvisii}.

However, synthetic data generators are challenging to create; so researchers have focused on leveraging game environments such as GTA V to render synthetic data and labels. 
Most of these datasets have pre-rendered frames, and researchers cannot manipulate the data generator. 
Hence, the barrier of entry into simulation-ready and user-friendly synthetic data generators is still high for computer vision and AI researchers. 
Consequently, exploiting synthetic data to its full potential is yet to be realized.
Motivated by these limitations, we present a privacy-preserving, ethically sourced, and fully manipulable synthetic data generator for human-centric computer vision named PSP-HDRI$+$, which is built upon PeopleSansPeople~\citep{ebadi2021peoplesanspeople} (PSP) in Unity~\footnote{The template Unity environment, benchmark binaries, and source code is available at: \url{https://github.com/Unity-Technologies/PeopleSansPeople}}. 

We demonstrate a strategy by which synthetic data can help surpass benchmark model performance. 
We explore two main tasks, namely human detection and keypoint localization. 
Having experimented with multiple training methodologies, we have consistently found pre-training with synthetic data and fine-tuning on real data through transfer learning to be the most successful strategy across many domains and tasks. 
We validate the above propositions with model performance on in-distribution and OOD benchmarks in our extensive studies.

This work argues that pre-training with synthetic human-centric data helps generalize model performance in the real world. 
These performance gains hold even when our synthetic dataset is not tuned to create a 1:1 replica (digital twin) of the real world. 
We show that even using na\"ively generated human-centric data makes it possible to pre-train models that perform better than ImageNet and other large-scale data. 
The performance gains from synthetic pre-training are substantial in the few-shot learning and small data regiments.

%% file: 03_RelatedWork.tex
\section{Related Work}
Models pre-trained using large-scale human-labeled datasets such as ImageNet~\citep{deng2009imagenet}, MS COCO~\citep{lin2014microsoftcoco}, PASCAL VOC~\citep{everingham2010pascalvoc}, NYU-Depth V2~\citep{silberman2012nyudepthv2}, and SUN RGB-D~\citep{song2015sunrgbd} have enabled rapid progress across many computer vision tasks. The pre-training usually expedites the training process by leveraging the already-learned representations. The task-specific nature of the pre-training stage tends to encode representations that are necessary to solve that specific task. If the fine-tuning stage involves solving a different task, the pre-learned features may harm or reduce the model's ability to perform at its full potential and capacity on the downstream tasks~\citep{goyal2022vision}.

Furthermore, research has shown that metric sensitivity is a problem with universal generalist pre-training datasets such as ImageNet~\citep{kynkaanniemi2022role}, which can trickle down into the kinds of learned representations. Recently, in~\citep{madan2020and} the authors argued that data diversity improves OOD performance but degrades in-distribution performance. Research has shown that pre-training without natural images (with fractals) can help in many natural image tasks~\citep{kataoka2020pre}.
Therefore, it is necessary to ask, how much of the universal generalist pre-training data is paramount to better fine-tuning? Do the labeling inaccuracies in such datasets have a detrimental effect on transfer learning? More importantly, can superior alternatives be found that are easy to create and label?
Therefore, we are motivated to understand whether na\"ively generated task-specific synthetic data of any size can replace universal generalist pre-training. 

%% file: 04_PSP_HDRI.tex
\section{PSP-HDRI}
\subsection{Baseline Environment Design}
We use PeopleSansPeople~\citep{ebadi2021peoplesanspeople} (PSP) as our baseline data generator. PSP is a parametric data generator, created in the Unity game engine, and contains simulation-ready and fully rigged 3D human assets, a diverse animation library for humans, a parameterized lighting and camera system, and generates synthetic RGB images with ground truth annotations of 2D/3D bounding box, human keypoints, and semantic/instance segmentation. PSP leverages \textit{Domain Randomization}~\citep{tobin2017domain} where aspects of the simulation environment are randomized to introduce variations in the generated synthetic data. These variations are necessary to increase the generalization of models trained with the synthetic data to the real or other domains.
The Unity Perception package provides a domain randomization framework~\citep{borkman2021unity}, which allows for diversifying the synthetic dataset with a large number of parametric variations, using a ``randomizer'' paradigm. During the simulation, the randomizers act on predefined Unity scene components (e.g., lighting, camera, environment, human assets' placement and orientation, human assets' clothing and pose, etc.). The randomizers can use various sampling techniques to randomly select parameters for each parametric attribute of the scene components. More information can be found in~\citep{ebadi2021peoplesanspeople, borkman2021unity}.

\subsection{Bridging the Visual Quality and Label Gaps}

The existing domain gap between the synthetic and other types of data -- whether real or synthetic -- poses challenges for OOD performance of models trained solely on synthetic data. Some factors that will exacerbate the domain gap are content, material, and texture quality. These however, are very hard and expensive to source and usually require hours of extensive work by artists. 
PSP uses low-resolution COCO images as background textures. 
Whilst the background regions in PSP seem to affect 
the visual quality of the generated images
more than the foreground regions, other factors such as simulated sensor noise and sensor type, asset geometry, asset quality, material textures, etc. also do contribute to the overall perceived quality.
To increase the visual quality,
we complement PSP with High Dynamic Range Image (HDRI) backgrounds from Poly Haven\footnote{\url{https://polyhaven.com/hdris}} as skyboxes, which will enable high-quality rendered backgrounds for our dataset. These skyboxes contribute to ambient scene lighting and global illumination. We also use a combination of scene lighting intensity and Sun randomization (which simulates the time of the day and day of the year) to capture more realistic and diverse lighting settings. 

\begin{figure*}[htb!] 
    \centering
    \begin{subfigure}[t]{0.16\textwidth}
        {\includegraphics[height=2.75cm]{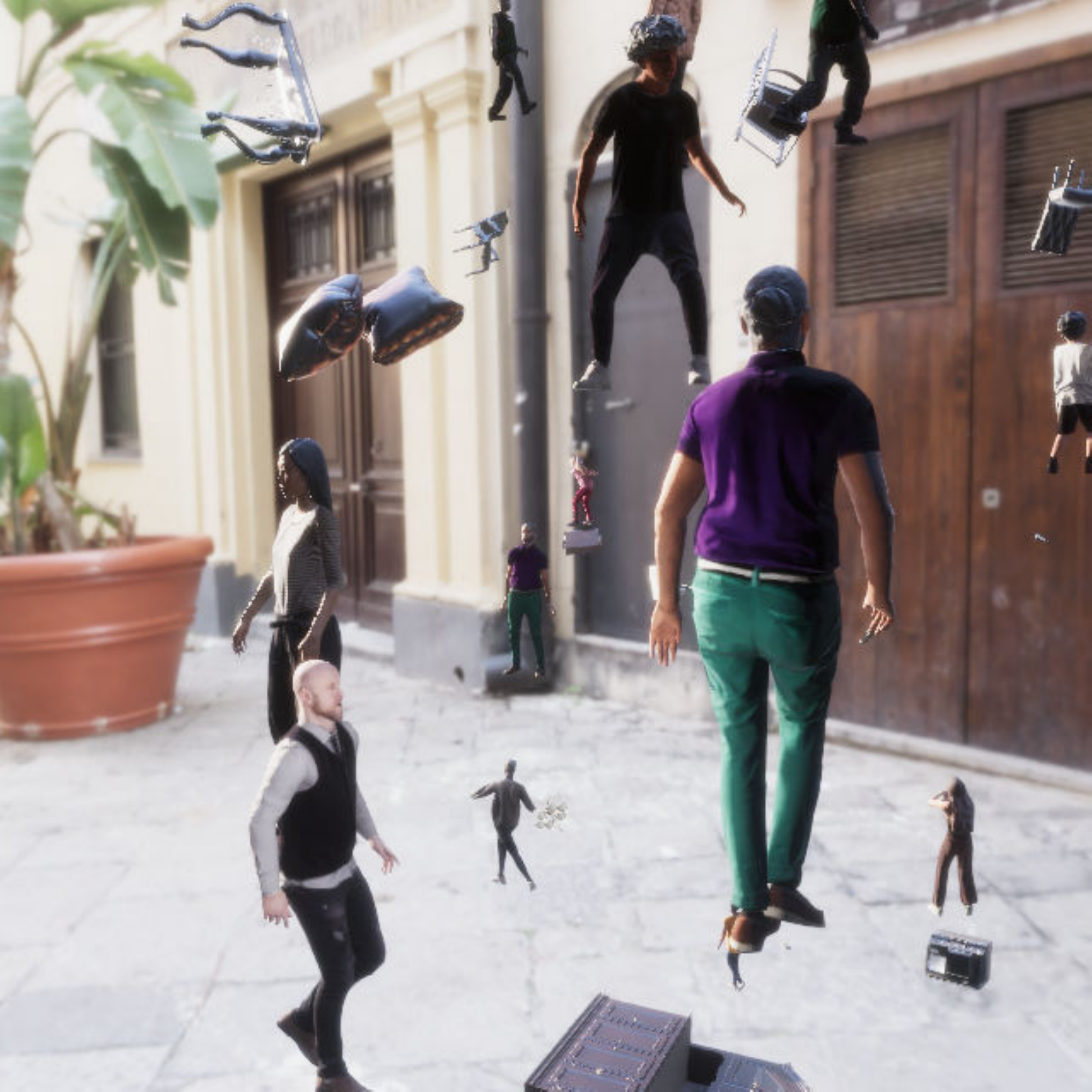}}
    \end{subfigure}
    \begin{subfigure}[t]{0.16\textwidth}
        {\includegraphics[height=2.75cm]{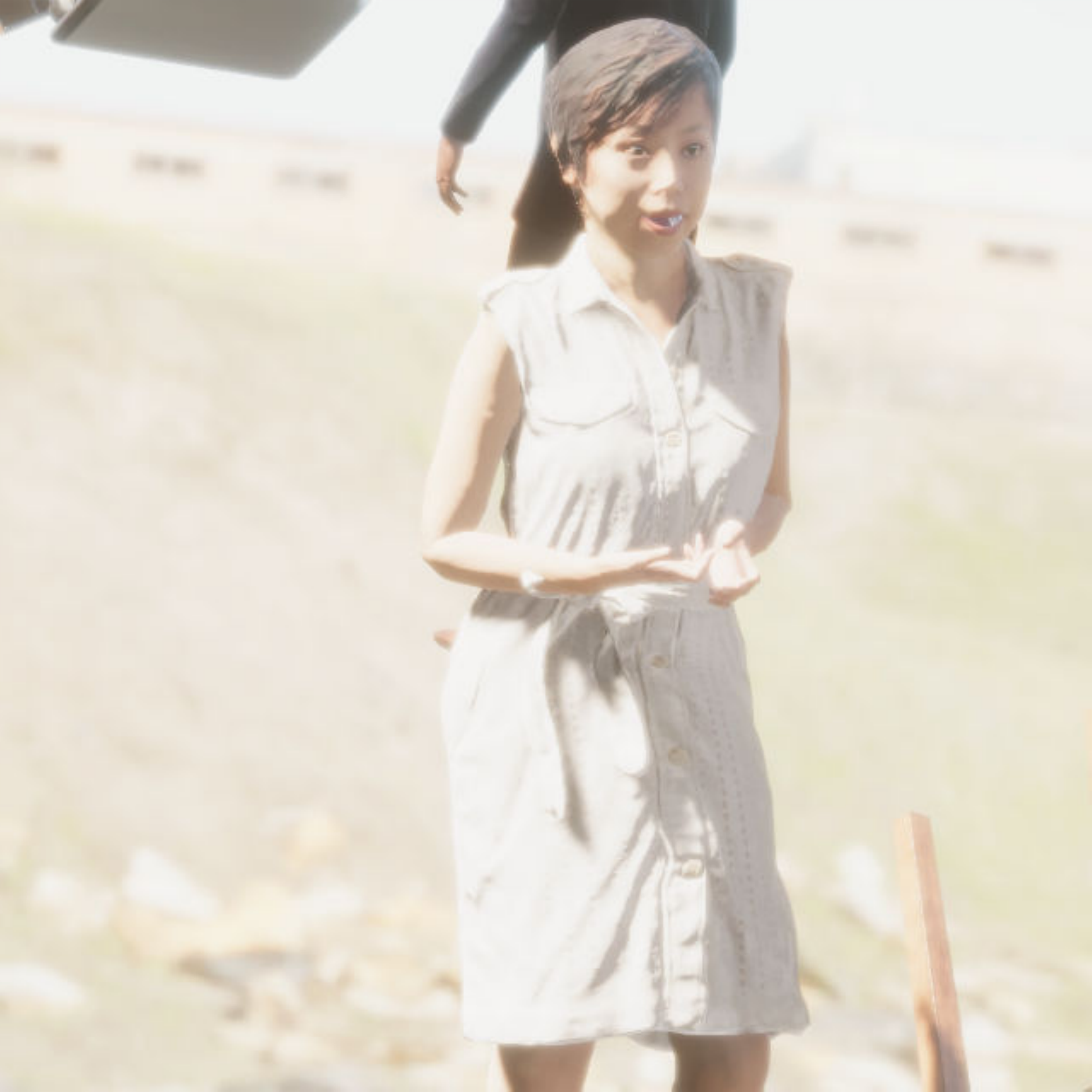}}
    \end{subfigure}
    \begin{subfigure}[t]{0.16\textwidth}
        {\includegraphics[height=2.75cm]{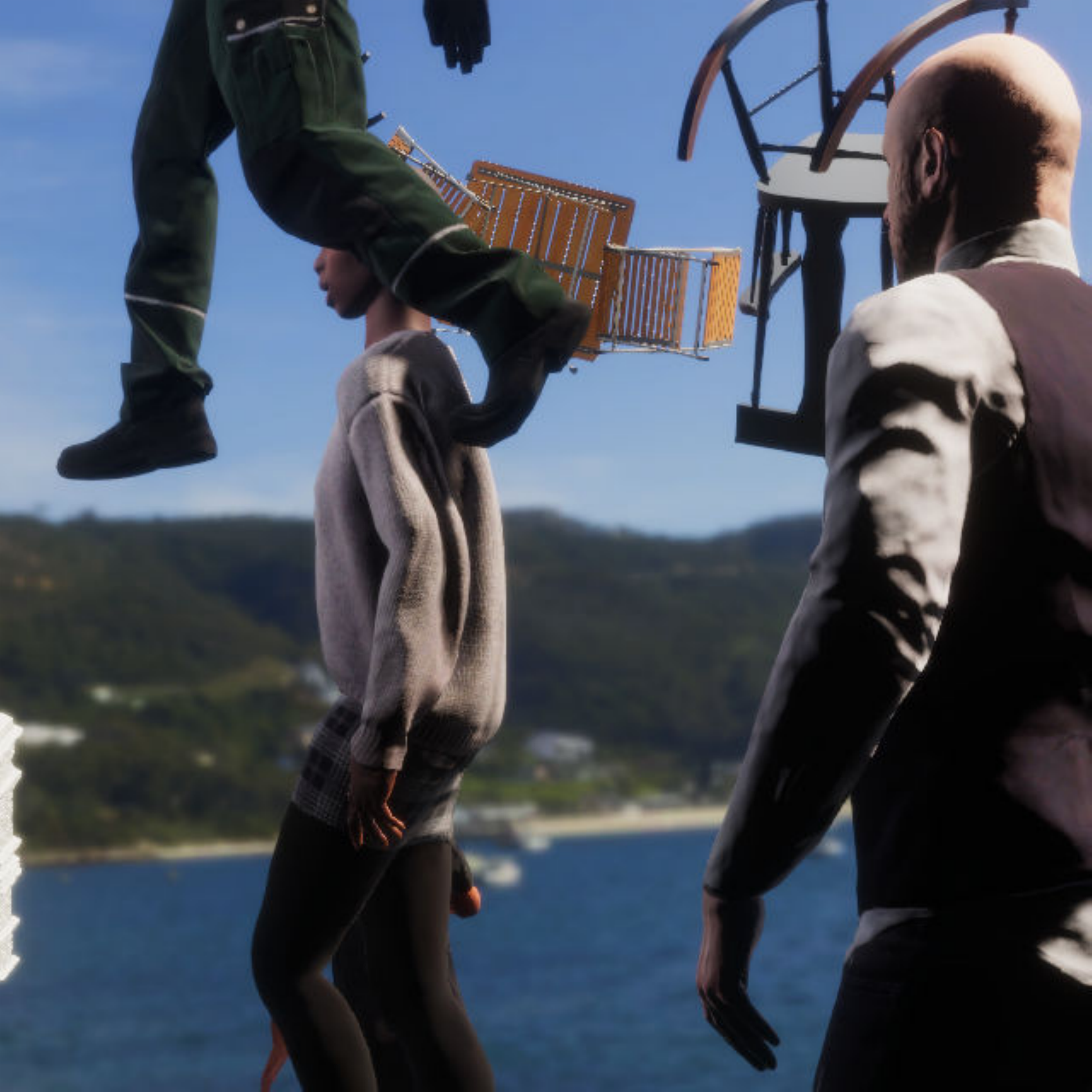}}
    \end{subfigure}
    \begin{subfigure}[t]{0.16\textwidth}
        {\includegraphics[height=2.75cm]{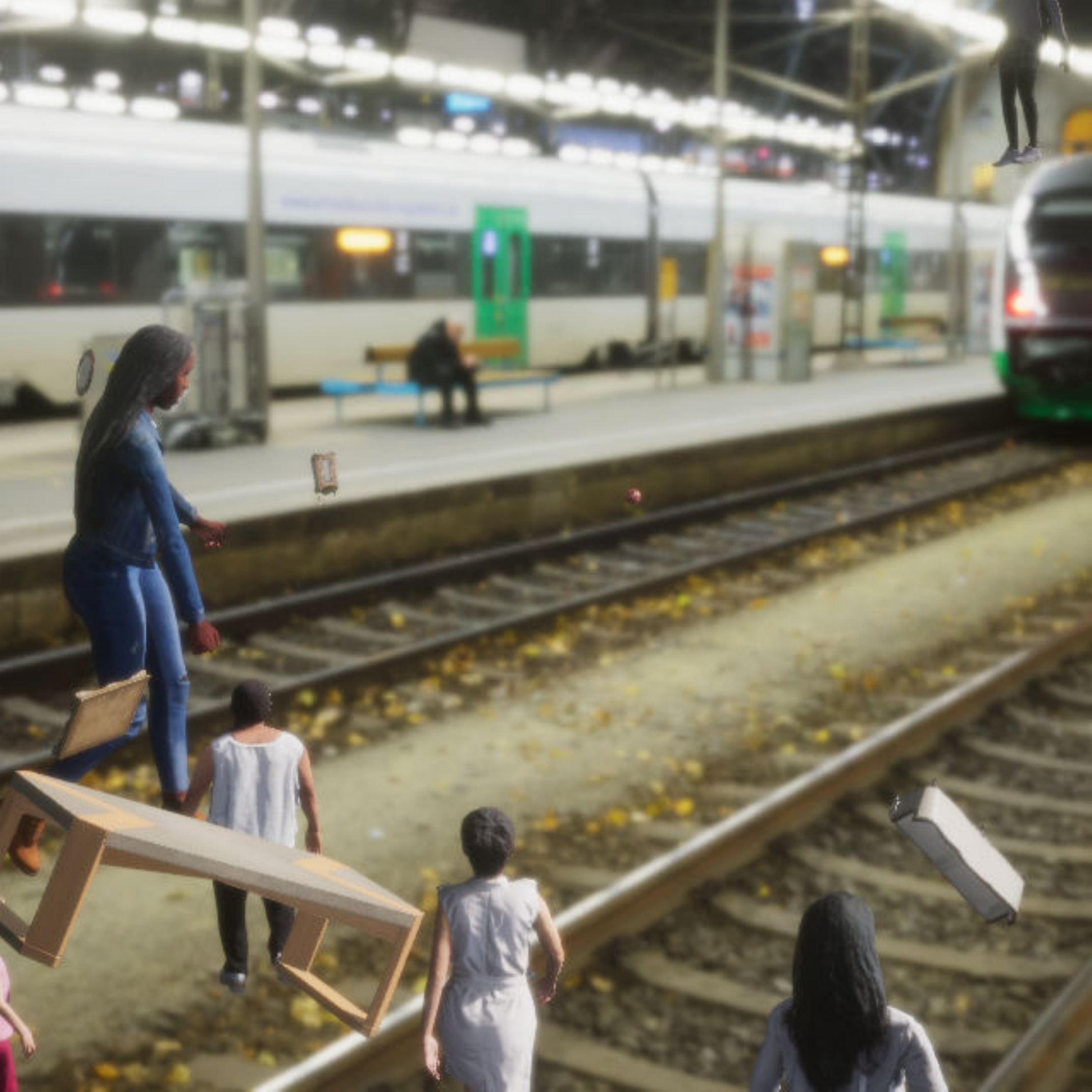}}
    \end{subfigure}
    \begin{subfigure}[t]{0.16\textwidth}
       {\includegraphics[height=2.75cm]{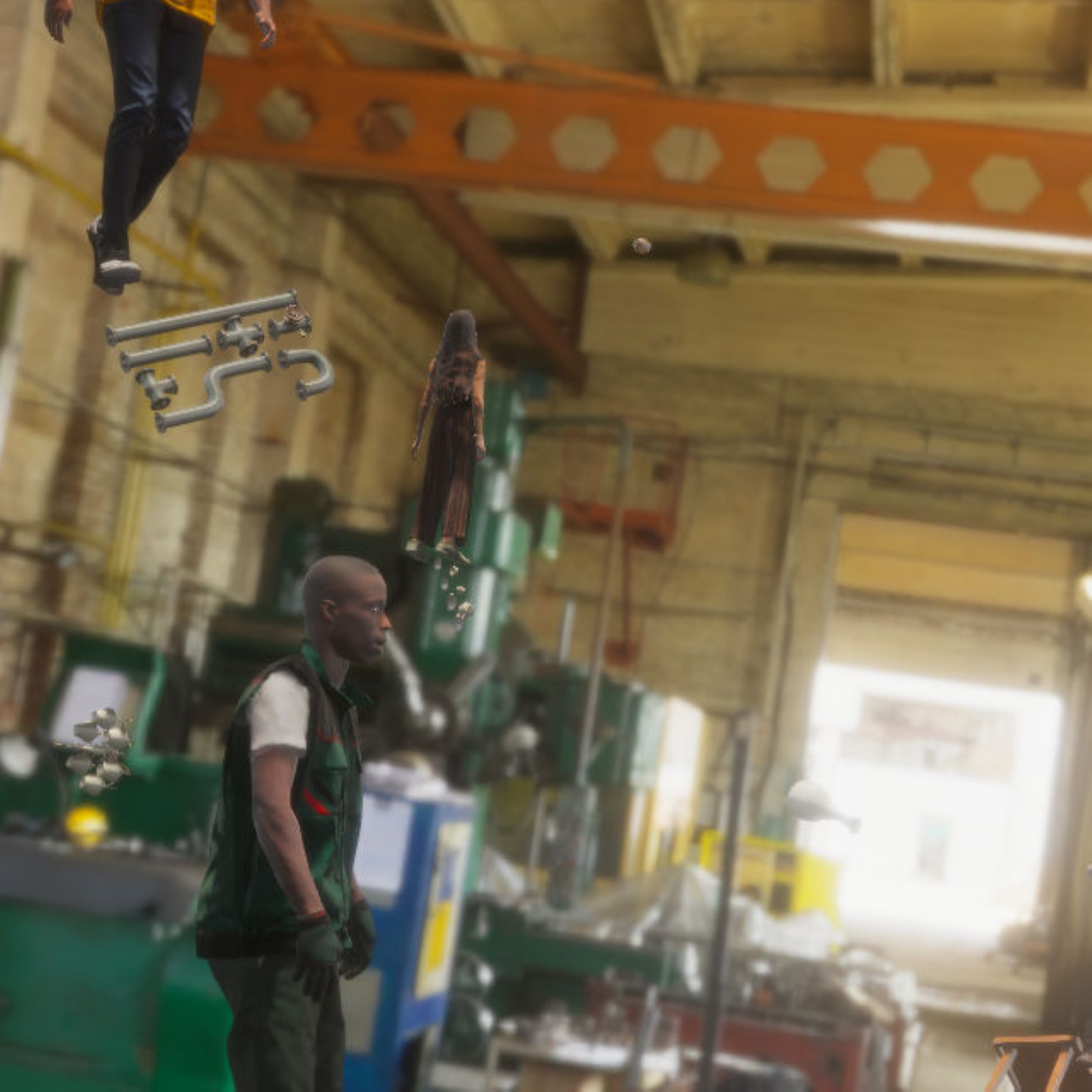}}
    \end{subfigure}
    \begin{subfigure}[t]{0.16\textwidth}
        {\includegraphics[height=2.75cm]{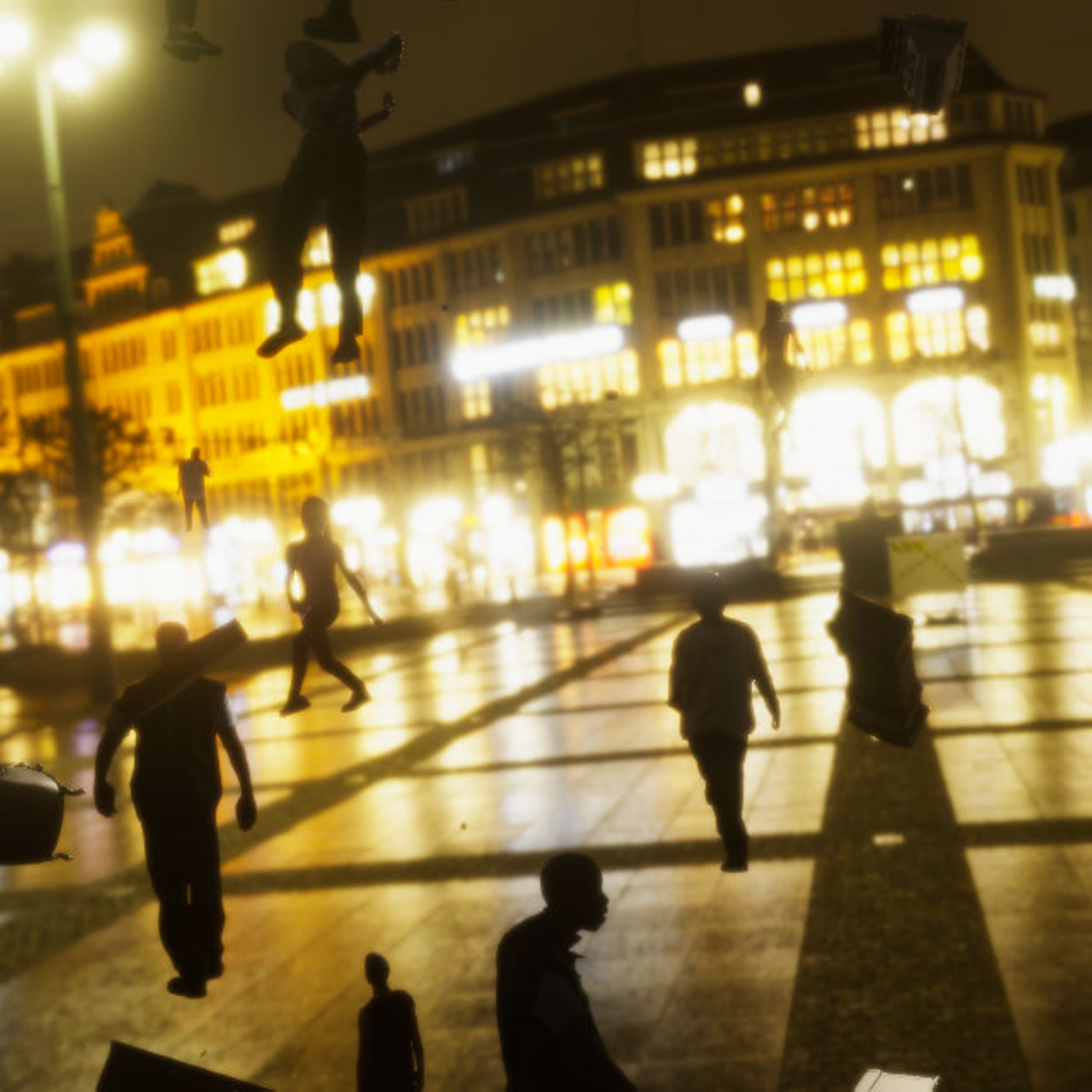}}
    \end{subfigure}
    \\
    \begin{subfigure}[t]{0.16\textwidth}
        {\includegraphics[height=2.75cm]{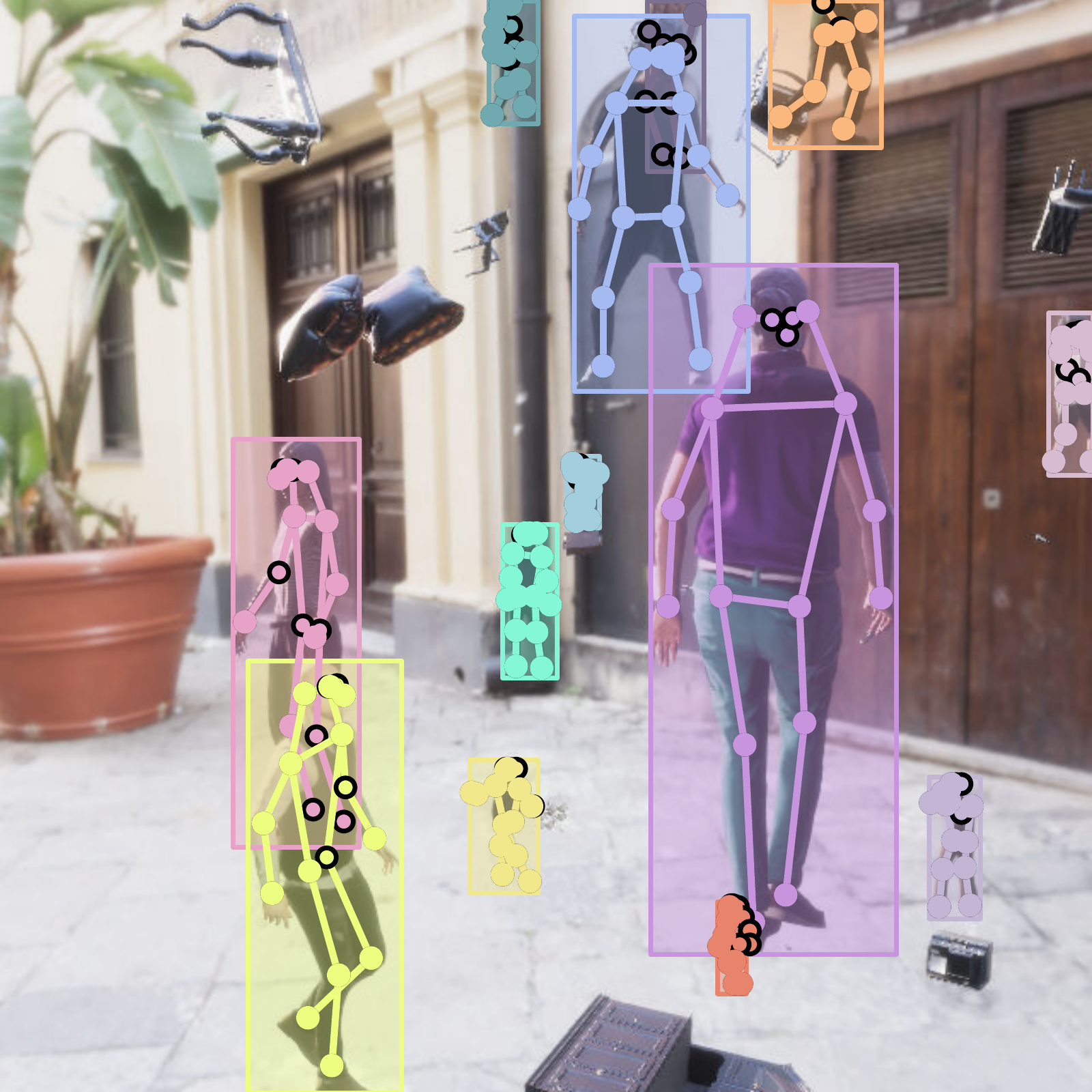}}
    \end{subfigure}
    \begin{subfigure}[t]{0.16\textwidth}
        {\includegraphics[height=2.75cm]{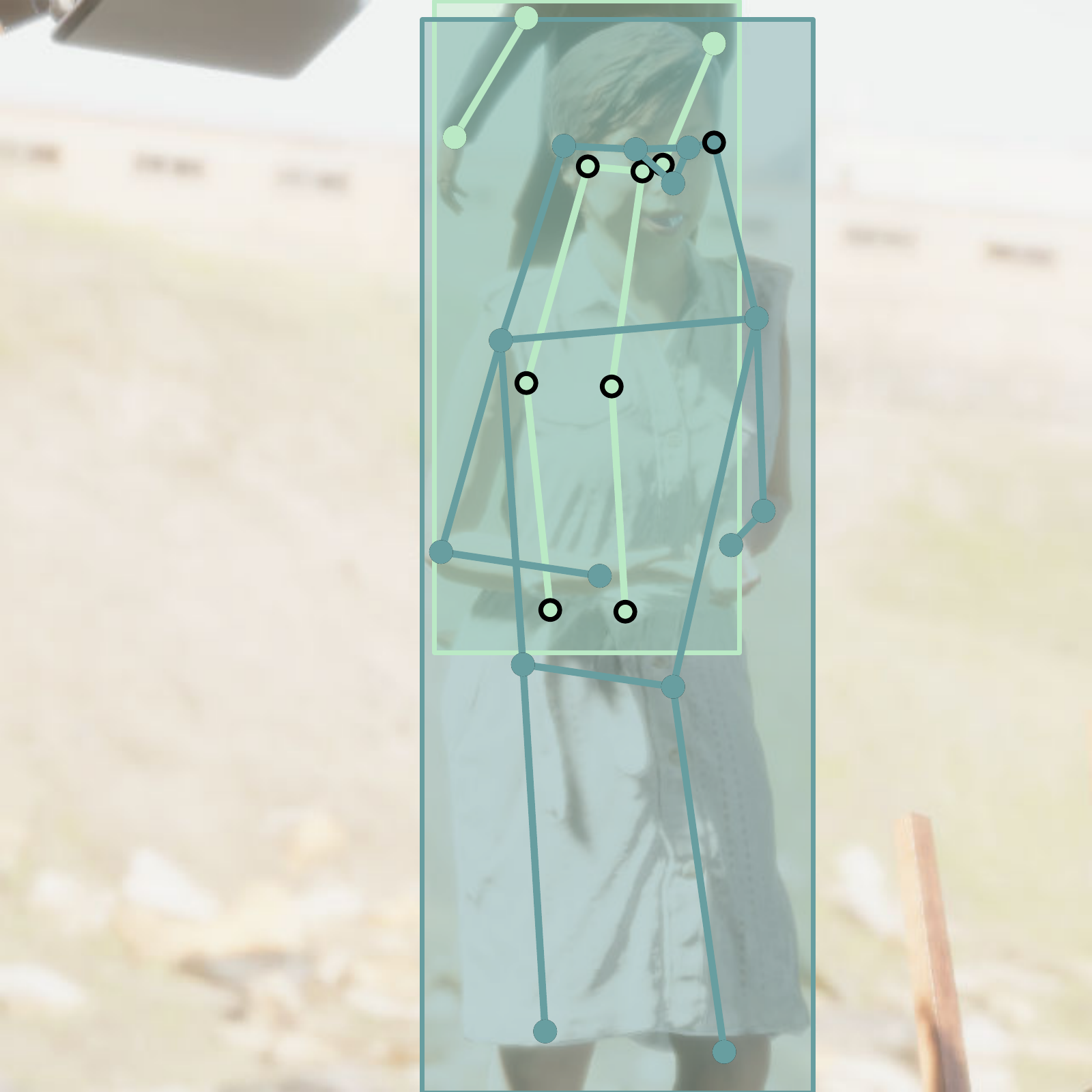}}
    \end{subfigure}
    \begin{subfigure}[t]{0.16\textwidth}
        {\includegraphics[height=2.75cm]{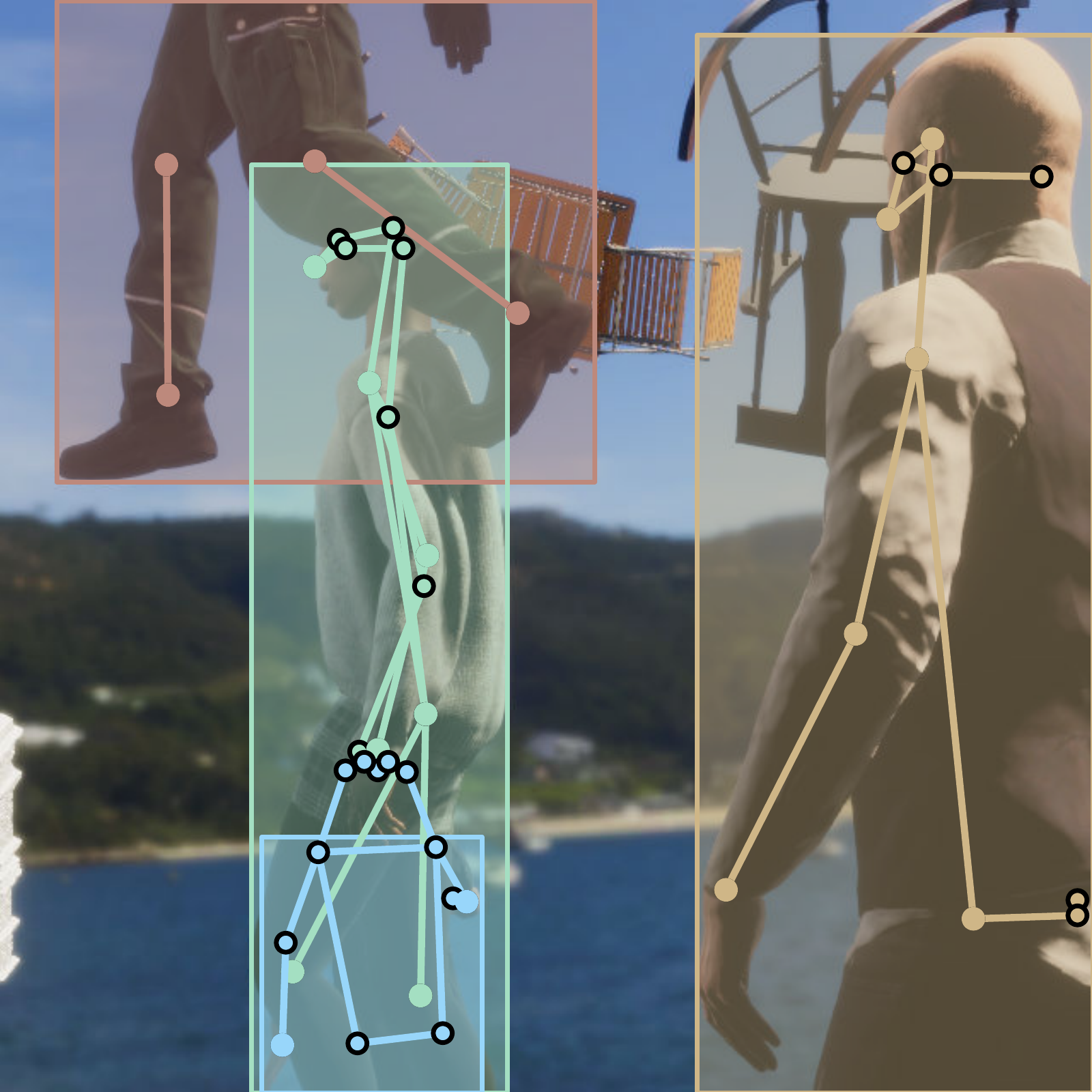}}
    \end{subfigure}
    \begin{subfigure}[t]{0.16\textwidth}
        {\includegraphics[height=2.75cm]{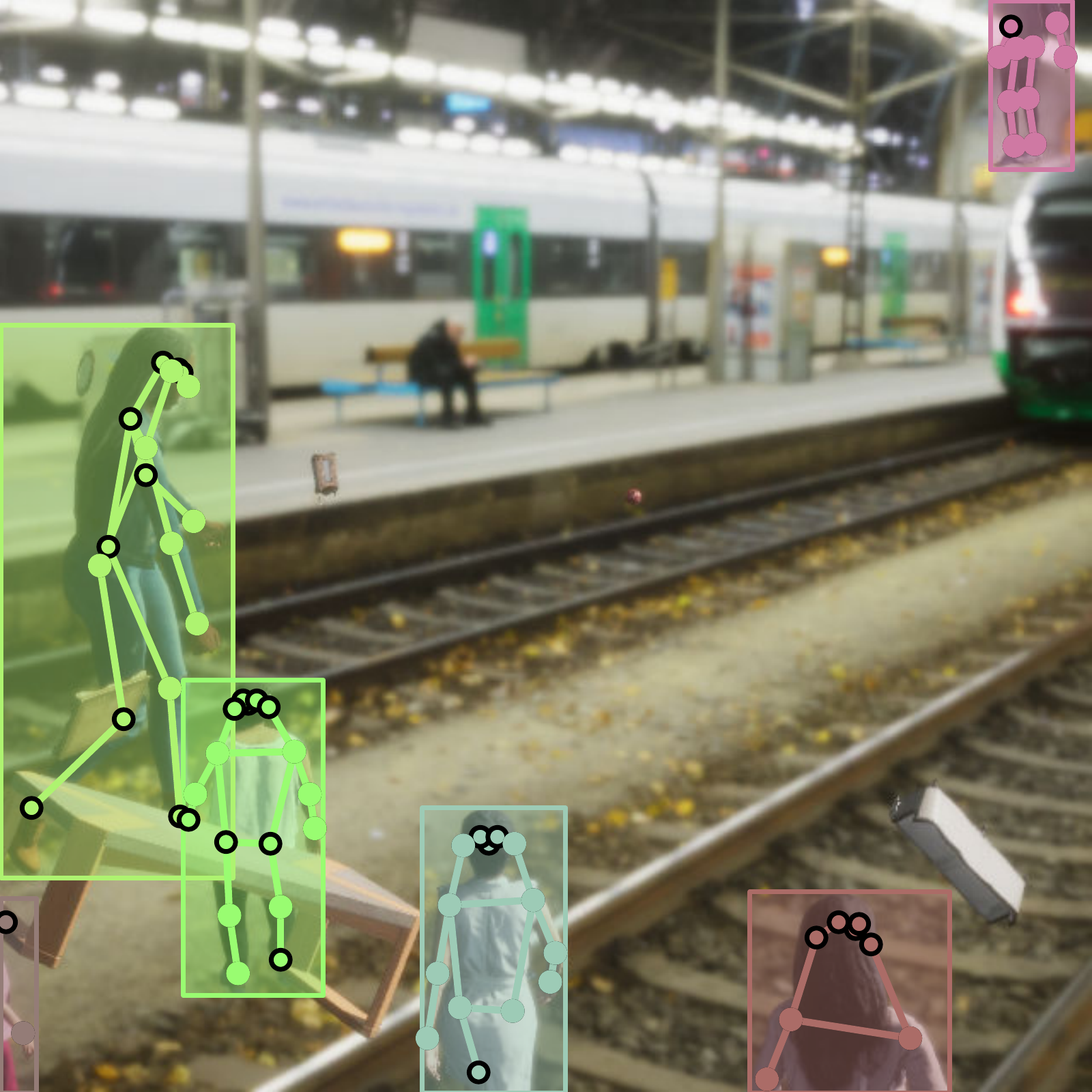}}
    \end{subfigure}
    \begin{subfigure}[t]{0.16\textwidth}
       {\includegraphics[height=2.75cm]{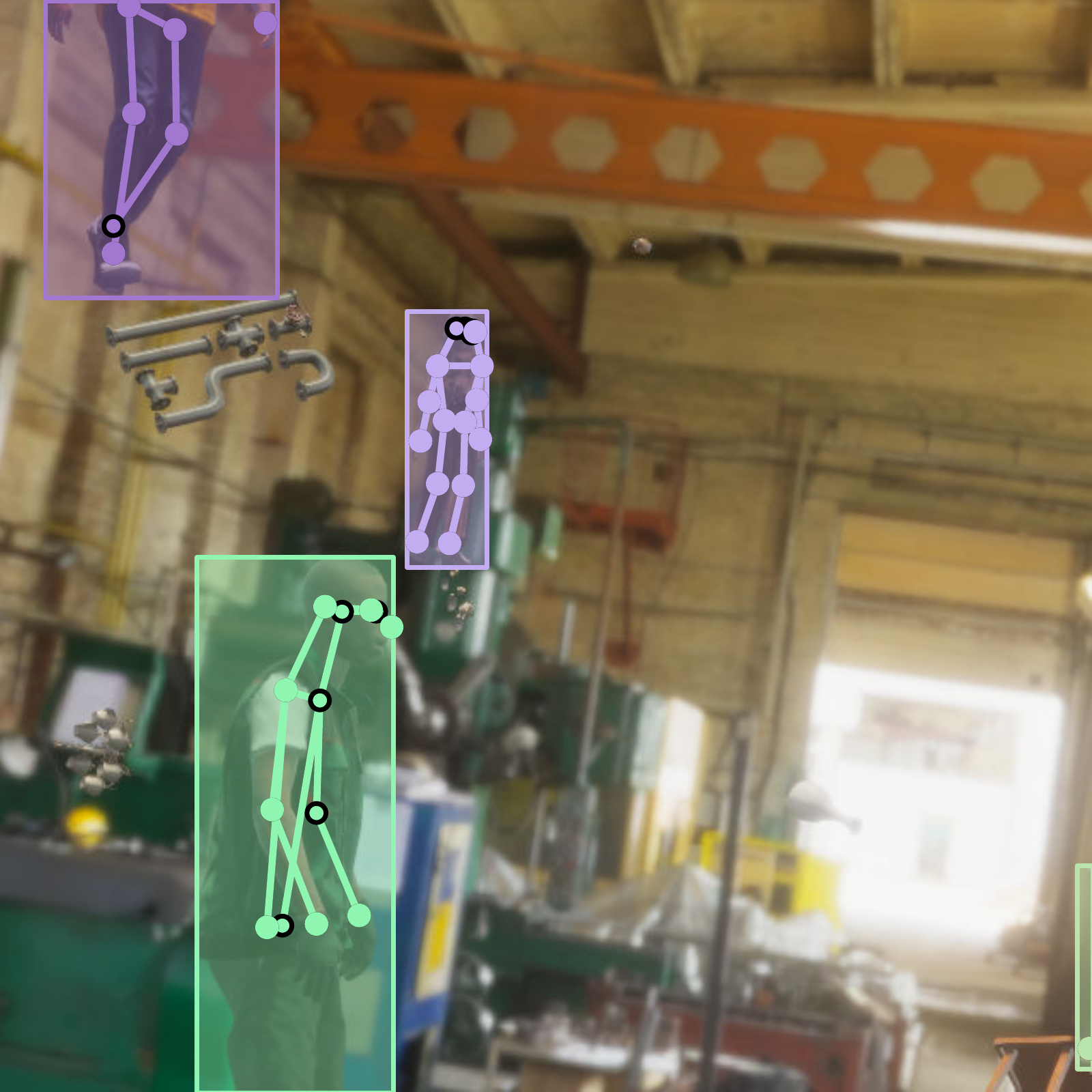}}
    \end{subfigure}
    \begin{subfigure}[t]{0.16\textwidth}
        {\includegraphics[height=2.75cm]{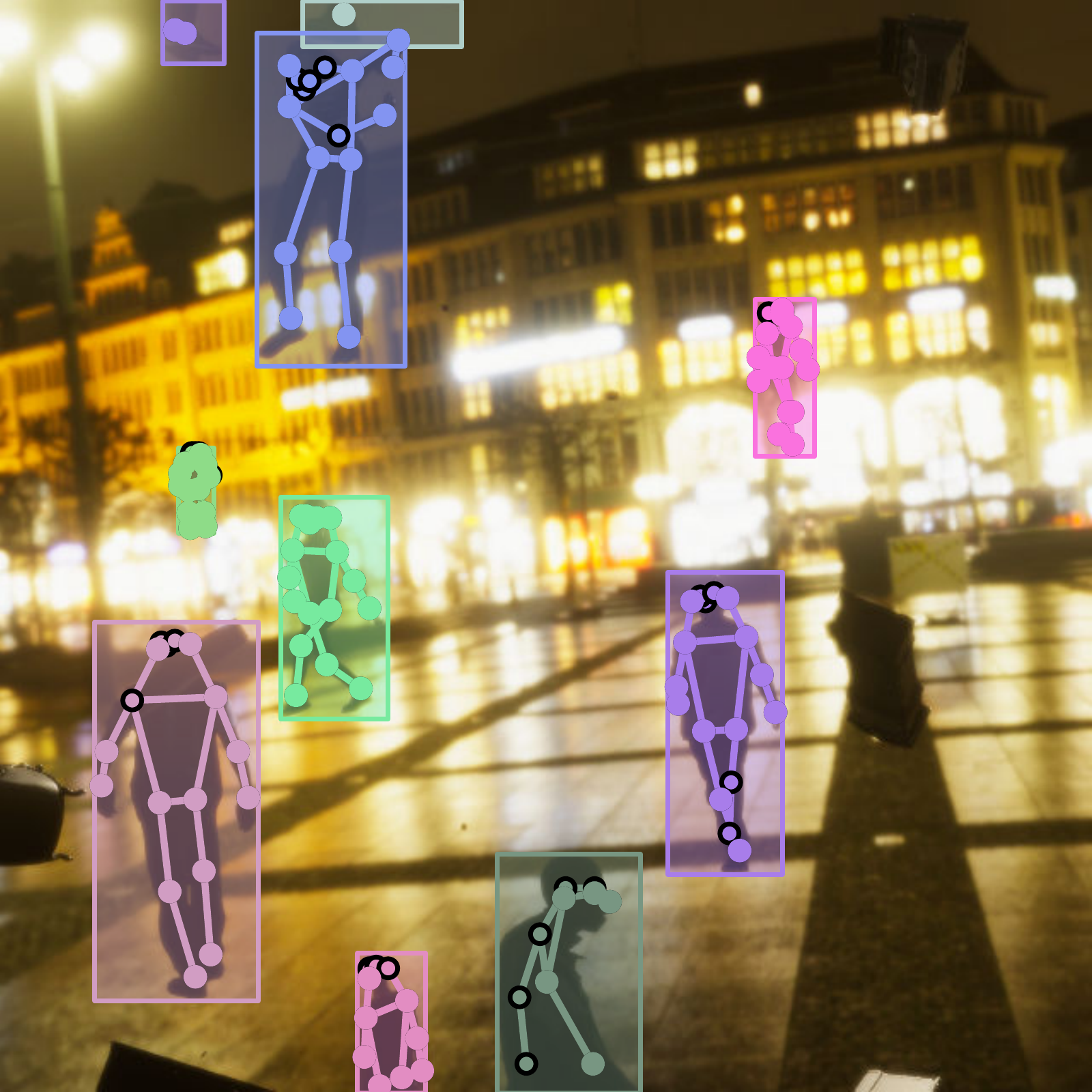}}
    \end{subfigure}
    \\
    \vspace{0.1cm}
    \caption{\textbf{Examples from PSP-HDRI$+$}. Top: RGB image; Bottom: bounding box and keypoint annotations.}
    \label{fig:psp-hdri+}
\end{figure*}

We spawn the human assets and other random shapes (occluders and distractors) at the scene's center, and the HDRI skybox background wraps around the scene. As with PSP, we use primitive 3D game objects as occluders and distractors, albeit with randomized high-quality HDRI textures. We modified the camera randomization to orbit around the scene and capture a diverse range of perspectives of the human assets and take advantage of all viewpoints in the HDRI skyboxes. To this end, we have produced a new version of PSP, dubbed PSP-HDRI. We performed additional ablation studies to guide the design of our data generator for better pre-training and OOD generalization, which we call PSP-HDRI$+$. Some examples are shown in fig.~\ref{fig:psp-hdri+}.

Our data generator produces sub-pixel-perfect labels for all objects, regardless of the amount of occlusion, and how far away or how small the object appears from the camera. However, real datasets rarely have very accurate labeling as human labeling is a subjective task and prone to the errors and negligence of human annotators. Further, there exist inevitable label distribution discrepancies between different domains. In PSP-HDRI$+$, we experimented with crude label adaptation for bounding boxes and keypoints to overcome the label gap. We use the bounding box and keypoint annotations from the COCO training set as our comparison baseline. Our criteria for bounding box label adaptation are: firstly, remove boxes smaller than the smallest box in COCO that has keypoint annotations; and secondly, remove boxes whose size to image size ratio is less than or equal to the same ratio in COCO. This strategy effectively removes boxes that are too small. For keypoint label adaptation, we ensure that for each box area range, the probability of having annotations for each keypoint matches between the synthetic and COCO, by randomly removing surplus keypoint annotations. Note that COCO has six box area ranges. This keypoint adaptation statistically matched the labeling inaccuracies of human annotators and the label distribution in the COCO dataset. As a result, we observed improvements in large-scale human-annotated benchmark datasets.

%% file: 05_Experiments.tex
\section{Experiments and Results}
\input{tables/tables_transfer_coco_testdev_results}
\input{tables/tables_hybrid_finetuned_coco_to_ood_results}
\input{tables/tables_hdri_ablations_coco_val}

\input{tables/tables_transfer_psphdriplus_motsynth_to_coco_mpii}
We generated our synthetic datasets using domain randomization with na\"ive (random uniform) sampling from the set of our data generator parameters; we also generated our synthetic data from three random generation seeds in order to obtain a fair comparison. 
For our experiments we considered the tasks of human detection and keypoint localization. To obtain a suite of benchmarks, we used three pre-training strategies: 
\begin{enumerate*}[label=(\arabic*)]
    \item random initialization (no pre-training),
    \item pre-training with ImageNet,
    \item and pre-training with various sizes of synthetic data.
\end{enumerate*} 
We then fine-tuned all these models on various sizes of real data.
We use the same training regiment without any hyperparameter tuning for all our models. Lastly, we compared the performance of these models on in-distribution and OOD sets to establish the generalization and applicability of each model across a wide range of real-world examples. 

For our real data training, we used the COCO person training dataset, divided into overlapping sets of $641$, $6411$, $32057$, and $64115$ images. The COCO person-val2017 and test-dev2017 have $2693$ and $20288$ images, respectively. We also used the MPII Human Pose dataset \citep{andriluka14mpii} divided into $16712$ training and $696$ validation sets. We compare our synthetic data to another large-scale synthetic dataset generated from scenes of the GTA V game with crowds of people walking in them, called MOTSynth~\citep{fabbri2021motsynth}; we randomly selected $50000$ training and $5008$ validation images from MOTSynth to obtain a fair comparison.
For our OOD tests we used the following datasets with their respective image numbers: CrowdPose~\citep{li2019crowdpose} Trainval ($12000$ images), Leeds Sports Pose~\citep{Johnson10lspet} ($10000$ images), Occluded Humans~\citep{zhang2019pose2seg} ($4731$ images), and MOT17~\citep{milan2016mot16} ($5316$ images).

\subsection{Training Strategy}
For all our experiments, we use the Detectron2 Keypoint R-CNN \texttt{R50-FPN} variant~\citep{he2017mask} with ResNet-50~\citep{he2016deep} plus Feature Pyramid Network (FPN)~\citep{lin2017feature} backbones. 
We trained our models from scratch (random weight initialization) with Group Normalization (GN)~ \citep{wu2018group, wu2019detectron2, he2019rethinking}.
Similar to~\citep{ebadi2021peoplesanspeople}, for all our models, we use a learning rate annealing strategy, where we reduce the learning rate when the validation keypoint Average Precision (AP) metric has stopped improving. Our models benefited from reducing the learning rate by a factor of $10\times$ once learning has stagnated based on a threshold (epsilon) for several epochs (patience period). Every time the patience period ends, we reduce the learning rate, and halve epsilon and the next patience period. We perform the learning rate reduction three times. Every time the learning rate is reduced, we restore the weights from the model checkpoint that achieves the highest metrics on the validation. Thus we ensure that the last model checkpoint is also the best performing one. 

For both pre-training and fine-tuning experiments, we set the initial learning rate to $0.02$, the initial patience to $38$ epochs, and the initial epsilon to $5$. We perform a \emph{linear} warm-up period of $1000$ iterations at the start of training, where we slowly increase the learning rate to the initial learning rate. The weight decay is $0.0001$, and momentum is $0.9$.
We use $8$ NVIDIA Tesla V100 GPUs on synchronized SGD with a mini-batch size of $2$ images per GPU; the mean pixel value and standard deviation from ImageNet is used for image normalization. We do not change the default augmentations used by Detectron2 and perform the evaluation every two epochs. Additionally, we fix the model seed to improve reproducibility. When we train on real data, we also evaluate real data from the same distribution. Likewise, we evaluate on synthetic data from the same distribution when we train on synthetic sets.

\subsection{Pre-Training Benchmarks}
Tab.~\ref{tab:transfer_coco_multi} shows a comparison between models with no pre-training, ImageNet, and synthetic pre-training. 
Unsurprisingly, we find that ImageNet pre-training improves the model performance over training from scratch for any real data size. 
Interestingly, even small sets of $4.9 \times 10^3$ synthetic images obtain performance that is better or on par with ImageNet pre-training, with larger effects for few-shot transfer. 
While we observe improvements with pre-training on all sizes of synthetic data, our largest set of $245 \times 10^3$ images achieves the best performance. In~\citep{ebadi2021peoplesanspeople} the authors showed positive trends for much larger sets of synthetic data, and we do expect the numbers on tab.~\ref{tab:transfer_coco_multi} to improve if more synthetic data is used for pre-training.

Further, we measure the generalization ability of models with no pre-training, ImageNet, and synthetic pre-training on a wide range of OOD sets for the tasks of human detection and keypoint localization. 
In tab.~\ref{tab:id_ood_coco_100} all our models are fine-tuned with the entire COCO person training data. 
We observe only marginal improvements with ImageNet pre-training over training from scratch. 
Additionally, on average, MOTSynth pre-trained models are inferior to those of a model pre-trained with a similarly-sized PSP-HDRI dataset of $49 \times 10^3$ images. 
We also reconfirm that OOD generalization of even $4.9 \times 10^3$ synthetic images for pre-training is on par with ImageNet pre-training. The best overall performance is also still achieved with larger synthetic pre-training sets. 
We hypothesize that since PSP-HDRI synthetic data is task-specific, it contains the necessary signals and representations needed for fine-tuning and better generalization on human-centric tasks. 

\subsection{Zero-Shot Ablation Studies}
We performed a set of ablation studies using metrics obtained from a suite of OOD datasets to guide our data generator design choices. In tab.~\ref{tab:ablation_ood} we list the zero-shot performance of models trained with approximately $50000$ synthetic images in each row. Beyond PSP-HDRI, we explored the effect of the following: bounding box label adaptation; bounding box$+$keypoint label adaptation; using no occluder/distractor objects; using high-quality 3D models from Poly Haven~\footnote{\url{https://polyhaven.com/models}} instead of our primitive 3D game objects as occluders/distractors; disabling the clothing texture randomization (shader graph randomizer); using Subpixel Morphological Anti-Aliasing (SMAA) which gives graphics a smoother appearance; and lastly, using only simple animations, such as walking, running, and standing idle, instead of PSP-HDRI's diverse animation library. 

Interestingly, we observed performance boosts over the PSP-HDRI baseline with every modification, except when we removed our occluder/distractor objects from the scene. We then combined all the modifications that hinted toward positive trends together to form the PSP-HDRI$+$ set, which includes: ``box + kpt adapt.", ``Poly Haven occluders", ``no shadergrpah", ``SMAA", and ``simple anims". PSP-HDRI$+$ obtains the best overall zero-shot performance. We also trained PSP-HDRI$+$ with random crop augmentation during training, which improved zero-shot performance on the MOTSynth and MOT17. The MOTSynth trained model performs exceedingly well on MPII val and Crowdpose Trainval. We also report its performance on in-distribution MOTSynth val set, which is unsurprisingly better than the rest. The MOTSynth dataset was designed to expedite pre-training for the MOT17 challenge, hence its better performance on the MOT17 set. However, on the larger and more diverse COCO test-dev2017 and COCO person-val2017, the PSP-HDRI$+$ performs better than MOTSynth.

\subsection{Comparison with Another Synthetic Counterpart}
Lastly, we pre-train our models on PSP-HDRI$+$ and a similarly-sized set from MOTSynth. We fine-tuned either of them on two real datasets COCO and MPII and tested them in-distribution (COCO test-dev2017, person-val2017, and MPII val) and OOD (Crowdpose Trainval, Leeds Sports, Occluded Humans, MOT17) as listed in tab.~\ref{tab:scratch_results_ood_mpii}. As before, models pre-trained on PSP-HDRI$+$ outperform those pre-trained on MOTsynth when tested on in-distribution test sets (COCO test-dev2017, person-val2017, and MPII val).
Furthermore, the OOD generalization of PSP-HDRI$+$ trained models also tends to be more robust than MOTSynth. Interestingly, a model pre-trained on MOTSynth and then fine-tuned on COCO (second row) is only marginally better than a model pre-trained on PSP-HDRI$+$ (first row) when tested on MOTSynth again. When tested on MOT17, the model pre-trained on MOTSynth and then fine-tuned on COCO data (second row) performs on par with a model trained only on MOTSynth (bbox AP of $32.46$ vs. $32.75$).
Conversely, the model pre-trained on PSP-HDRI$+$ and fine-tuned on COCO (first row) benefits from a more than double improvement on bounding box AP when tested on MOT17 (bbox AP of $32.01$ vs. $13.97$).
This result indicates that after fine-tuning, the model forgets the MOTSynth pre-training. 

It is worth noting that the entire MOTSynth dataset has $764$ sequences where the background is primarily static. In PSP-HDRI$+$, we use $510$ HDRI backgrounds, but the camera randomizer allows us to capture unique and diverse perspectives from these spherical images and the human assets. Additionally, PSP-HDRI$+$ has more diverse human poses, the image quality is improved with SMAA, the lighting randomization produces unique and diverse light settings for our scenes and provides image augmentation out-of-the-box, the occluder objects play an important role in making the model more robust in cases of partially or almost entirely occluded foreground objects, and finally our label adaptation can statistically match the target label distributions, further reducing the visual quality and label gaps.

%% file: tables/tables_transfer_coco_testdev_results.tex
\begin{table*}[htb!]
\caption{\textbf{Comparison between models with no pre-training, ImageNet, and synthetic pre-training.}
All models are fine-tuned on different sizes of real data as shown on the first column. We used three different dataset sizes for our synthetic pre-training, each of which are generated using three different random seeds.
The results are reported for COCO test-dev2017 keypoint metrics. 
We refer to PSP-HDRI as synth.
}
\centering
\begin{adjustbox}{max width=0.61\textwidth} 
\centering
\begin{tabular}{crccccc} 
\toprule
    \begin{tabular}{@{}c@{}}real\\fine-tune\end{tabular} & \begin{tabular}{@{}c@{}}pre-train\end{tabular} &                AP &              $\textrm{AP}^{\textrm{\textit{IoU=.50}}}$ &             $\textrm{AP}^{\textrm{\textit{IoU=.75}}}$ &               $\textrm{AP}^{\textrm{\textit{large}}}$ &              $\textrm{AP}^{\textrm{\textit{medium}}}$                \\
\midrule
\parbox[t]{2mm}{\multirow{5}{*}{\rotatebox[origin=c]{90}{641}}} & - &              6.40 &             20.30 &              2.40 &              7.90 &              5.60      \\
 & ImageNet &              21.90 &             50.90 &              15.90 &              26.90 &              18.80      \\
      & $\text{4.9}\times10^3$ synth &  25.00 $\pm$ 0.14 &  52.37 $\pm$ 0.45 &  20.67 $\pm$ 0.21 &  29.23 $\pm$ 0.34 &  22.60 $\pm$ 0.00      \\
      & $\text{49}\times10^3$ synth &  41.73 $\pm$ 0.17 &  69.00 $\pm$ 0.33 &  42.53 $\pm$ 0.25 &  47.33 $\pm$ 0.33 &  38.77 $\pm$ 0.09      \\
      & \textbf{$\text{245}\times\text{10}^\text{3}$ synth} &  \textbf{46.00 $\pm$ 0.08} &  \textbf{72.93 $\pm$ 0.17} &  \textbf{48.17 $\pm$ 0.12} &  \textbf{52.00 $\pm$ 0.08} &  \textbf{42.70 $\pm$ 0.08}      \\
\midrule
\parbox[t]{2mm}{\multirow{5}{*}{\rotatebox[origin=c]{90}{6411}}} & - &             37.30 &             67.60 &             35.60 &             43.80 &             33.30      \\
 & ImageNet &             44.20 &             73.90 &             45.00 &             52.40 &             38.80      \\
      & $\text{4.9}\times10^3$ synth &  42.50 $\pm$ 0.29 &  71.73 $\pm$ 0.29 &  43.13 $\pm$ 0.29 &  49.30 $\pm$ 0.37 &  38.37 $\pm$ 0.26      \\
      & $\text{49}\times10^3$ synth &  51.90 $\pm$ 0.92 &  79.30 $\pm$ 0.57 &  55.53 $\pm$ 1.16 &  59.17 $\pm$ 0.90 &  47.60 $\pm$ 0.92      \\
      & \textbf{$\text{245}\times\text{10}^\text{3}$ synth} &  \textbf{53.50 $\pm$ 0.65} &  \textbf{80.50 $\pm$ 0.36} &  \textbf{57.83 $\pm$ 0.87} & \textbf{ 61.07 $\pm$ 0.60} &  \textbf{48.97 $\pm$ 0.74}      \\
\midrule
\parbox[t]{2mm}{\multirow{5}{*}{\rotatebox[origin=c]{90}{32057}}} & - &             55.80 &             82.00 &             60.60 &             64.20 &             50.70      \\
 & ImageNet &             57.50 &             83.60 &             62.40 &             66.40 &             51.70      \\
      & $\text{4.9}\times10^3$ synth &  56.47 $\pm$ 0.12 &  82.90 $\pm$ 0.00 &  61.03 $\pm$ 0.17 &  64.70 $\pm$ 0.22 &  51.33 $\pm$ 0.17      \\
      & $\text{49}\times10^3$ synth &  59.13 $\pm$ 0.34 &  84.57 $\pm$ 0.17 &  64.43 $\pm$ 0.50 &  67.30 $\pm$ 0.37 &  54.03 $\pm$ 0.34      \\
      & \textbf{$\text{245}\times\text{10}^\text{3}$ synth} &  \textbf{60.30 $\pm$ 0.22} &  \textbf{85.10 $\pm$ 0.08} &  \textbf{66.00 $\pm$ 0.43} &  \textbf{68.67 $\pm$ 0.26} &  \textbf{55.07 $\pm$ 0.25}      \\
\midrule
\parbox[t]{2mm}{\multirow{5}{*}{\rotatebox[origin=c]{90}{64115}}} & - &             62.00 &             86.20 &             68.10 &             70.50 &             56.70      \\
 & ImageNet &             62.40 &             86.60 &             68.60 &             71.20 &             56.80      \\
      & $\text{4.9}\times10^3$ synth &  62.03 $\pm$ 0.05 &  86.23 $\pm$ 0.05 &  68.20 $\pm$ 0.08 &  70.53 $\pm$ 0.12 &  56.73 $\pm$ 0.05      \\
      & $\text{49}\times10^3$ synth &  62.93 $\pm$ 0.12 &  86.90 $\pm$ 0.00 &  69.30 $\pm$ 0.16 &  71.30 $\pm$ 0.24 &  57.70 $\pm$ 0.14      \\
      & \textbf{$\text{245}\times\text{10}^\text{3}$ synth} &  \textbf{63.47 $\pm$ 0.24} &  \textbf{87.17 $\pm$ 0.12} &  \textbf{69.83 $\pm$ 0.42} &  \textbf{71.90 $\pm$ 0.16} &  \textbf{58.17 $\pm$ 0.31}      \\
\bottomrule
\end{tabular}
\end{adjustbox}
\label{tab:transfer_coco_multi}
\end{table*}

%% file: tables/tables_hybrid_finetuned_coco_to_ood_results.tex
\begin{table*}[htb!]
\caption{\textbf{Keypoint AP for in-distribution and OOD sets.} We compare models with no pre-training, pre-training with ImageNet, and synthetic data, where all are fine-tuned with the COCO 64115 set. We refer to PSP-HDRI as synth.}
\centering
\begin{adjustbox}{max width=0.8\textwidth}
\begin{tabular}{r|c|c|c|c|c|c|c|c} 
\toprule
    pre-training data        & COCO test-dev2017          & COCO person-val2017 & MPII val & Crowdpose Trainval & Leeds Sports     & Occluded Humans & MOTSynth & MOT17 (bbox AP)  \\
\midrule
    -                        & 62.00             & 65.12 & 69.42                           & 69.78              & 26.69            & 30.34 & 15.63 & 32.04            \\
    ImageNet                 & 62.40             & 65.10 & 69.74                           & 69.37              & 27.78            & 30.68 & 15.93 & 32.31             \\
MOTSynth & 62.60 & 65.81 & 70.07 & 69.85 & 26.09 & 30.56 & 16.53 & \textbf{32.46} \\
\midrule
$\text{4.9}\times10^3$ synth & 62.03 $\pm$ 0.05  & 65.34 $\pm$ 0.12 & 69.47 $\pm$ 0.40     & 69.72 $\pm$ 0.35   & 26.56 $\pm$ 0.47 & 30.62 $\pm$ 0.06 & 15.87 $\pm$ 0.18 & 32.01 $\pm$ 0.21  \\
$\text{49}\times10^3$ synth  & 62.93 $\pm$ 0.12  & 66.28 $\pm$ 0.07 & 70.15 $\pm$ 0.25      & 70.27 $\pm$ 0.14   & 28.53 $\pm$ 0.57 & \textbf{31.35 $\pm$ 0.51} & 16.37 $\pm$ 0.24 & 32.21 $\pm$ 0.35  \\
\textbf{$\text{245}\times\text{10}^\text{3}$ synth}  & \textbf{63.47 $\pm$ 0.24}  & \textbf{66.75 $\pm$ 0.20} & \textbf{70.38 $\pm$ 0.11}     & \textbf{70.57 $\pm$ 0.21}   & \textbf{29.85 $\pm$ 0.75} & 31.34 $\pm$ 0.25 & \textbf{16.72 $\pm$ 0.29} & 32.01 $\pm$ 0.11  \\
\bottomrule
\end{tabular}
\end{adjustbox}
\label{tab:id_ood_coco_100}
\end{table*}

%% file: tables/tables_hdri_ablations_coco_val.tex
\begin{table*}[htb!]
\centering
\caption{\textbf{PSP-HDRI ablation results for OOD sets.} 
All models are trained with $49\times10^3$ images.}
\begin{adjustbox}{max width=0.8\textwidth, center}
\centering
\begin{tabular}{r|r|r|r|r|r|r|r|r}
\toprule
training data & COCO test-dev2017 & COCO person-val2017 & MPII val & Crowdpose Trainval & Leeds Sports & Occluded Humans & MOTSynth & MOT17 (bbox AP) \\
\midrule
PSP-HDRI                   &  6.60 &  7.36 & 11.91 &  7.18 & 0.81 & 3.59 &  9.37 &  8.74 \\
box adapt.                 &  9.00 & 10.05 & 16.13 & 10.46 & 1.89 & 5.82 &  8.95 &  9.74 \\
box + kpt adapt.           & 10.10 & 11.12 & \underline{\textbf{19.08}} & 12.63 & \textbf{2.23} & 7.43 &  9.32 & 10.58 \\
No occluders               &  5.30 &  6.20 & 10.85 &  5.53 & 0.52 & 2.64 &  8.26 &  6.32 \\
Poly Haven occluders        & 10.80 & 11.31 & 15.59 & 11.18 & 1.82 & 5.54 & 11.49 & 11.61 \\
No shadergraph             &  9.50 & 10.41 & 12.66 & 10.45 & 0.99 & 5.75 & 10.91 &  8.51 \\
SMAA                       &  7.70 &  8.56 & 12.24 &  9.67 & 1.17 & 5.86 & 10.12 &  9.51 \\
Simple anims               &  8.70 &  9.27 & 15.64 & 10.31 & 0.25 & 5.81 & 11.89 & 11.49 \\
\textbf{PSP-HDRI$+$}                & \textbf{12.80} & \textbf{13.07} & 15.67 & \underline{\textbf{13.57}} & 0.72 & \textbf{8.09} & 11.07 & 13.97 \\ 
PSP-HDRI$+$ w/ random crop & 12.70 & 12.78 & 15.42 & 13.43 & 0.27 & 7.24 & \underline{\textbf{11.90}} & \underline{\textbf{15.66}} \\ 
\midrule
MOTSynth                  & 7.30 & 7.72 & \textbf{26.32} & \textbf{20.74} & 0.24 & 1.95 & \textbf{41.01} & \textbf{32.75} \\
\bottomrule
\end{tabular}
\end{adjustbox}
\label{tab:ablation_ood}
\end{table*}

%% file: tables/tables_transfer_psphdriplus_motsynth_to_coco_mpii.tex
\begin{table*}[htb!]
\centering
\caption{\textbf{Keypoint AP for in-distribution and OOD sets for transfer learning from PSP-HDRI$+$ and MOTSynth to COCO and MPII datasets.} All models are trained with $49\times10^3$ images and fine-tuned on the full fine-tuning dataset.}
\centering
\begin{adjustbox}{max width=0.8\textwidth, center}
\begin{tabular}{rrr|r|r|r|r|r|r|r|r} 
\hline
\noalign{\smallskip}
pre-train    & $\rightarrow$ &  fine-tune & COCO test-dev2017 & COCO person-val2017 & MPII val & Crowdpose Trainval & Leeds Sports & Occluded Humans & MOTSynth & MOT17 (bbox AP)  \\
\noalign{\smallskip}
\hline
\noalign{\smallskip}
\textbf{PSP-HDRI$+$} & $\rightarrow$ &  COCO & \textbf{62.80} & \textbf{66.33} & \textbf{70.33} & \textbf{70.07} & \textbf{27.45} & \textbf{31.84} & 16.15 &  32.01  \\
MOTSynth     & $\rightarrow$ &  COCO & 62.60 & 65.81 & 70.07 & 69.85 & 26.09 & 30.56 & \textbf{16.53} & \textbf{32.46} \\
\noalign{\smallskip}
\hline
\noalign{\smallskip}
\textbf{PSP-HDRI$+$} & $\rightarrow$ &  MPII & \textbf{17.30} & \textbf{16.29} & \textbf{72.55} & \textbf{50.12} & \textbf{33.78} & \textbf{10.53} & \textbf{7.97} & \textbf{12.03}   \\
MOTSynth     & $\rightarrow$ &  MPII & 14.30 & 13.54 & 71.21 & 47.90 & 30.17 & 8.13 & 7.46 & 11.06  \\
\noalign{\smallskip}
\hline
\end{tabular}
\end{adjustbox}
\label{tab:scratch_results_ood_mpii}
\end{table*}

%% file: 07_Conclusions.tex
\section{Conclusions}
We introduced a new synthetic data generator PSP-HDRI$+$ and showed that for human-centric computer vision, it provides a superior pre-training data compared with other common alternatives. We have also identified a training strategy whereby we obtained the best pre-training and fine-tuning results without the need for hyper-parameter search. This is made possible by automatic learning-rate annealing that adapts to the model performance on a validation set and is scaled based on the size of training data. We demonstrated that our open-source, privacy-preserving, ethically sourced, and fully manipulable human-centric synthetic data generator has the potential for improvement beyond its out-of-the-box capabilities and is an excellent choice for meta-learning and sim2real research. 

%% file: 08_Appendix.tex
\section{Appendix: More examples from PSP-HDRI$+$}
\setcounter{figure}{0} 
\setcounter{table}{0} 

\begin{figure}[htb!]
    \centering
    \begin{subfigure}[t]{0.132\textwidth}
        {\includegraphics[height=2.3cm]{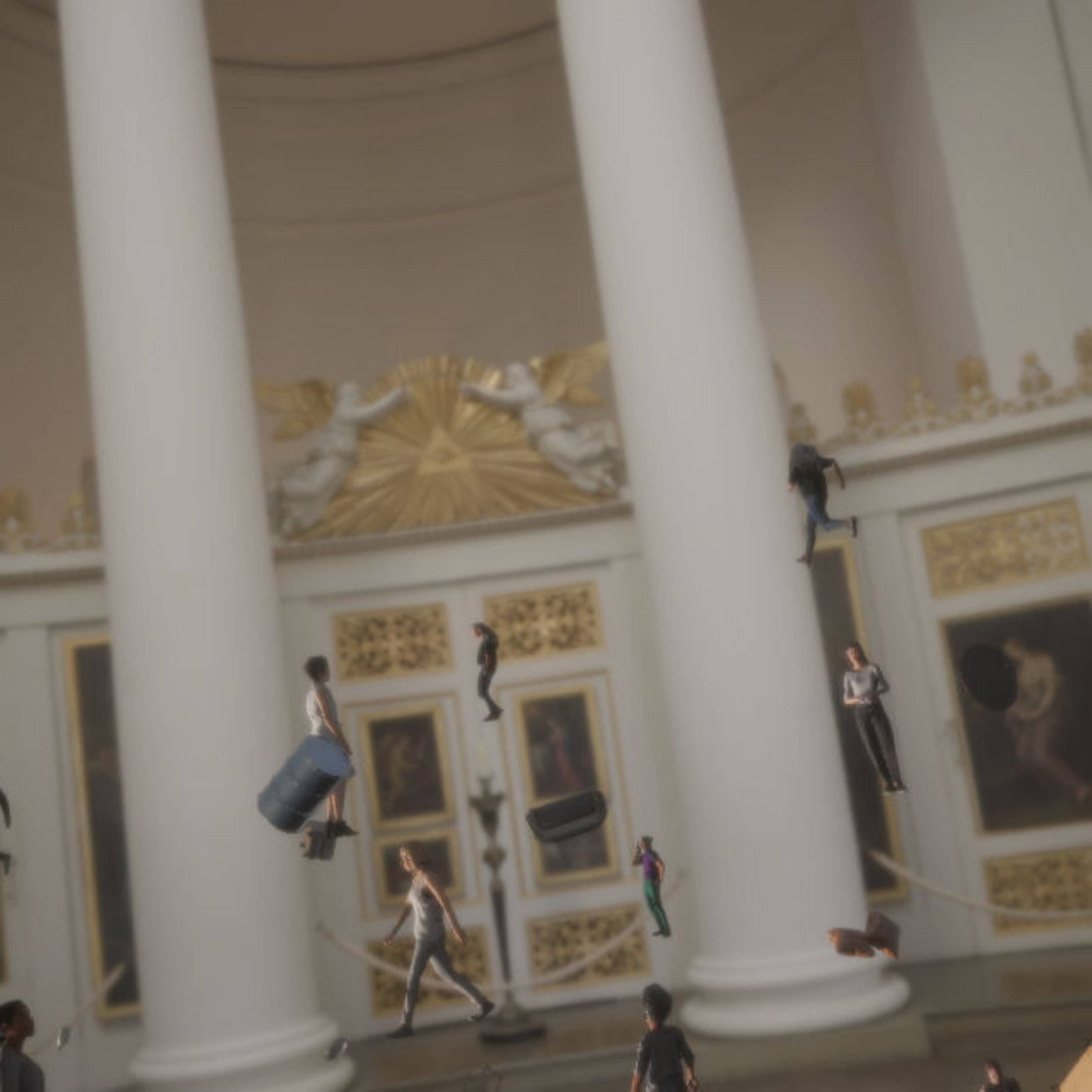}}
    \end{subfigure}
    \begin{subfigure}[t]{0.132\textwidth}
        {\includegraphics[height=2.3cm]{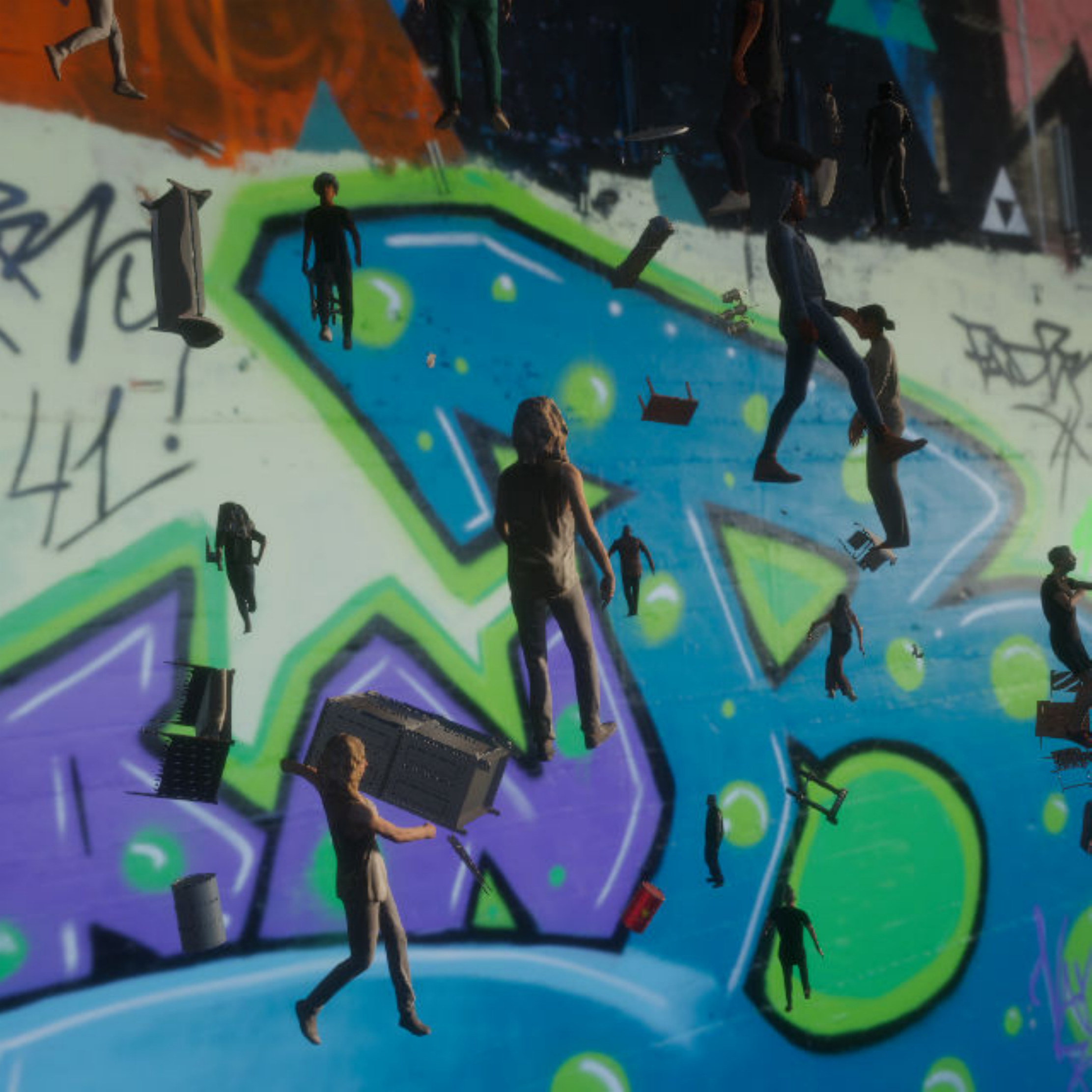}}
    \end{subfigure}
    \begin{subfigure}[t]{0.132\textwidth}
        {\includegraphics[height=2.3cm]{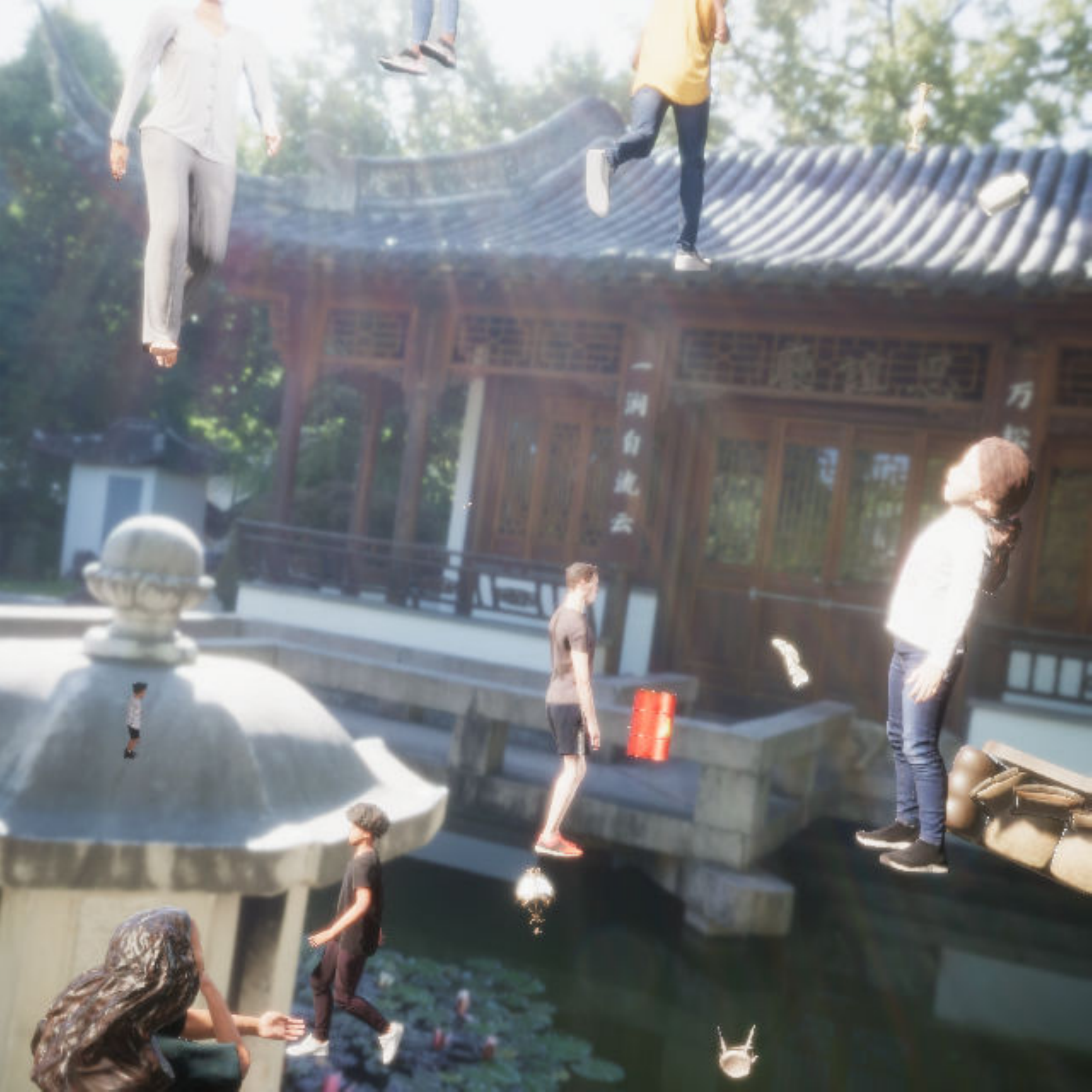}}
    \end{subfigure}
    \begin{subfigure}[t]{0.132\textwidth}
        {\includegraphics[height=2.3cm]{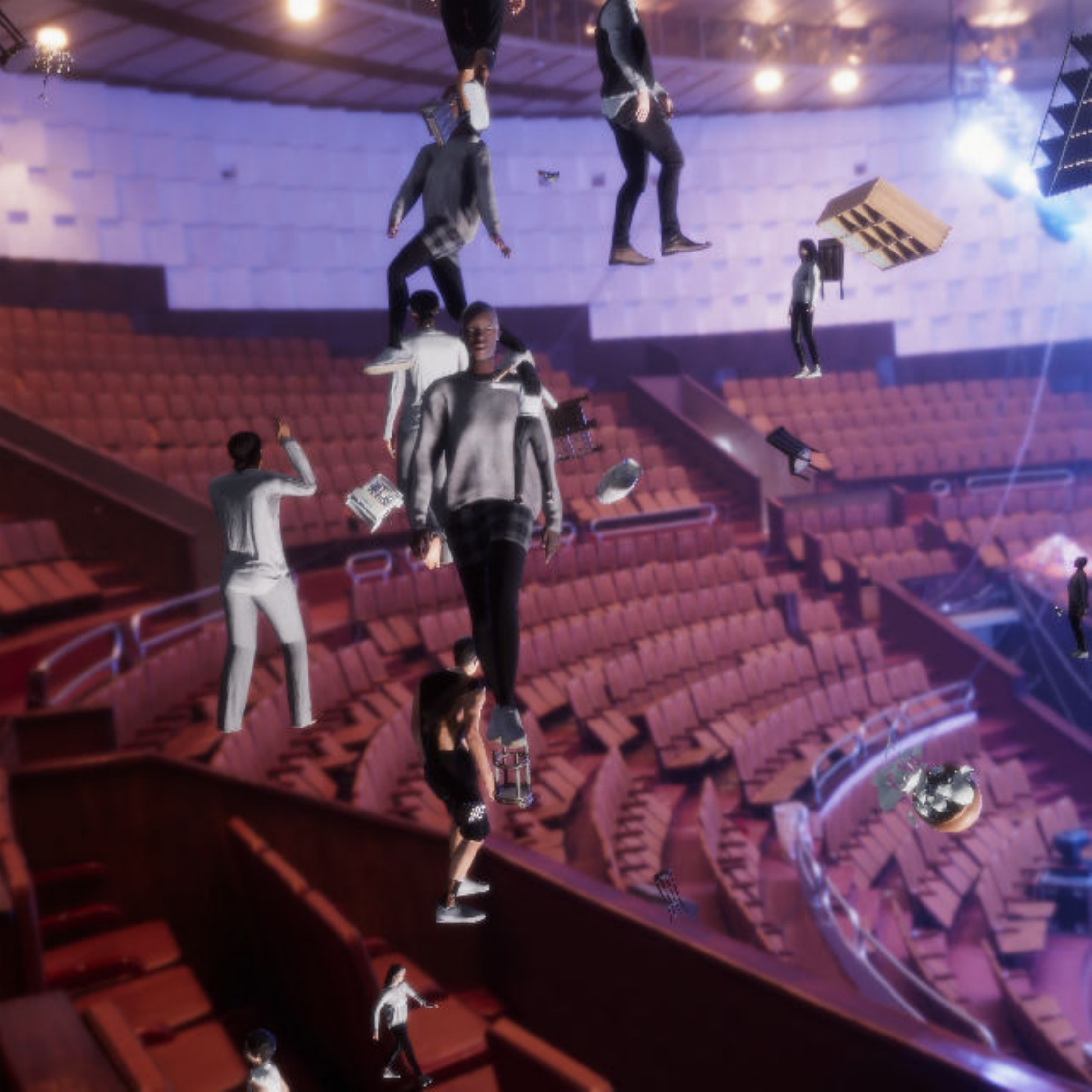}}
    \end{subfigure}
    \begin{subfigure}[t]{0.132\textwidth}
        {\includegraphics[height=2.3cm]{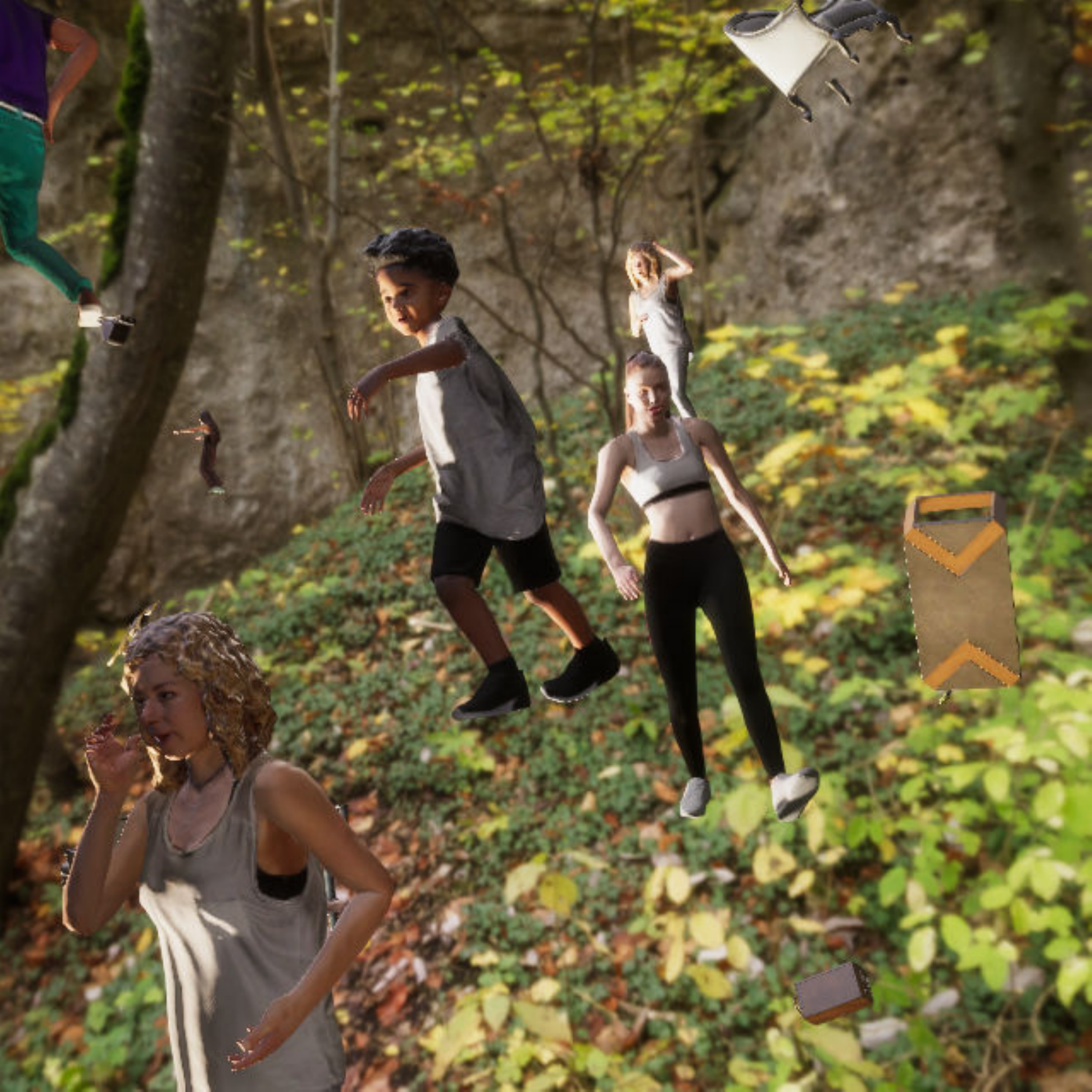}}
    \end{subfigure}
    \begin{subfigure}[t]{0.132\textwidth}
        {\includegraphics[height=2.3cm]{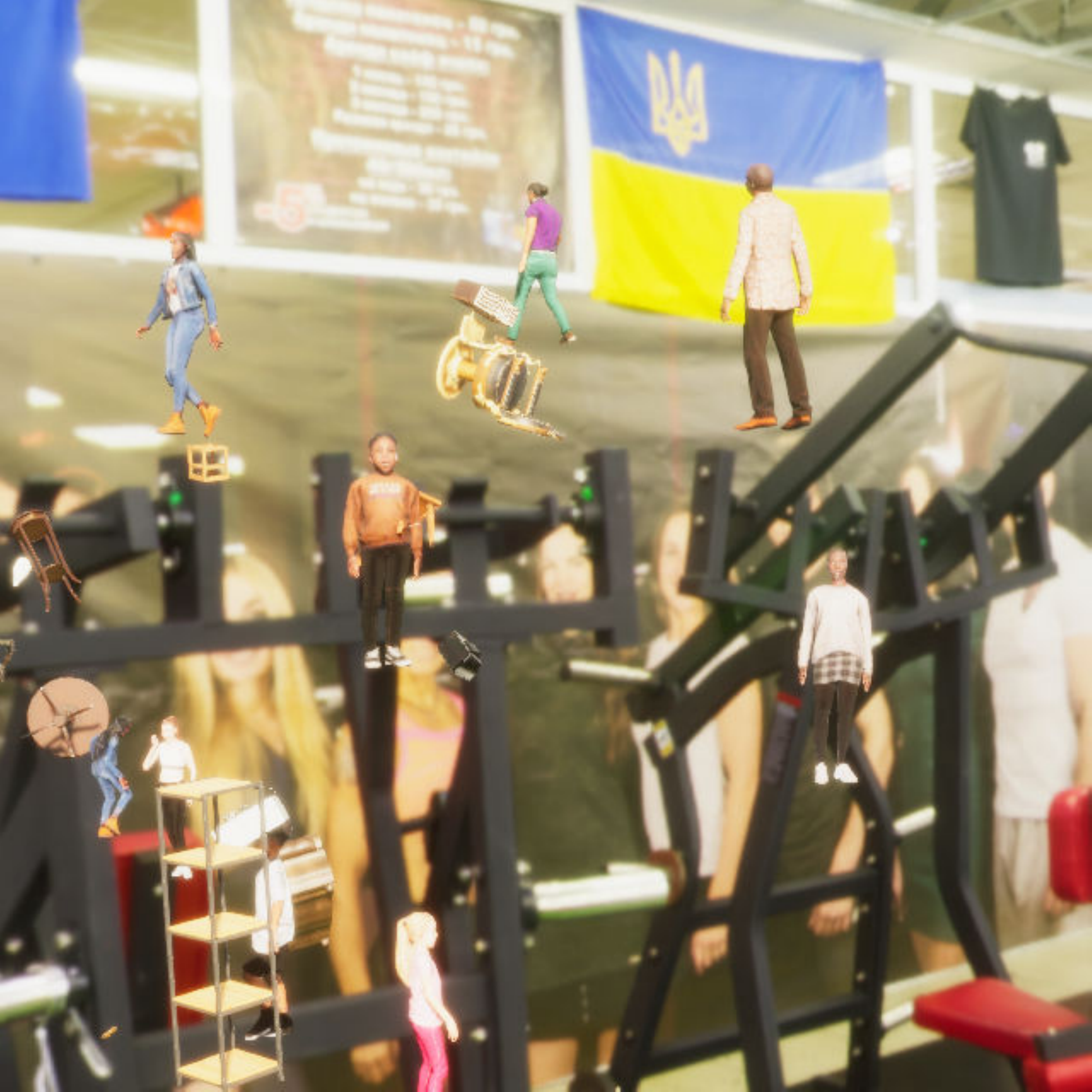}}
    \end{subfigure}
    \\
    \begin{subfigure}[t]{0.132\textwidth}
        {\includegraphics[height=2.3cm]{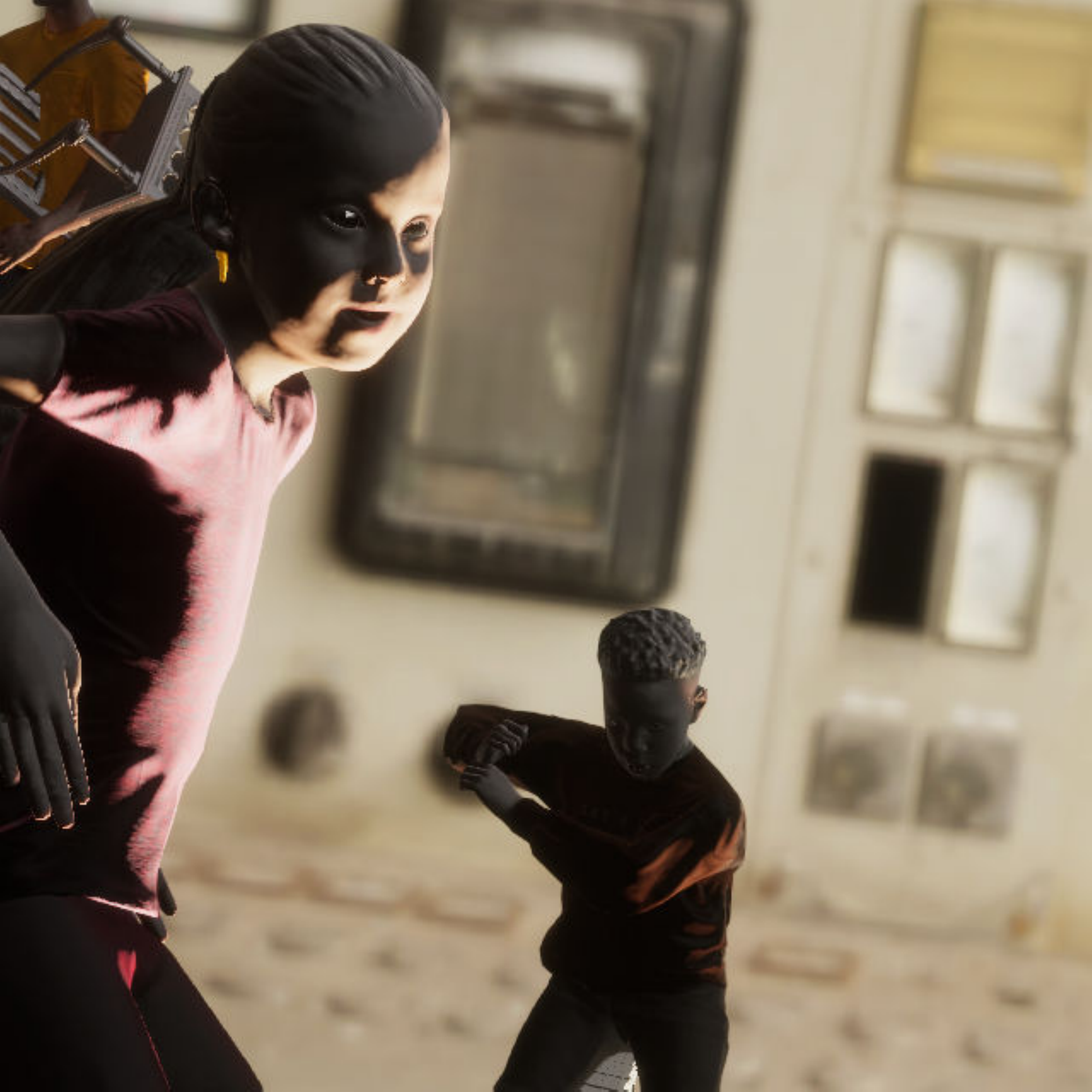}}
    \end{subfigure}
    \begin{subfigure}[t]{0.132\textwidth}
        {\includegraphics[height=2.3cm]{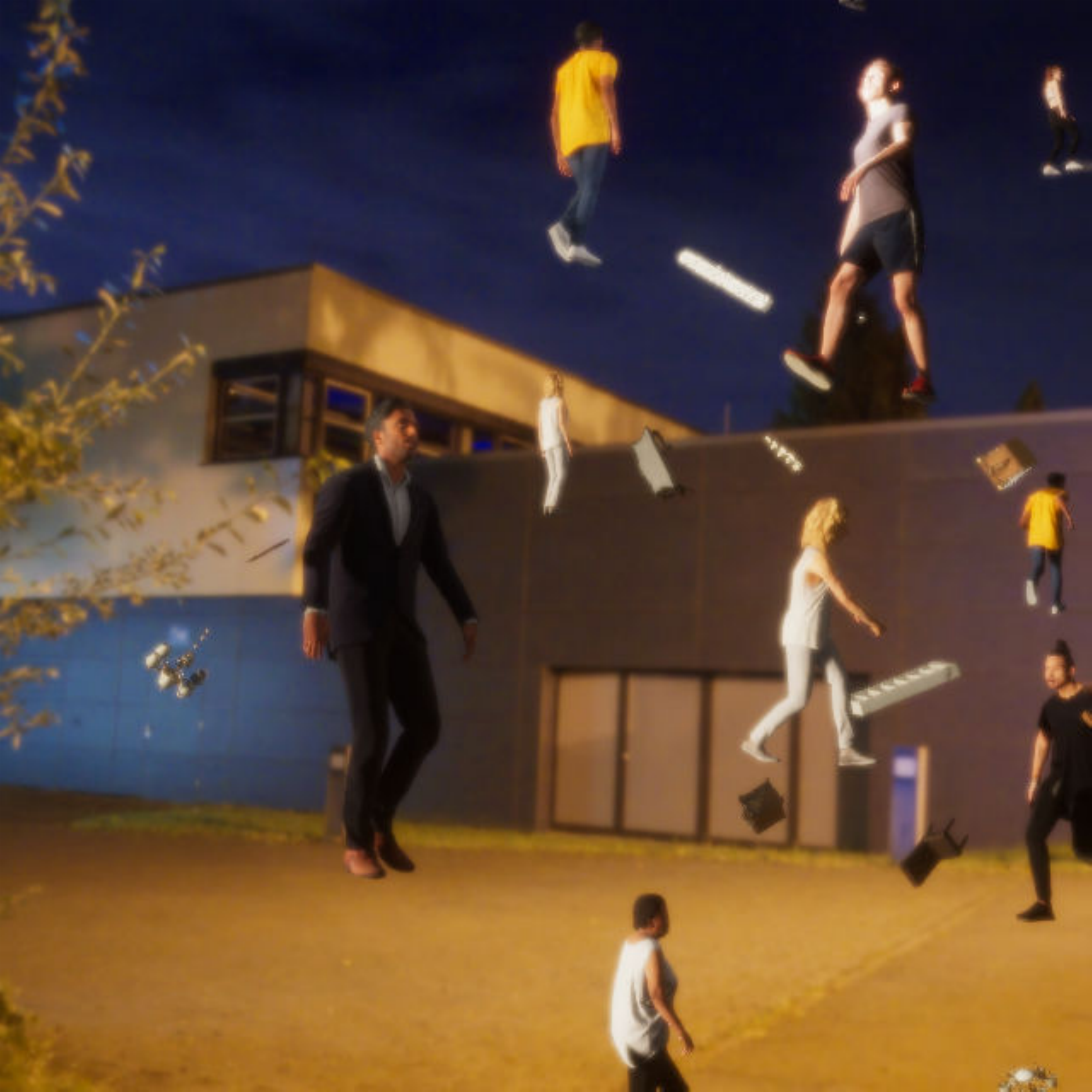}}
    \end{subfigure}
    \begin{subfigure}[t]{0.132\textwidth}
        {\includegraphics[height=2.3cm]{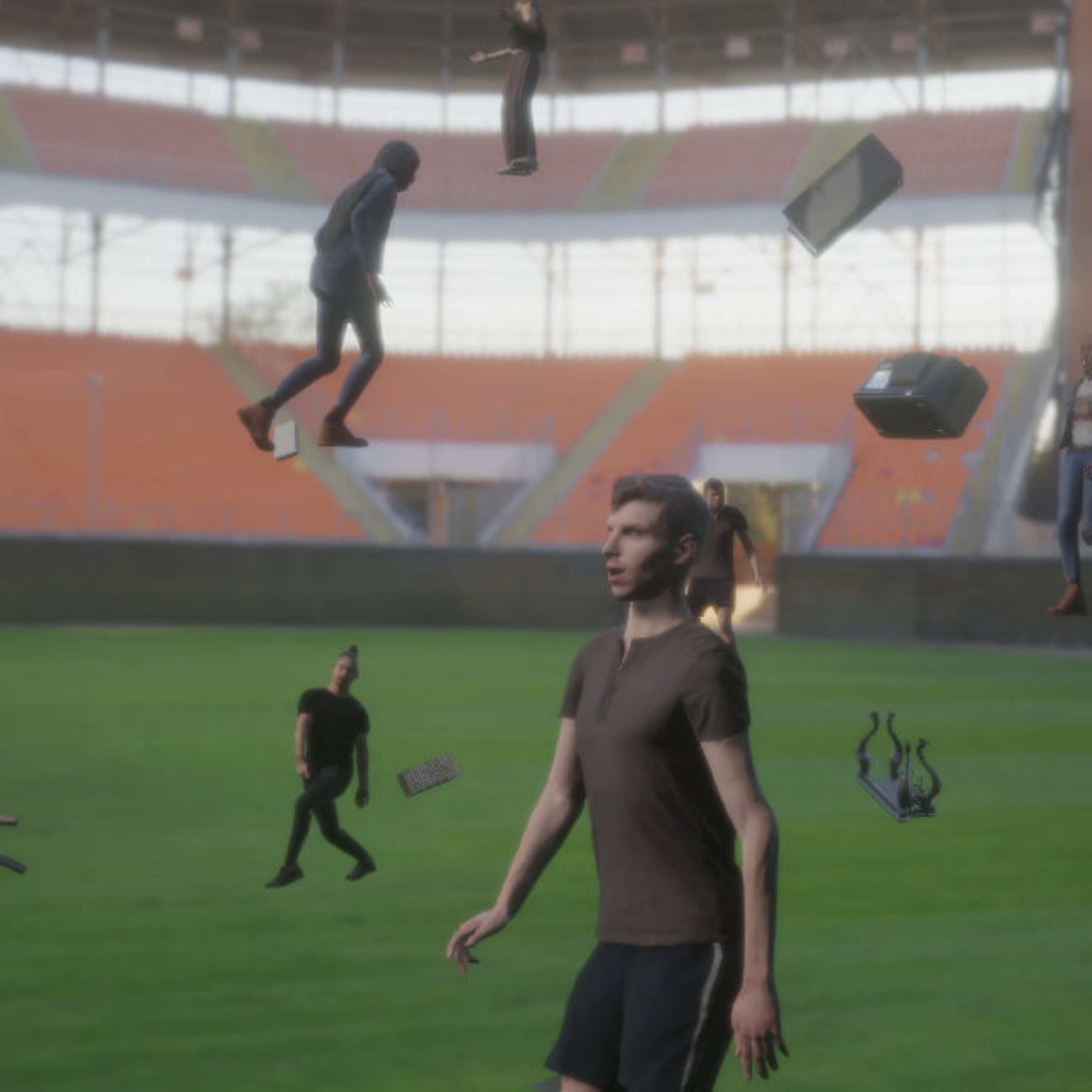}}
    \end{subfigure}
    \begin{subfigure}[t]{0.132\textwidth}
        {\includegraphics[height=2.3cm]{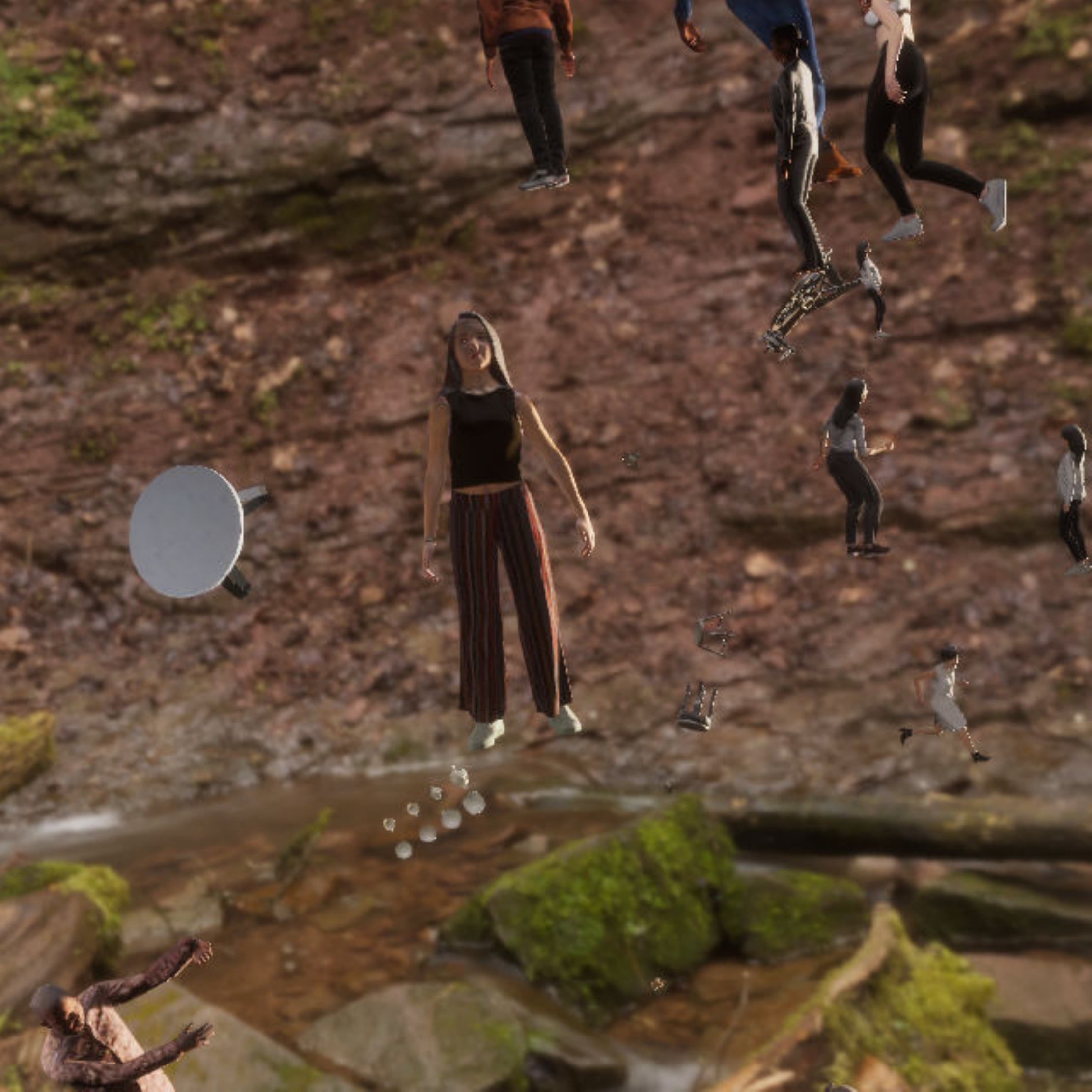}}
    \end{subfigure}
    \begin{subfigure}[t]{0.132\textwidth}
        {\includegraphics[height=2.3cm]{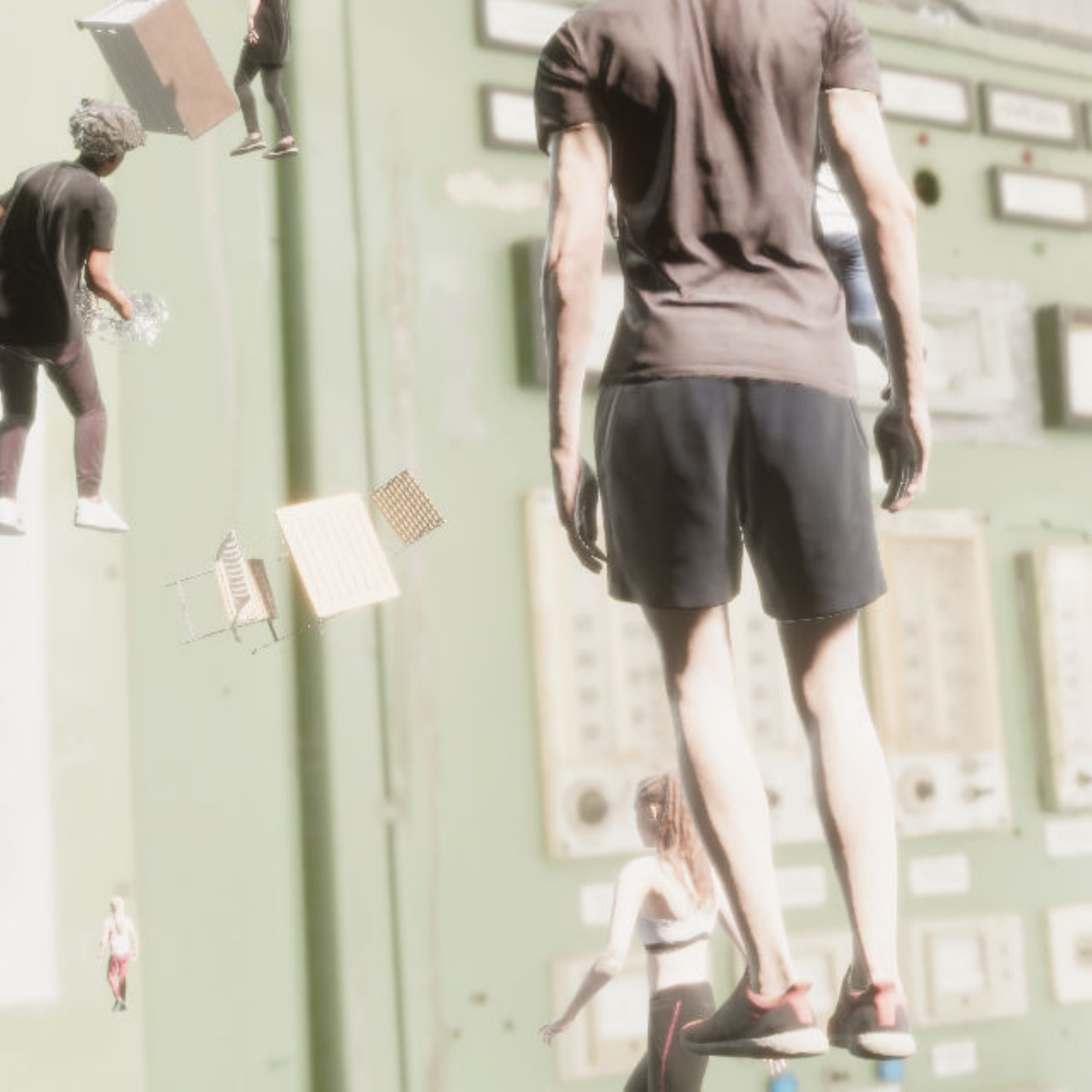}}
    \end{subfigure}
    \begin{subfigure}[t]{0.132\textwidth}
        {\includegraphics[height=2.3cm]{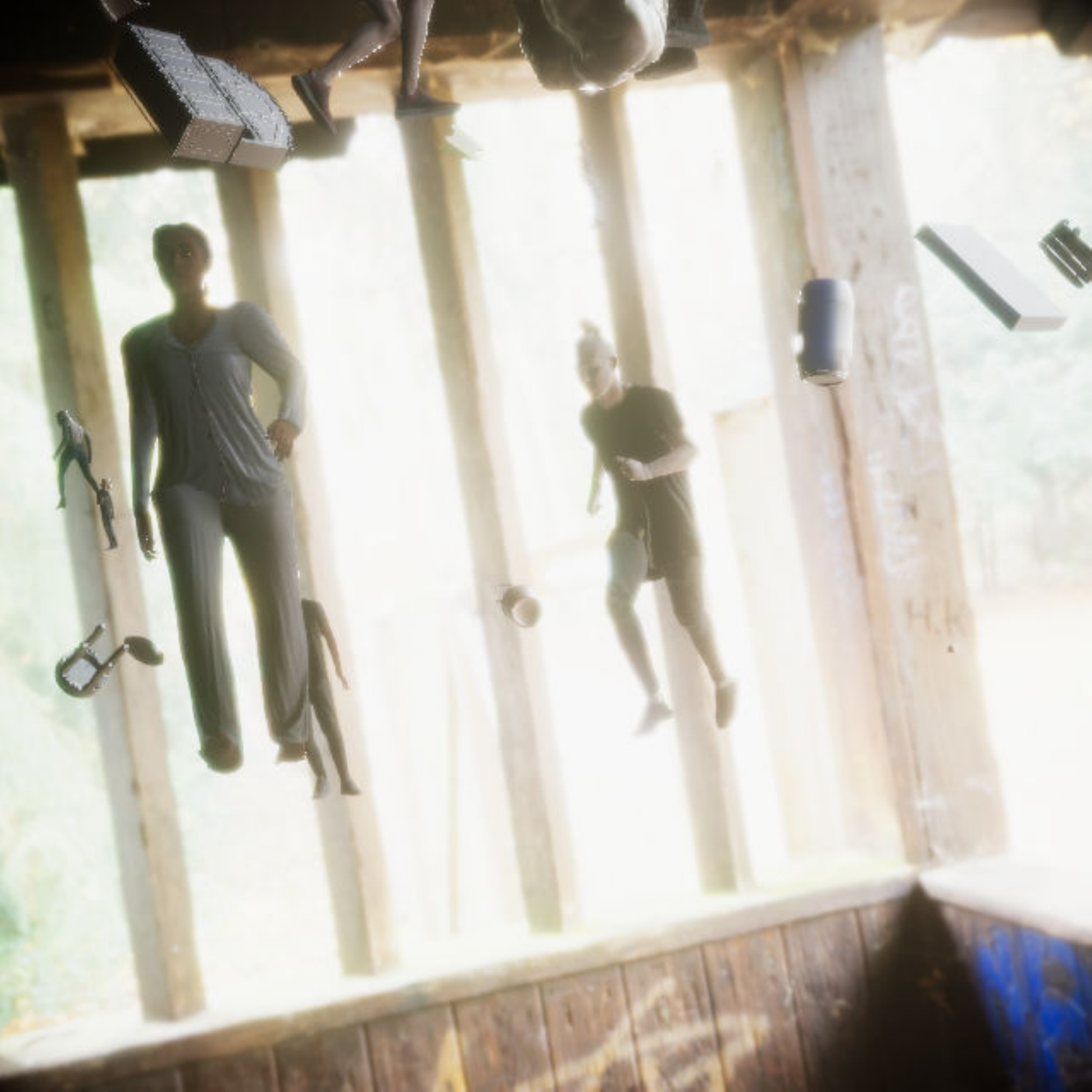}}
    \end{subfigure}
    \\
    \begin{subfigure}[t]{0.132\textwidth}
        {\includegraphics[height=2.3cm]{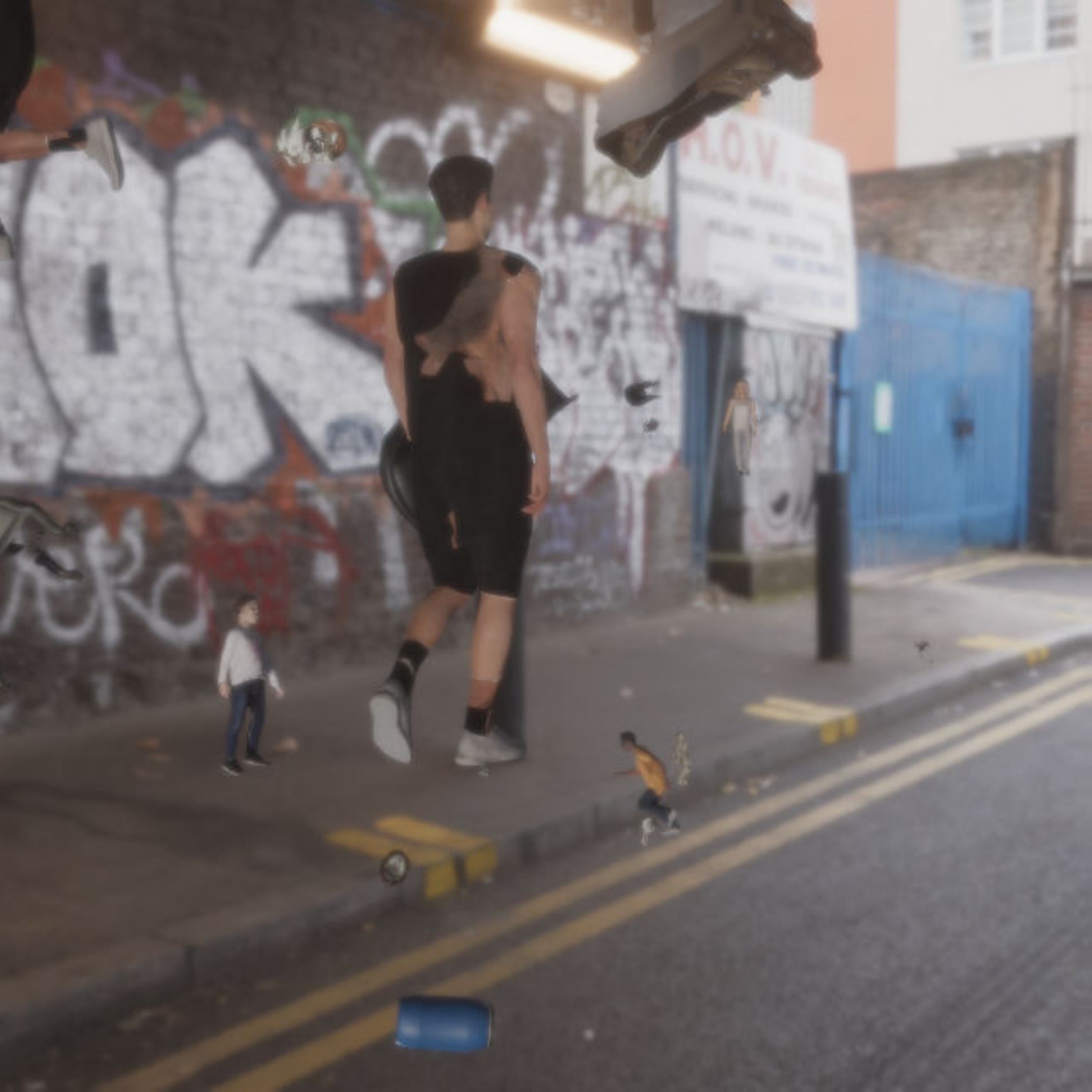}}
    \end{subfigure}
    \begin{subfigure}[t]{0.132\textwidth}
        {\includegraphics[height=2.3cm]{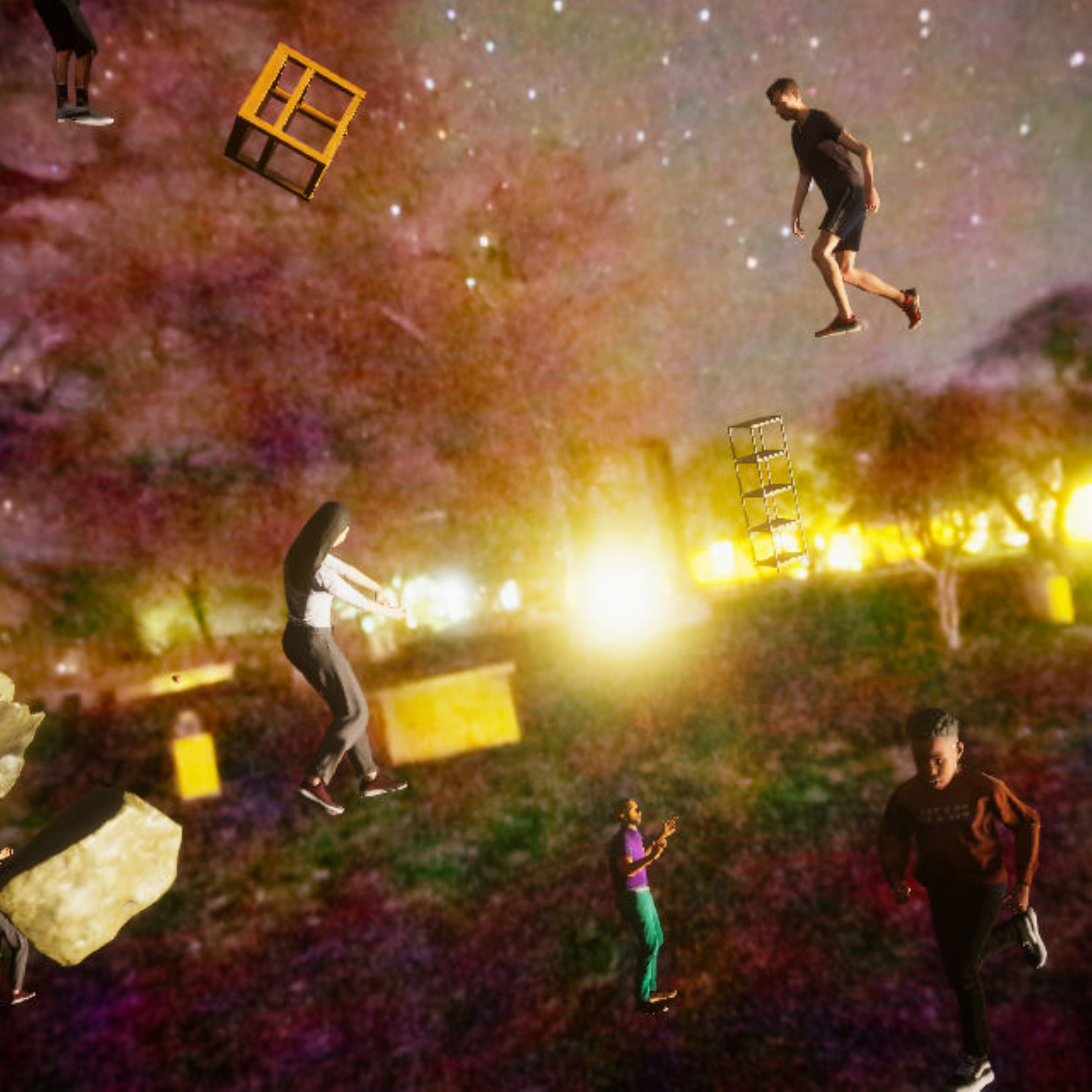}}
    \end{subfigure}
    \begin{subfigure}[t]{0.132\textwidth}
        {\includegraphics[height=2.3cm]{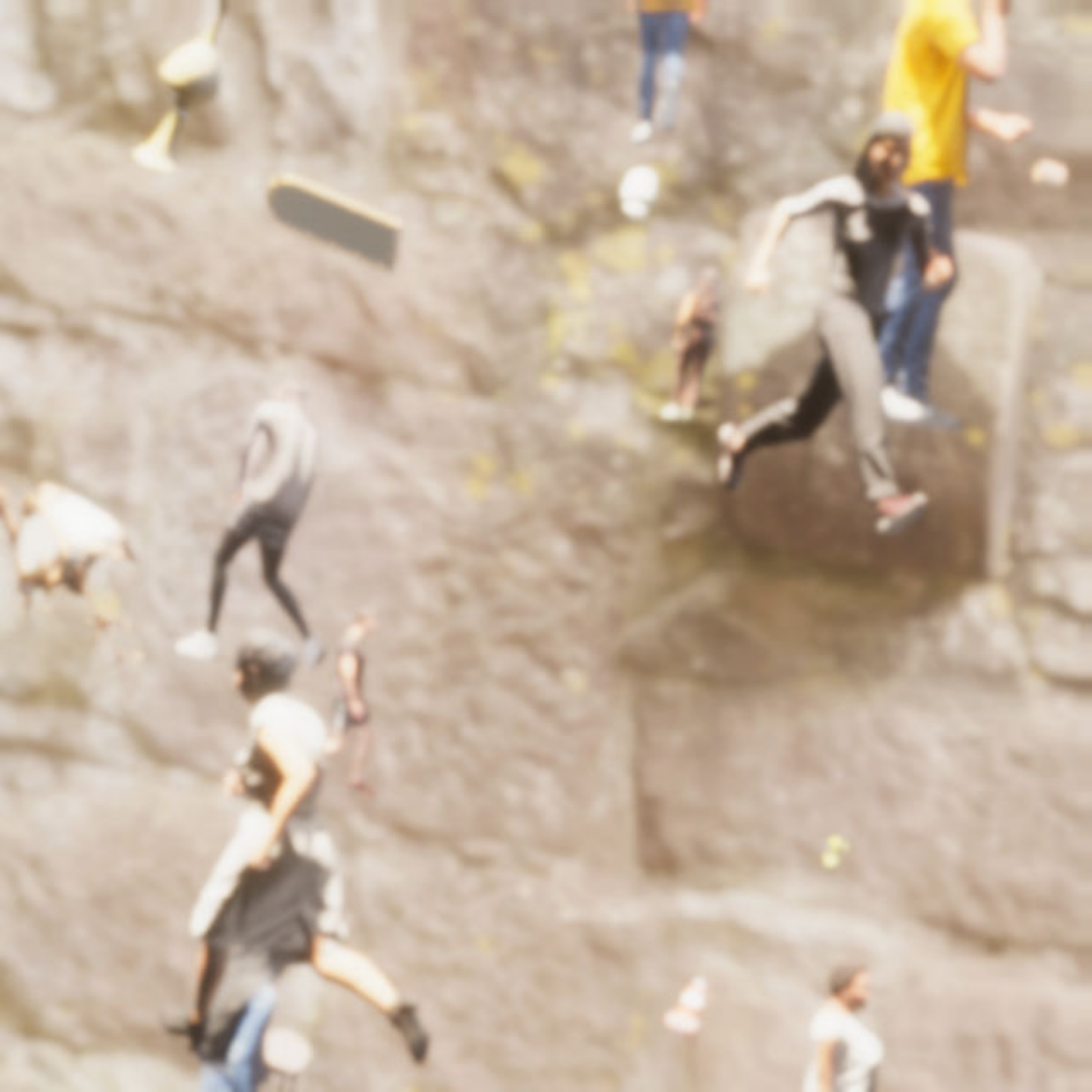}}
    \end{subfigure}
    \begin{subfigure}[t]{0.132\textwidth}
        {\includegraphics[height=2.3cm]{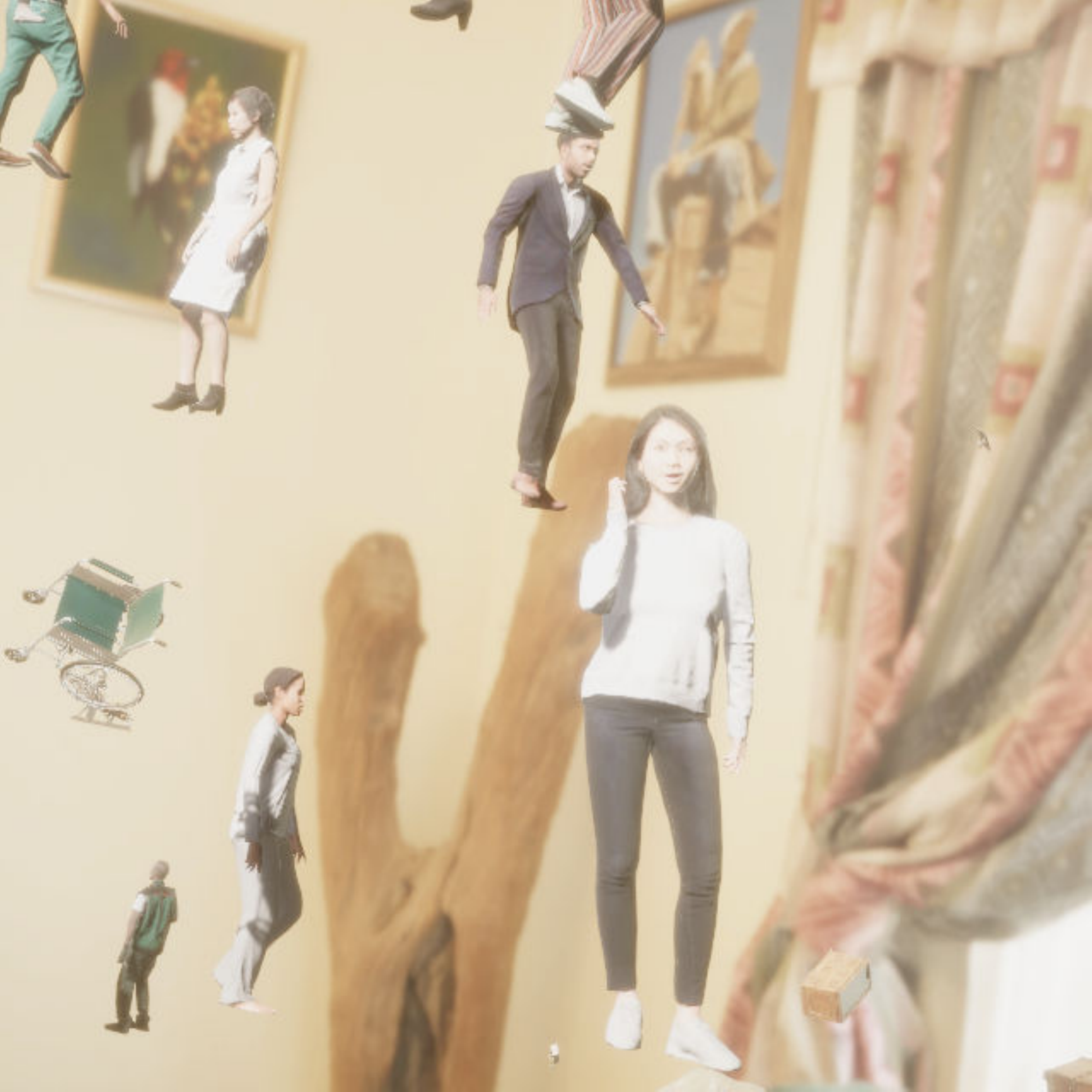}}
    \end{subfigure}
    \begin{subfigure}[t]{0.132\textwidth}
        {\includegraphics[height=2.3cm]{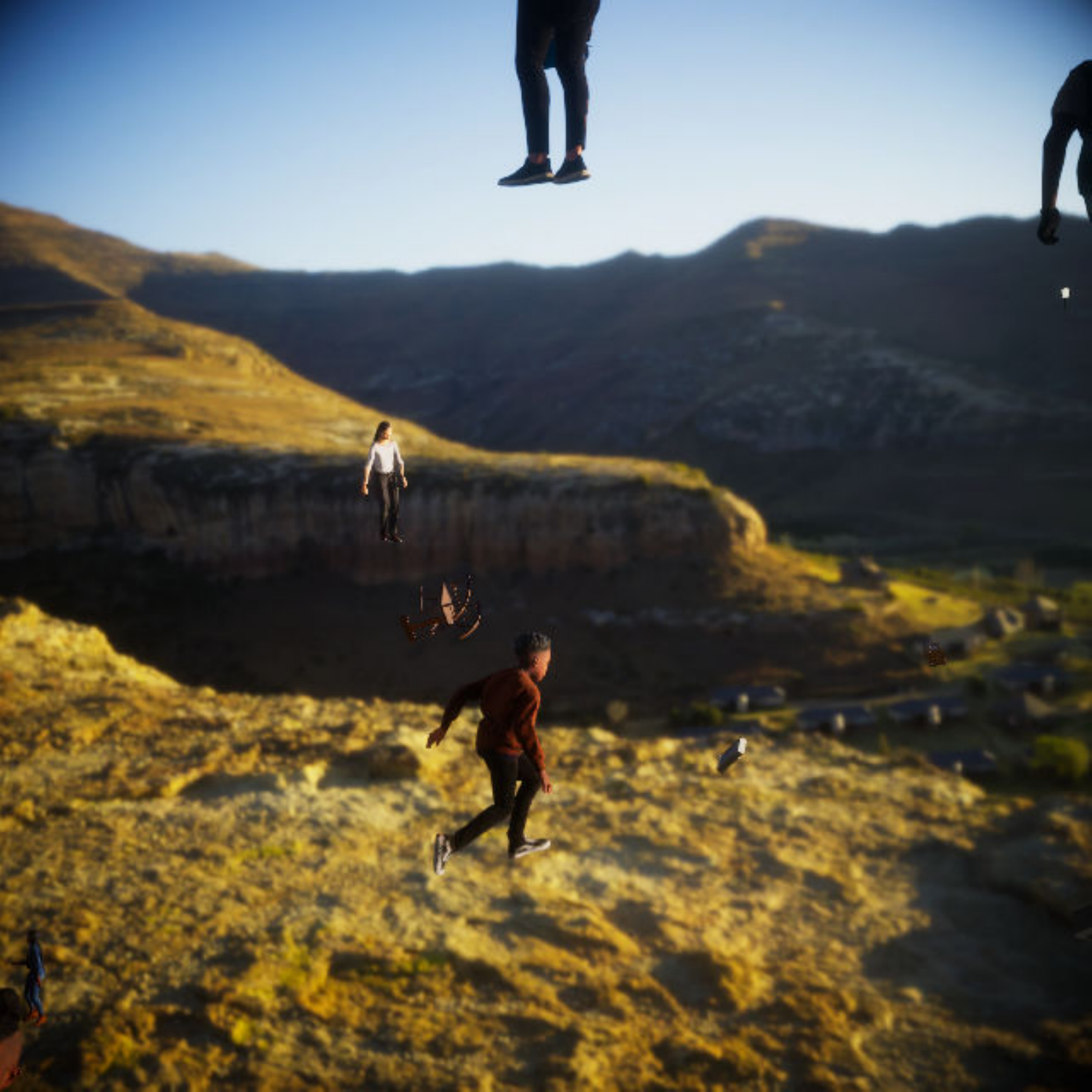}}
    \end{subfigure}
    \begin{subfigure}[t]{0.132\textwidth}
        {\includegraphics[height=2.3cm]{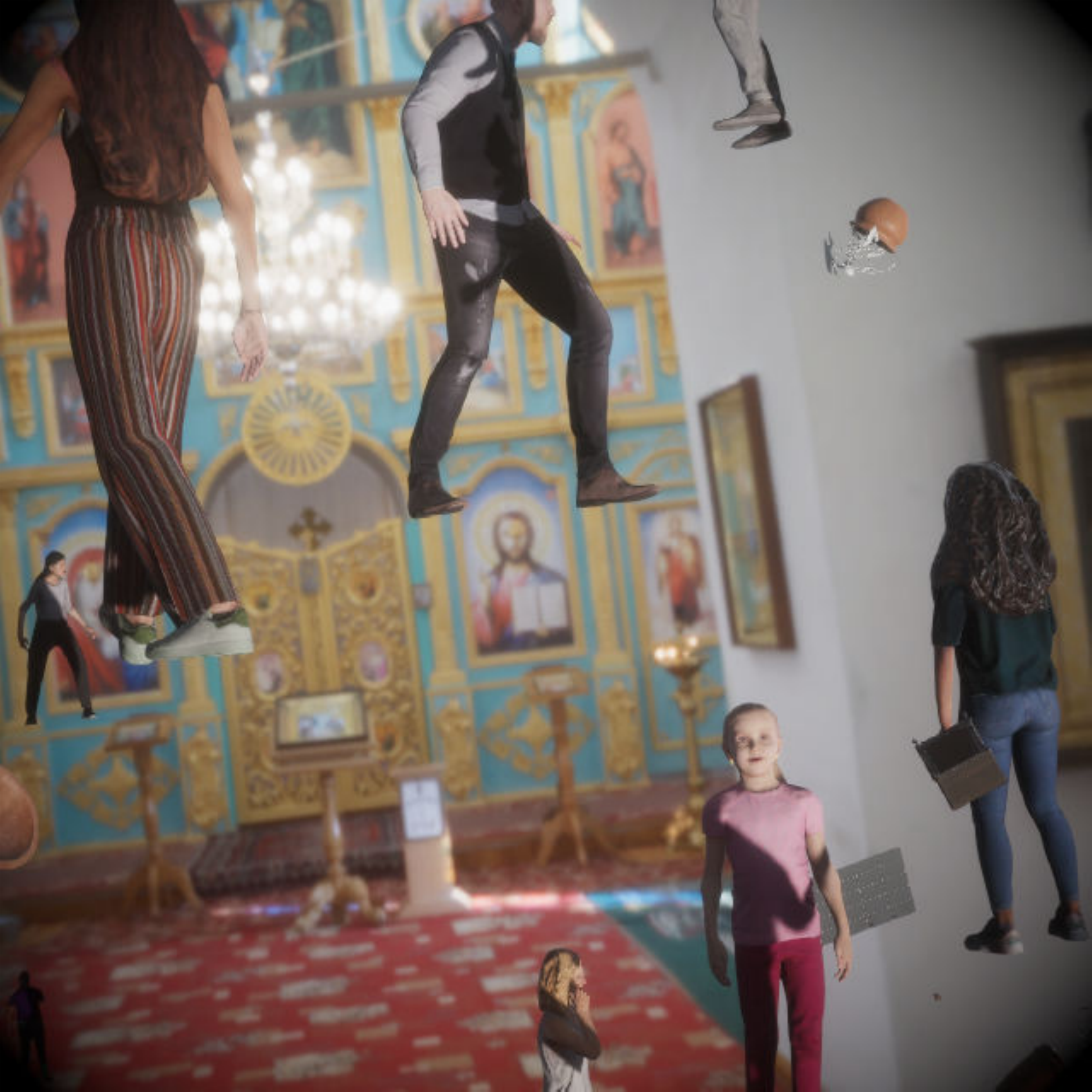}}
    \end{subfigure}
    \\
    \begin{subfigure}[t]{0.132\textwidth}
        {\includegraphics[height=2.3cm]{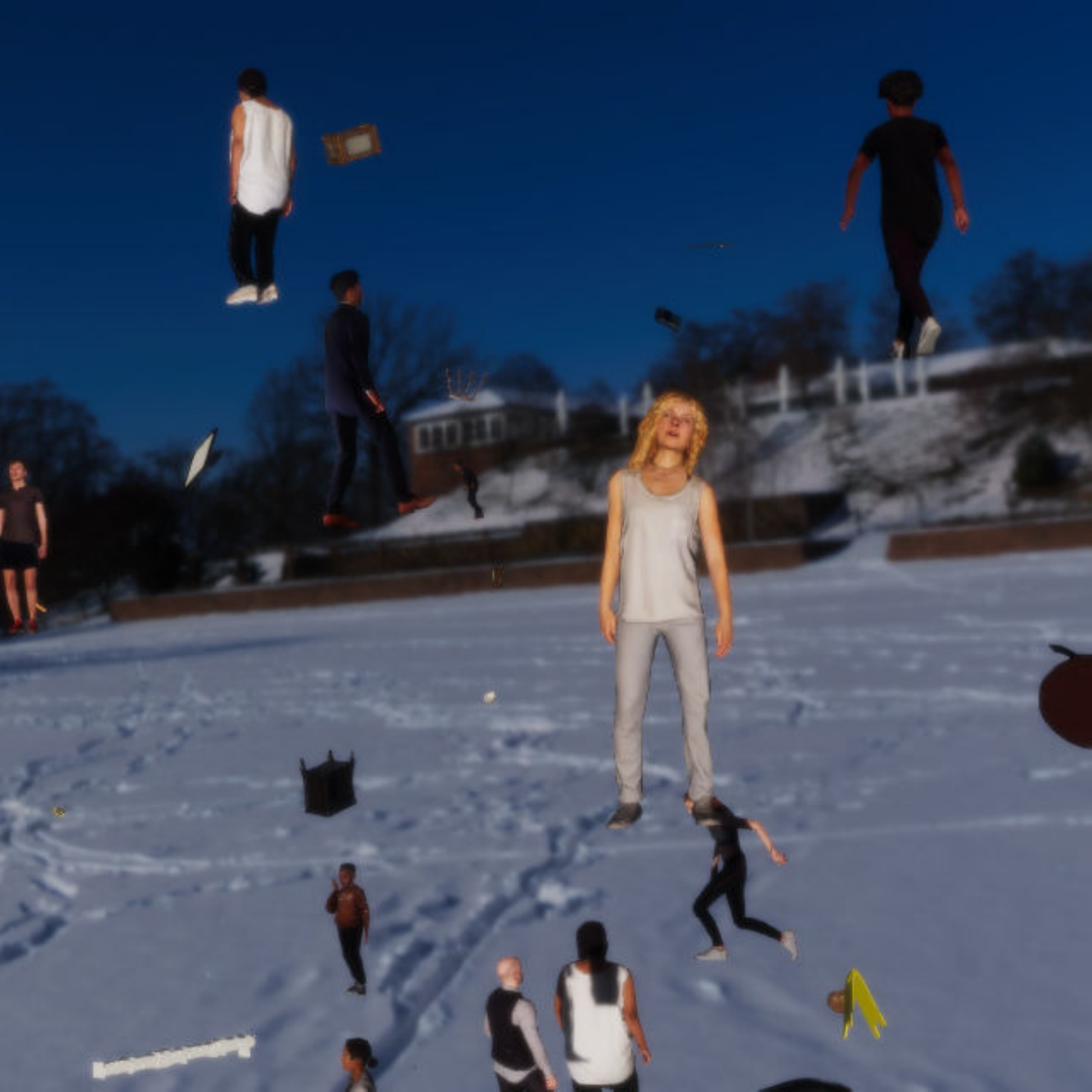}}
    \end{subfigure}
    \begin{subfigure}[t]{0.132\textwidth}
        {\includegraphics[height=2.3cm]{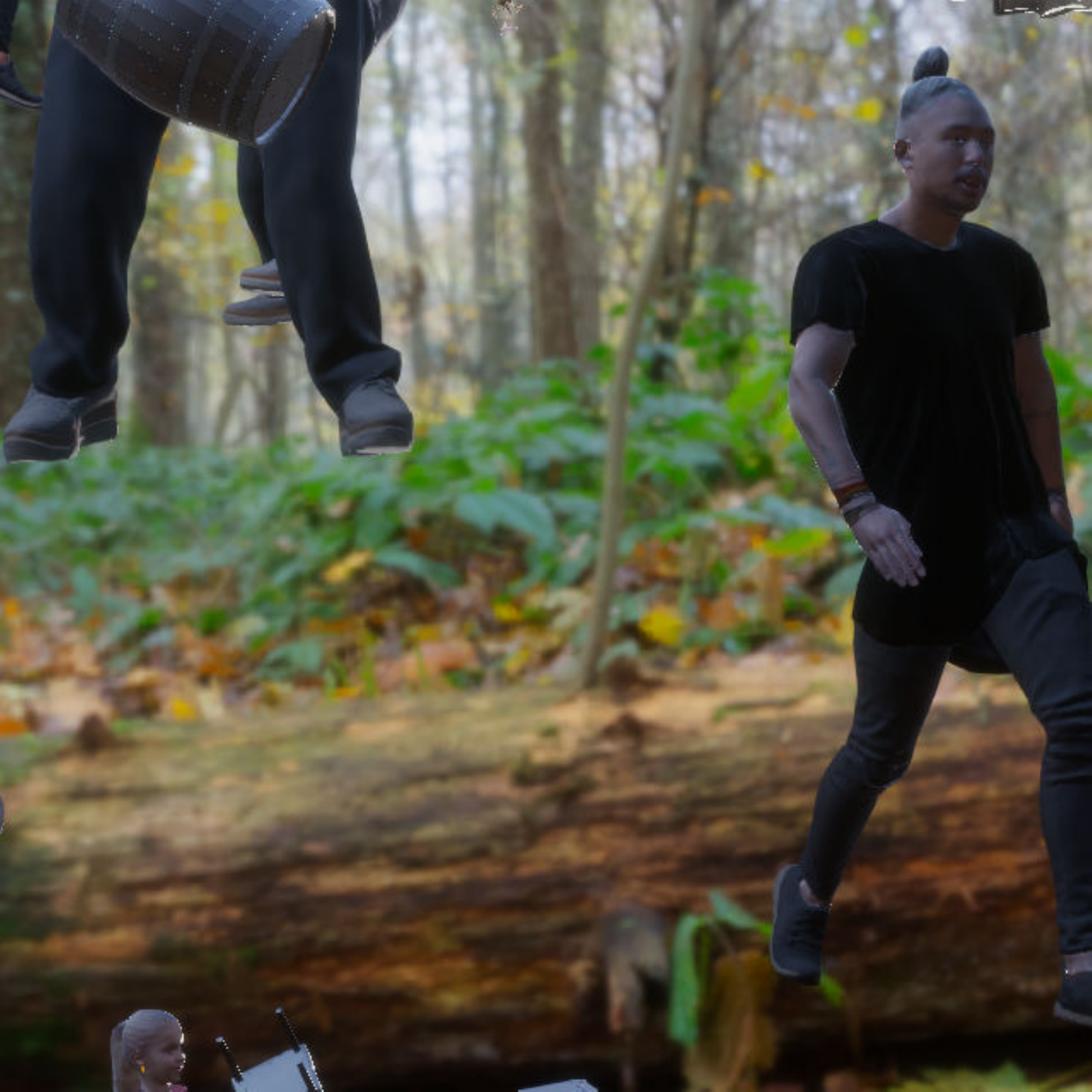}}
    \end{subfigure}
    \begin{subfigure}[t]{0.132\textwidth}
        {\includegraphics[height=2.3cm]{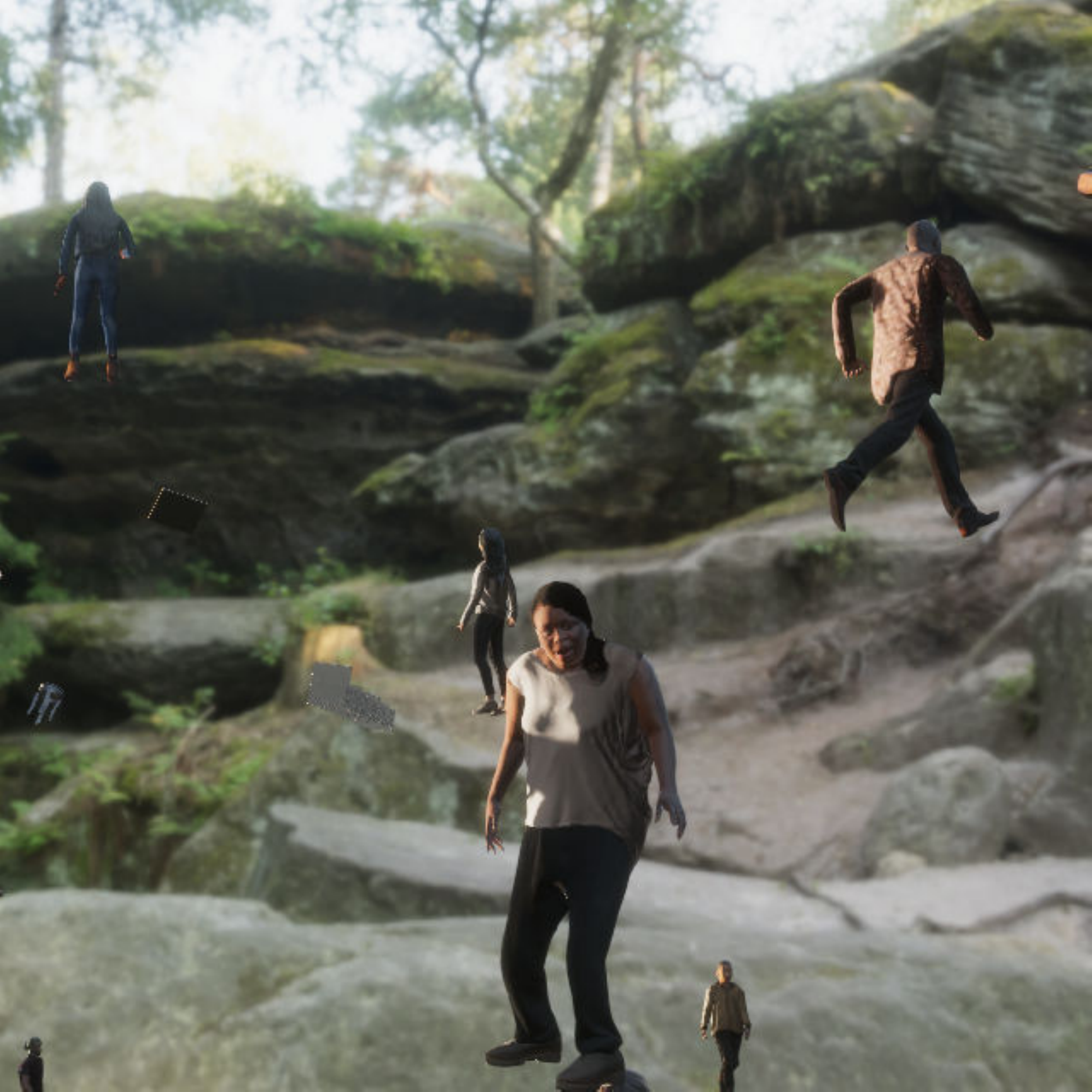}}
    \end{subfigure}
    \begin{subfigure}[t]{0.132\textwidth}
        {\includegraphics[height=2.3cm]{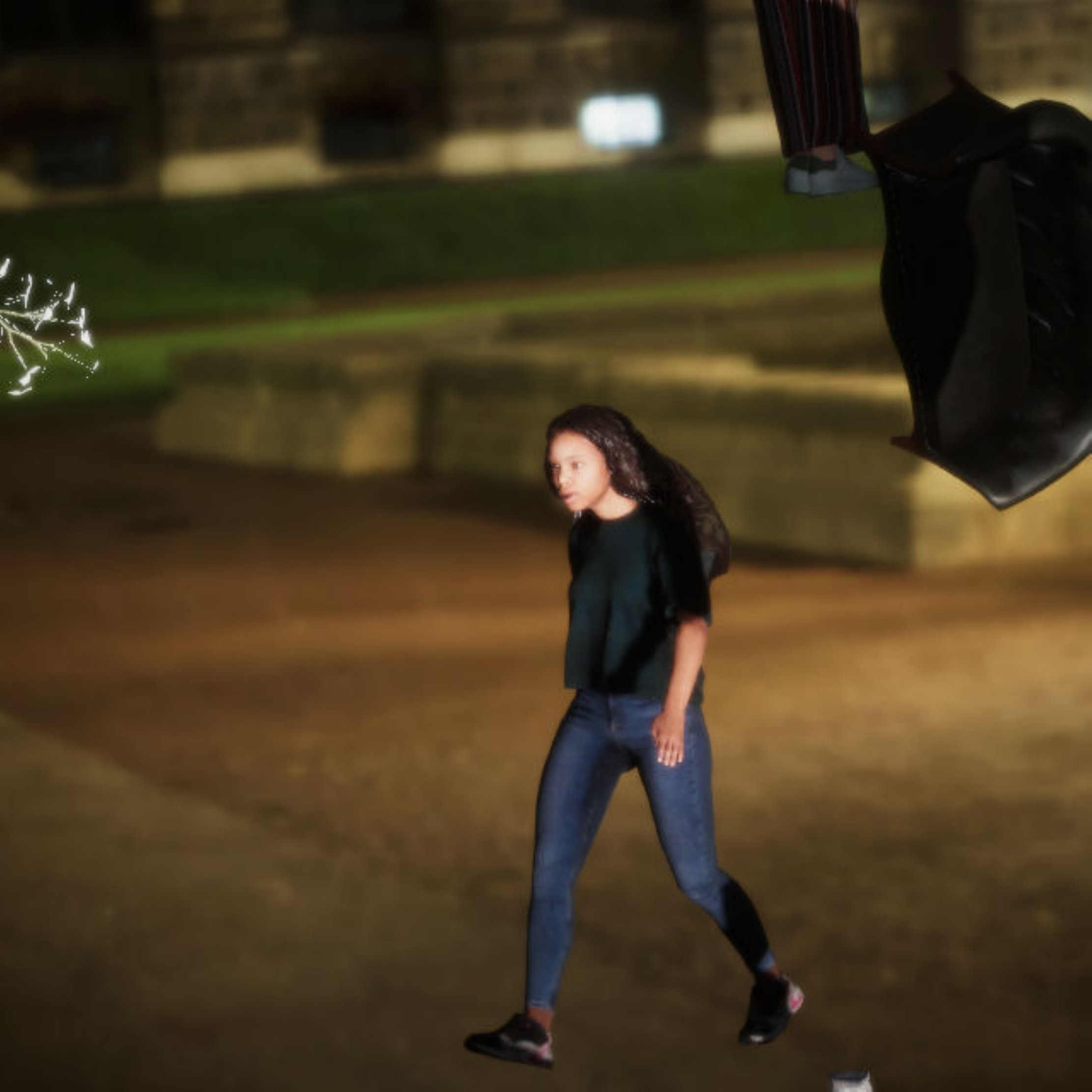}}
    \end{subfigure}
    \begin{subfigure}[t]{0.132\textwidth}
        {\includegraphics[height=2.3cm]{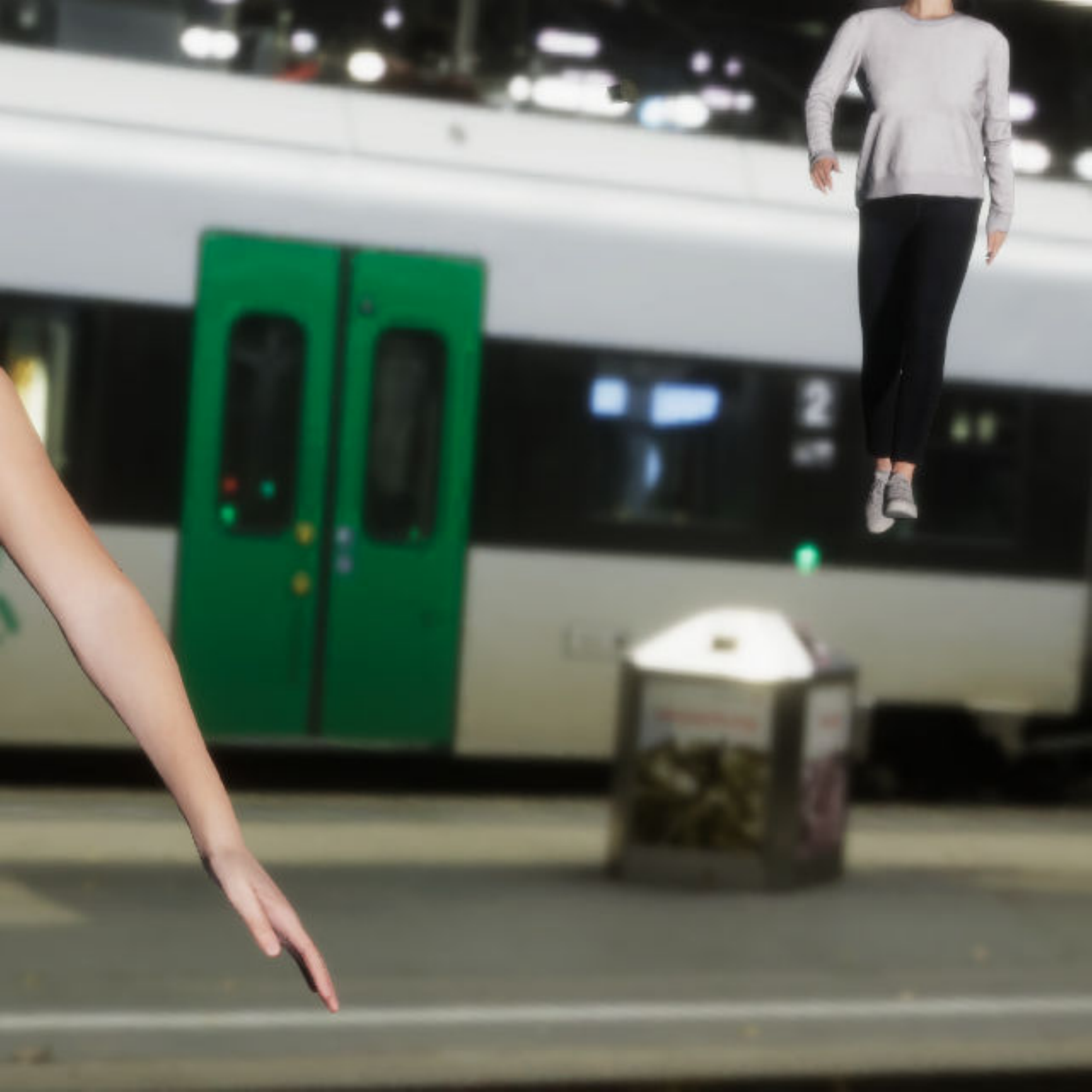}}
    \end{subfigure}
    \begin{subfigure}[t]{0.132\textwidth}
        {\includegraphics[height=2.3cm]{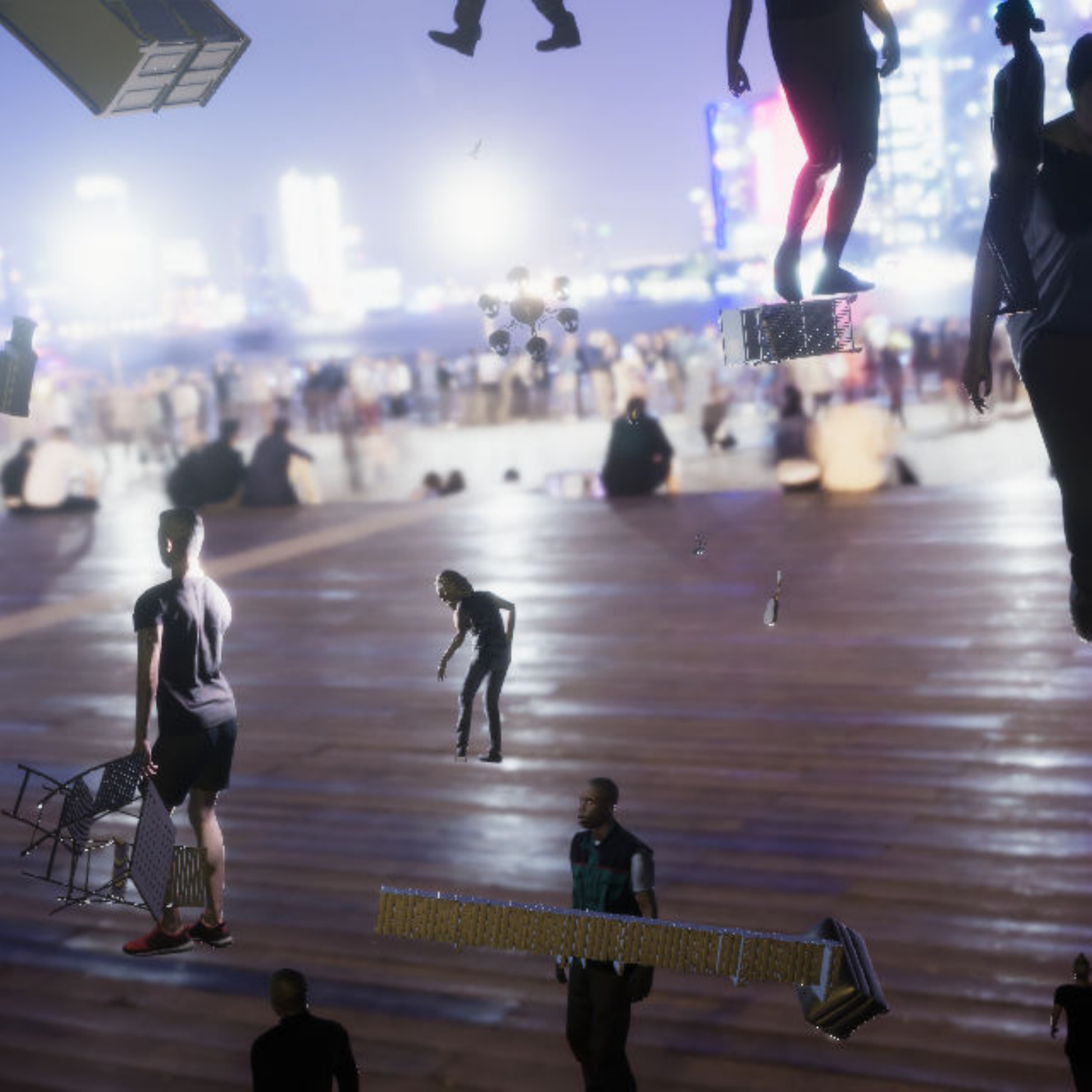}}
    \end{subfigure}
    \\
    \begin{subfigure}[t]{0.132\textwidth}
        {\includegraphics[height=2.3cm]{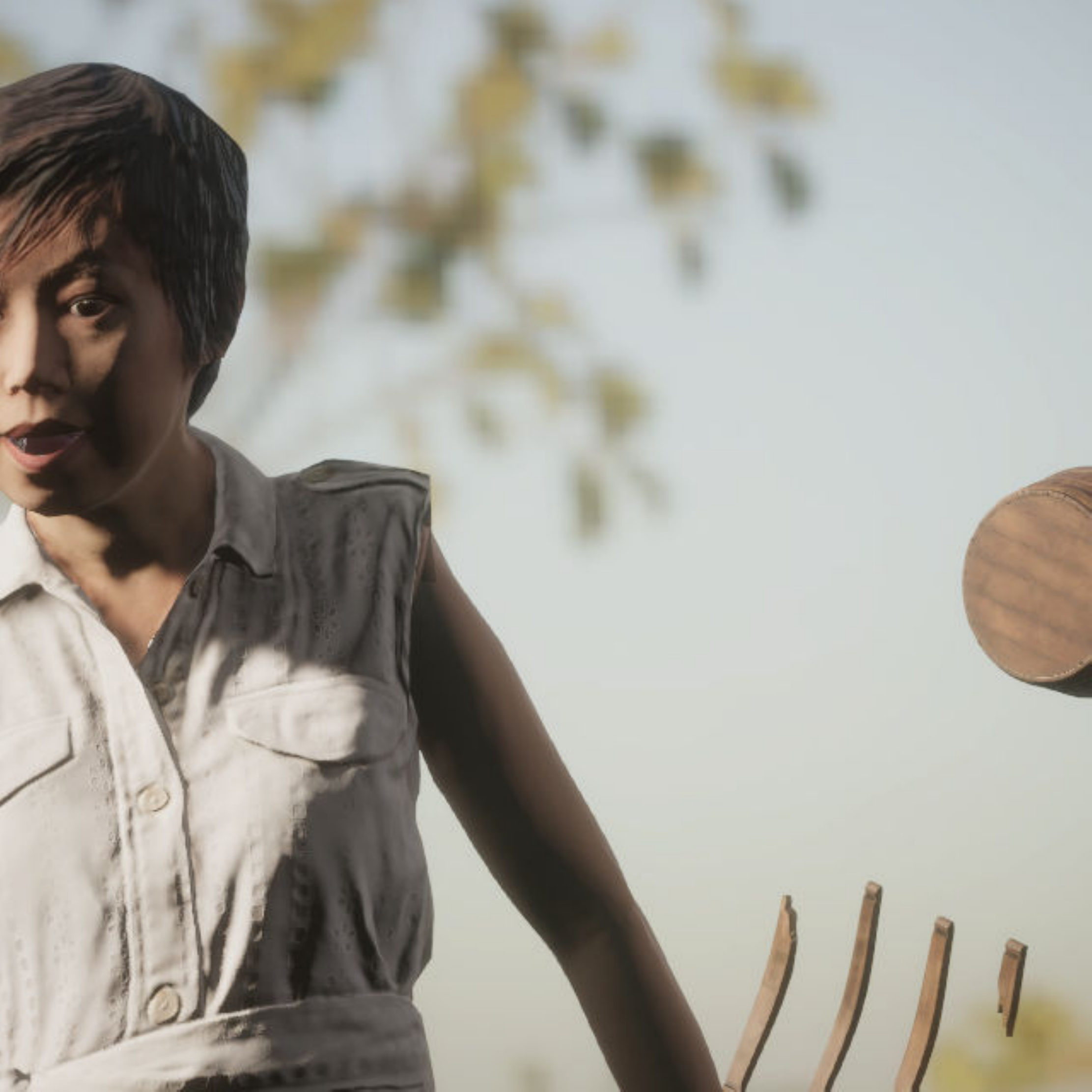}}
    \end{subfigure}
    \begin{subfigure}[t]{0.132\textwidth}
        {\includegraphics[height=2.3cm]{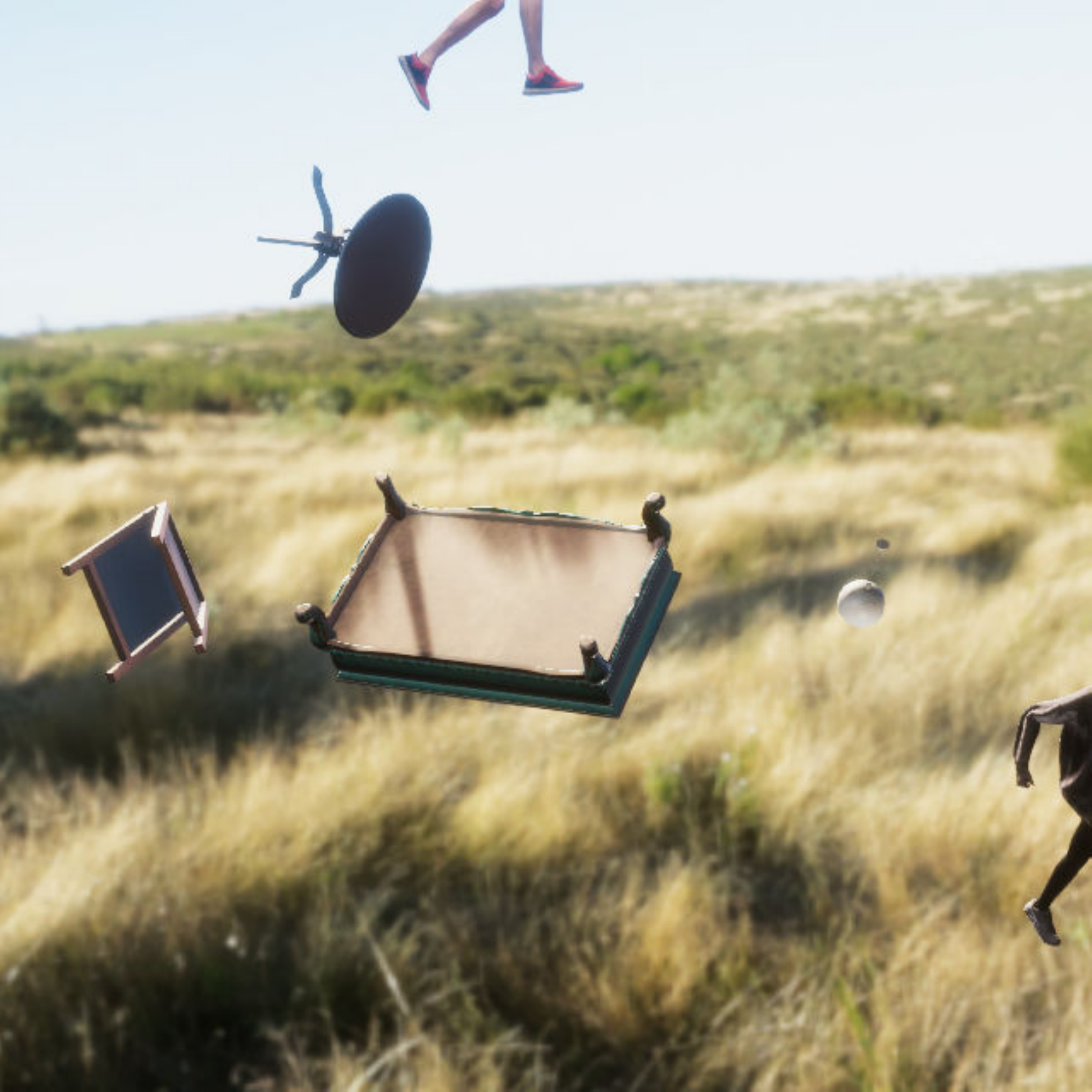}}
    \end{subfigure}
    \begin{subfigure}[t]{0.132\textwidth}
        {\includegraphics[height=2.3cm]{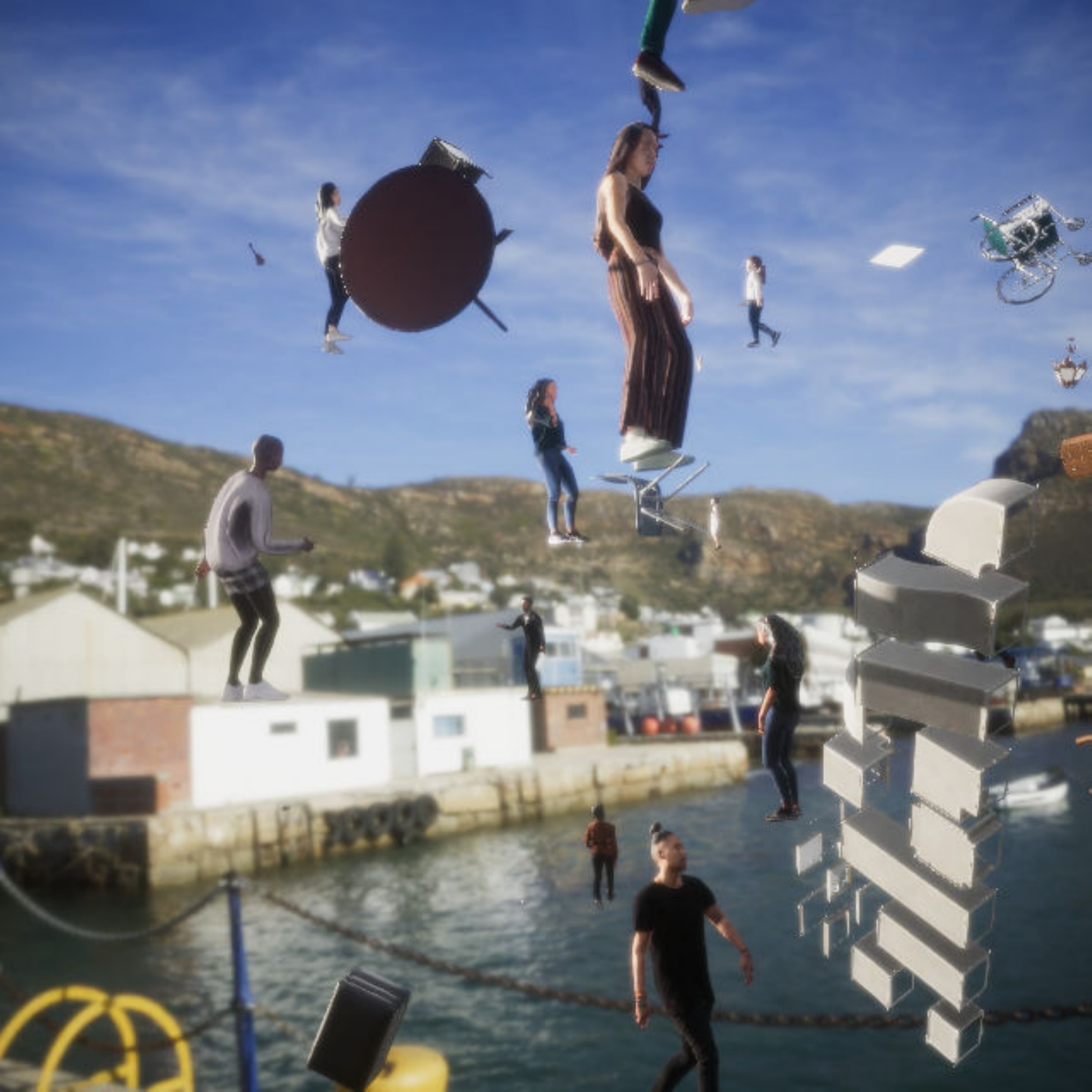}}
    \end{subfigure}
    \begin{subfigure}[t]{0.132\textwidth}
        {\includegraphics[height=2.3cm]{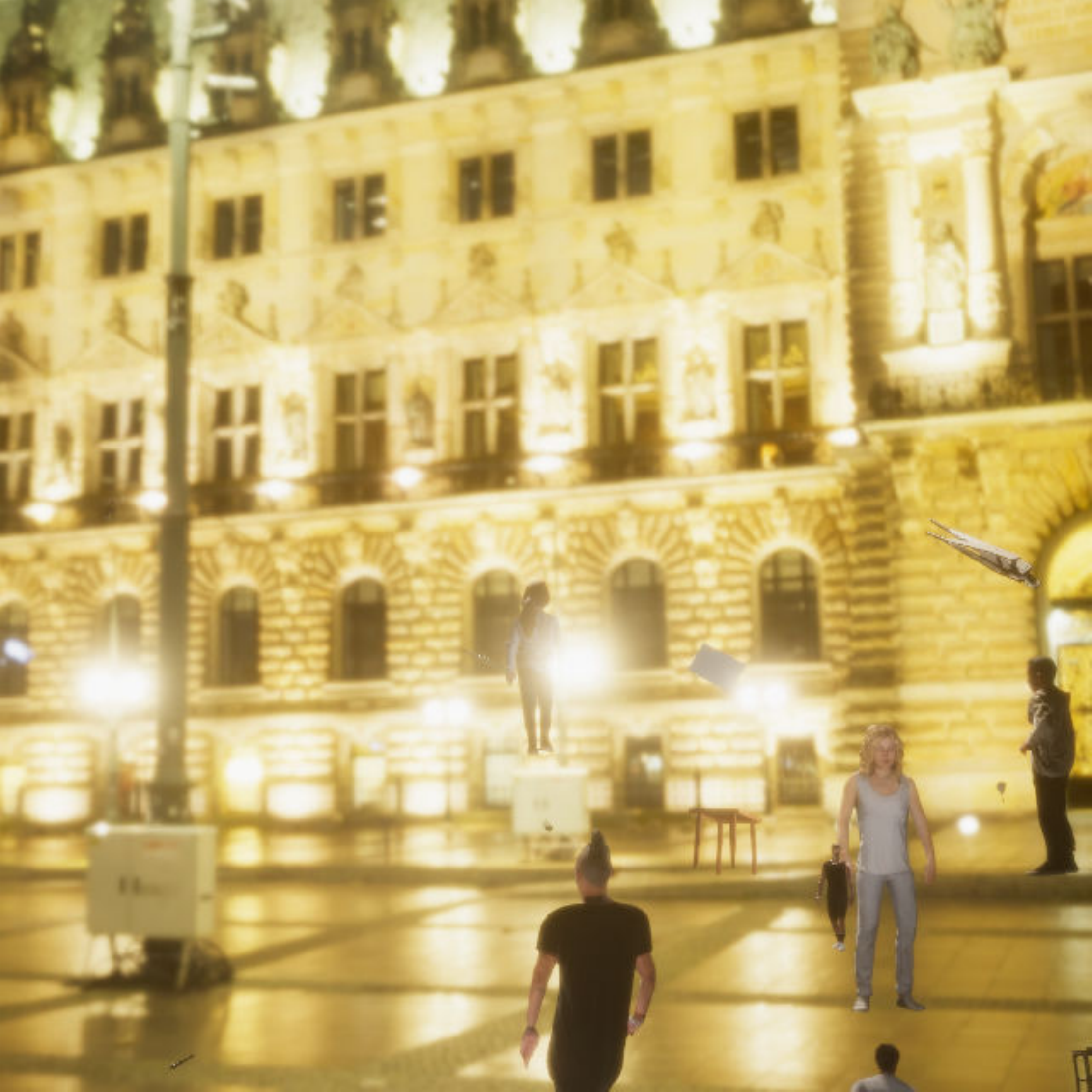}}
    \end{subfigure}
    \begin{subfigure}[t]{0.132\textwidth}
        {\includegraphics[height=2.3cm]{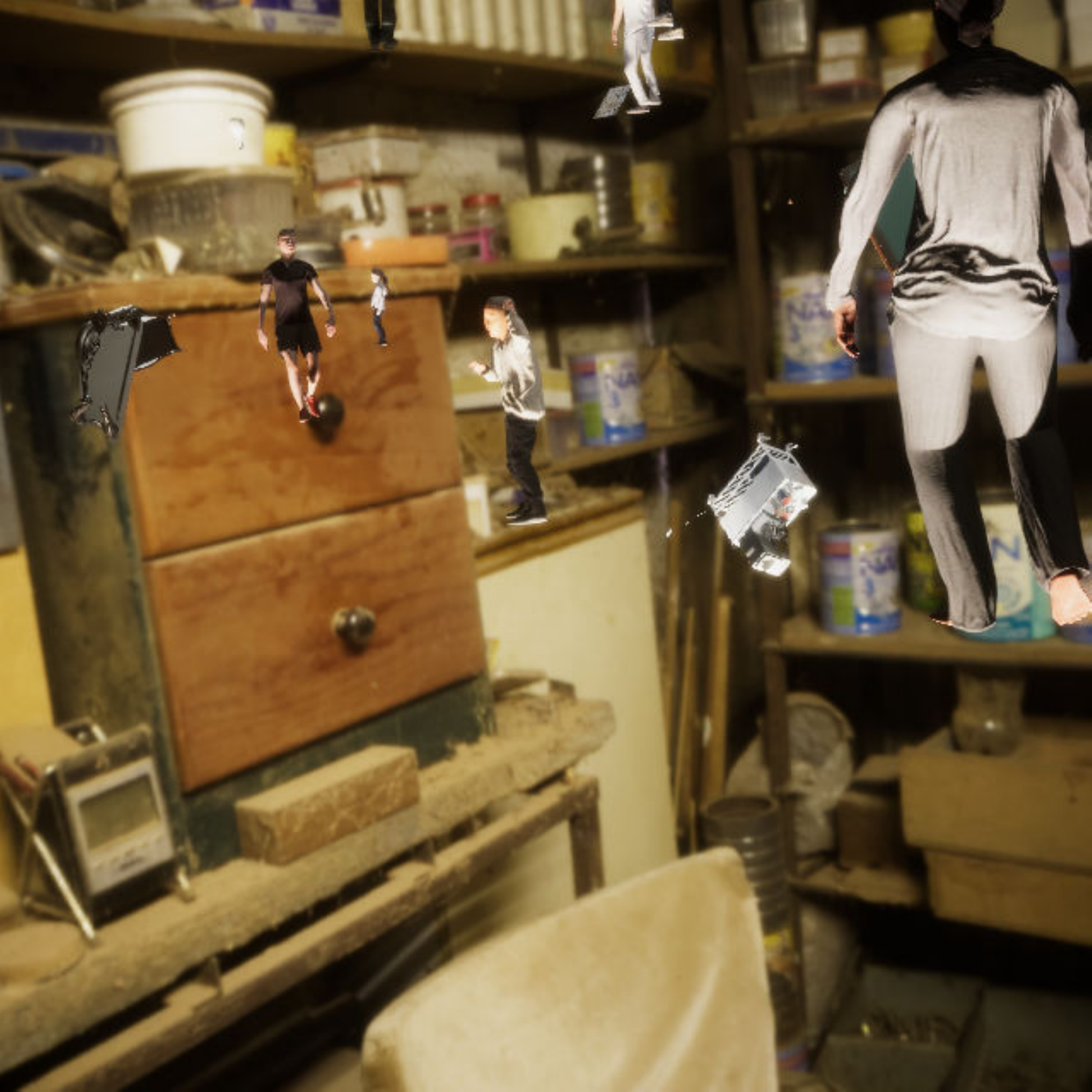}}
    \end{subfigure}
    \begin{subfigure}[t]{0.132\textwidth}
        {\includegraphics[height=2.3cm]{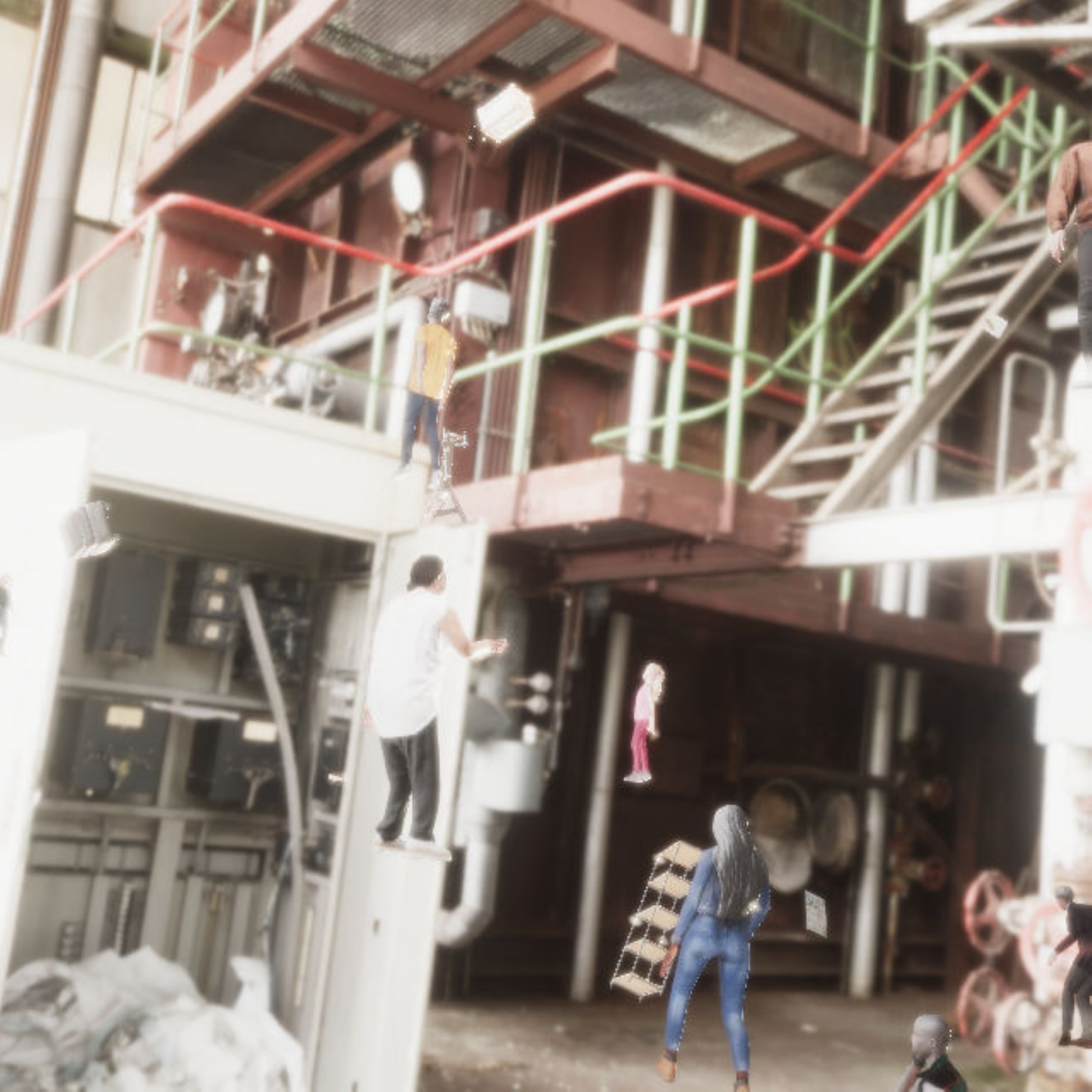}}
    \end{subfigure}
    \\
    \begin{subfigure}[t]{0.132\textwidth}
        {\includegraphics[height=2.3cm]{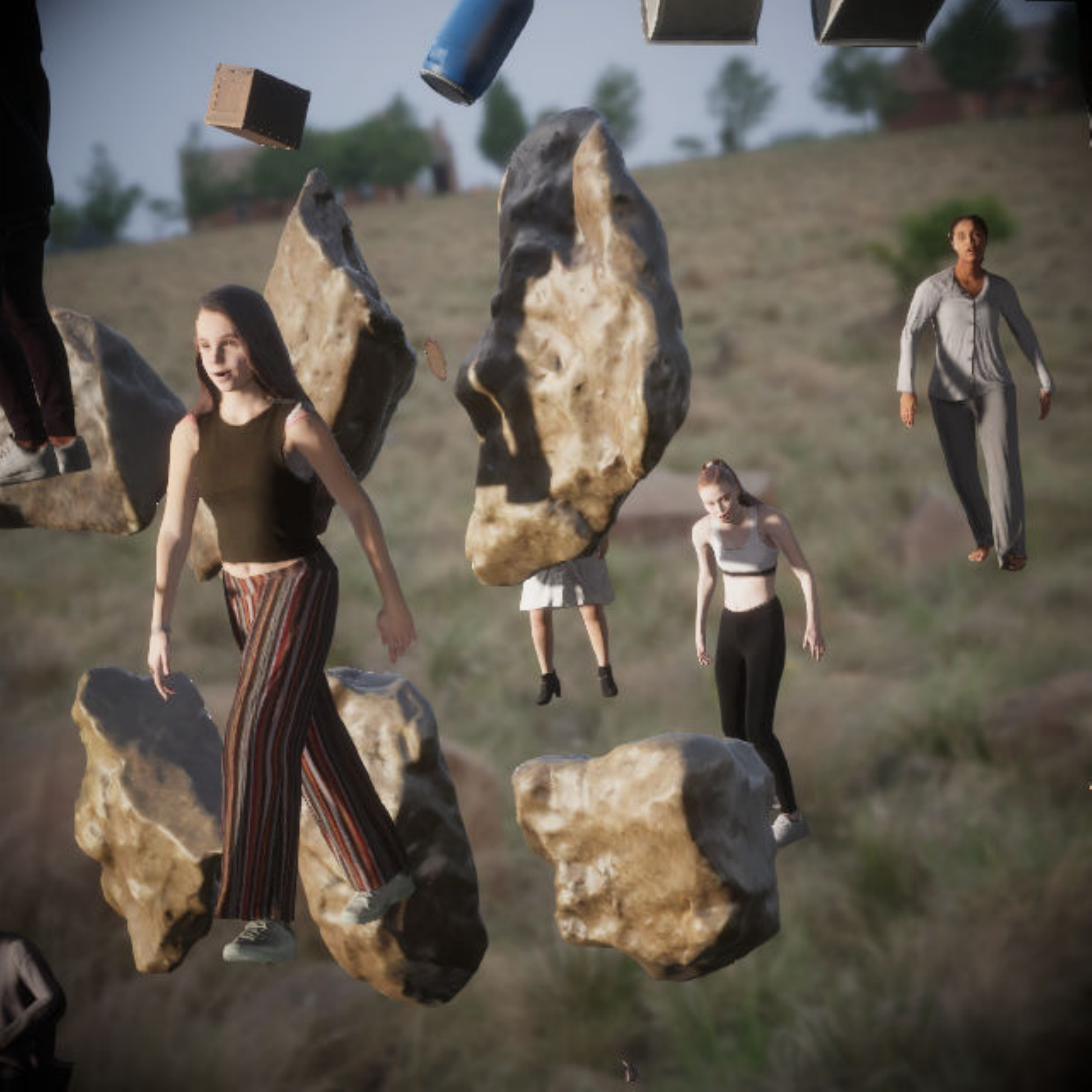}}
    \end{subfigure}
    \begin{subfigure}[t]{0.132\textwidth}
        {\includegraphics[height=2.3cm]{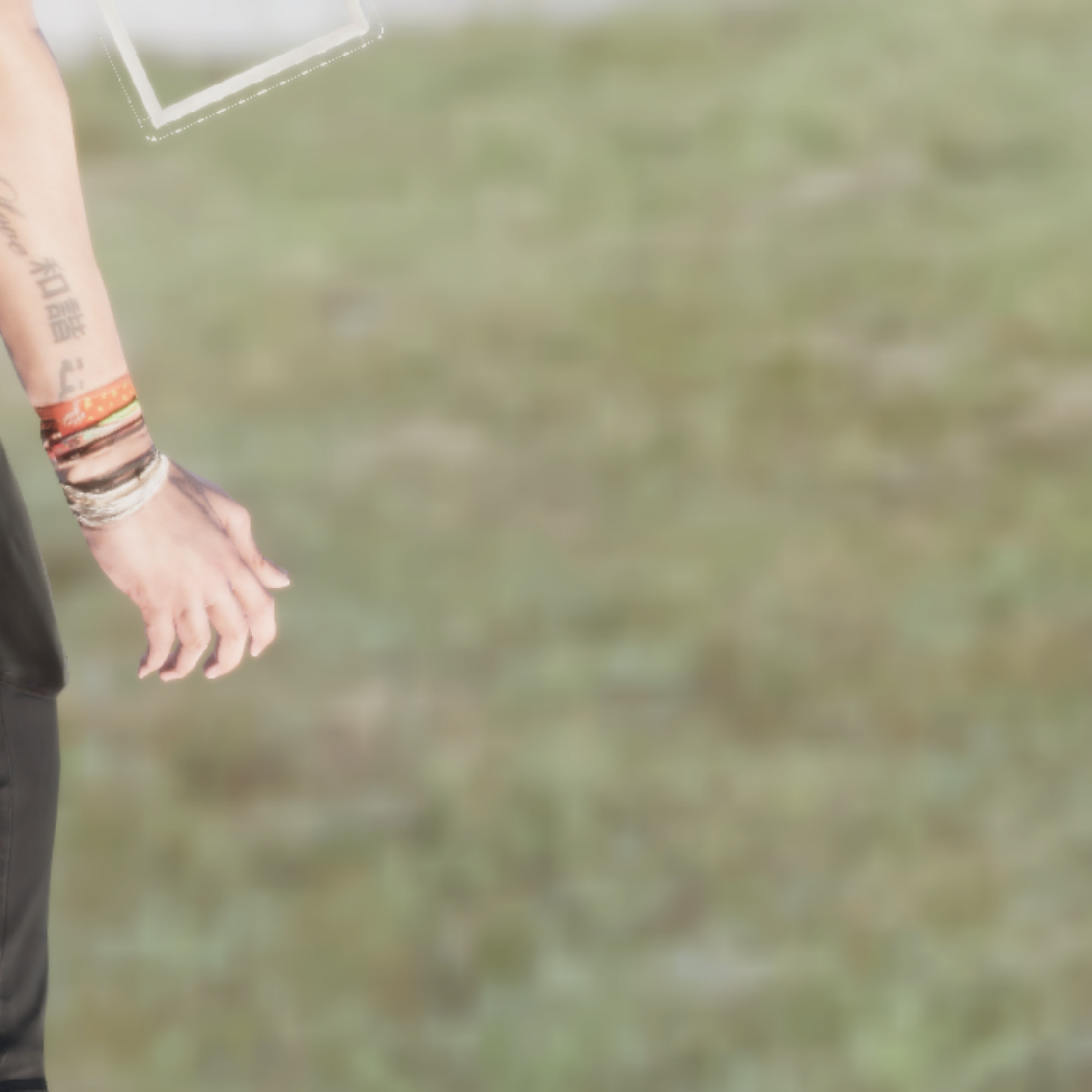}}
    \end{subfigure}
    \begin{subfigure}[t]{0.132\textwidth}
        {\includegraphics[height=2.3cm]{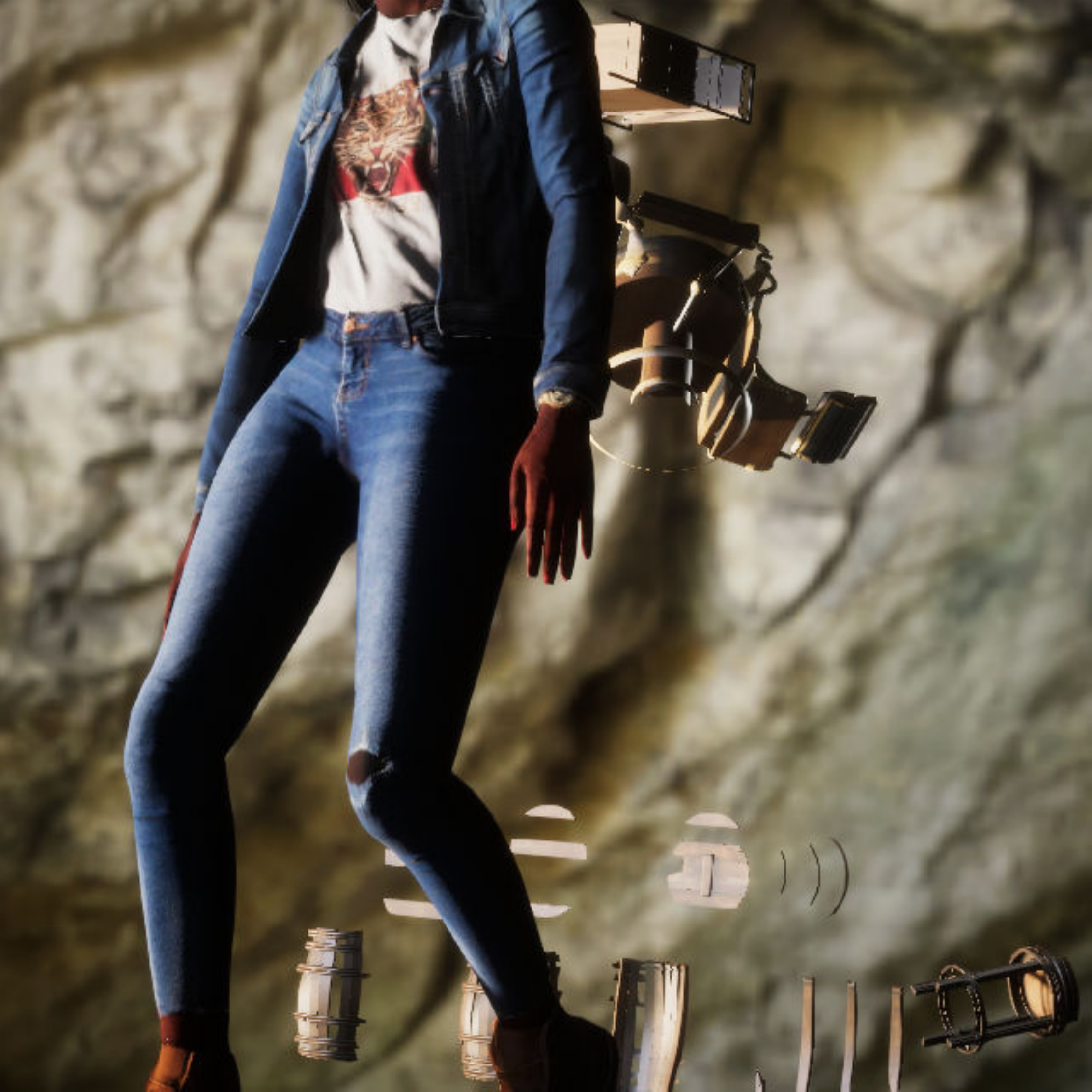}}
    \end{subfigure}
    \begin{subfigure}[t]{0.132\textwidth}
        {\includegraphics[height=2.3cm]{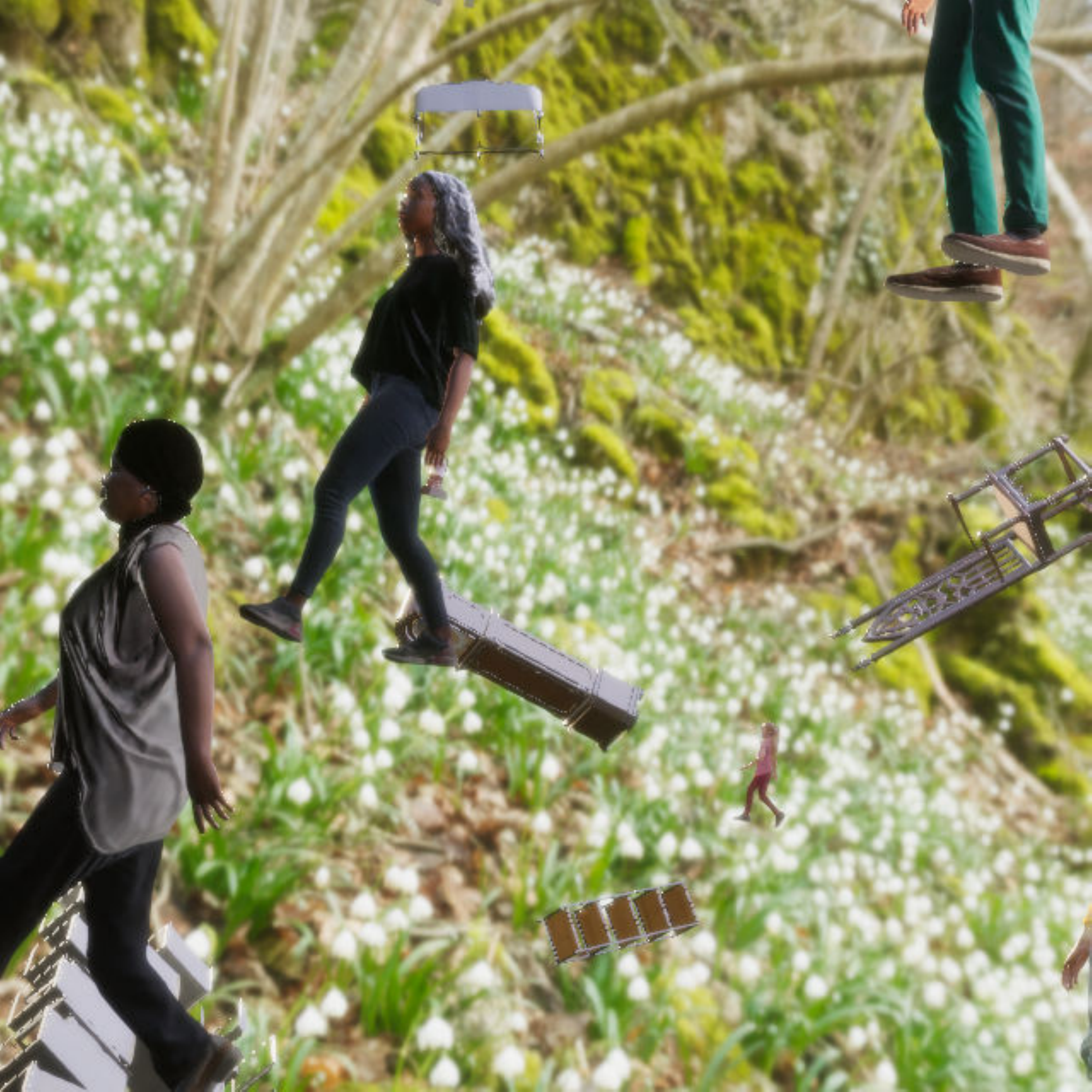}}
    \end{subfigure}
    \begin{subfigure}[t]{0.132\textwidth}
        {\includegraphics[height=2.3cm]{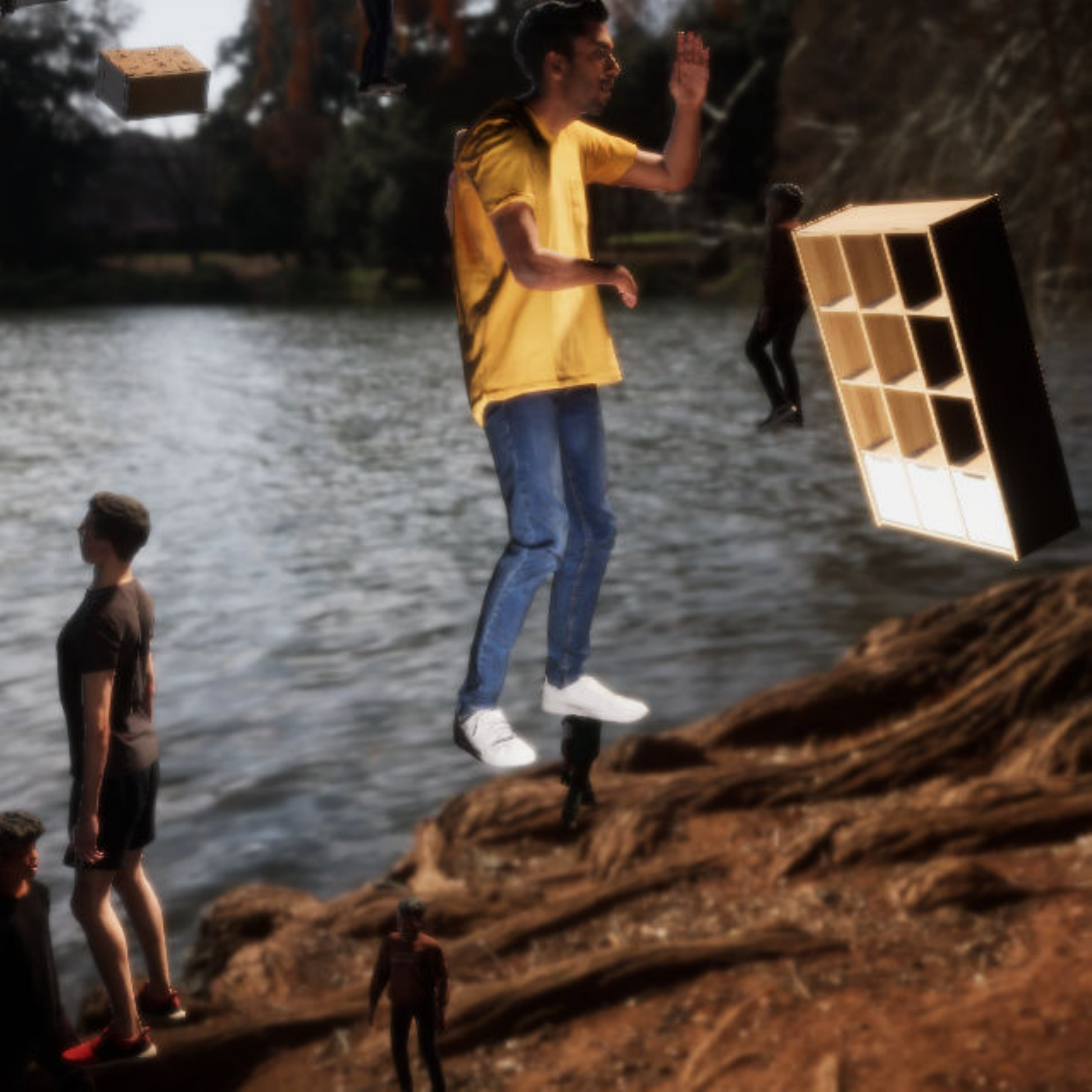}}
    \end{subfigure}
    \begin{subfigure}[t]{0.132\textwidth}
        {\includegraphics[height=2.3cm]{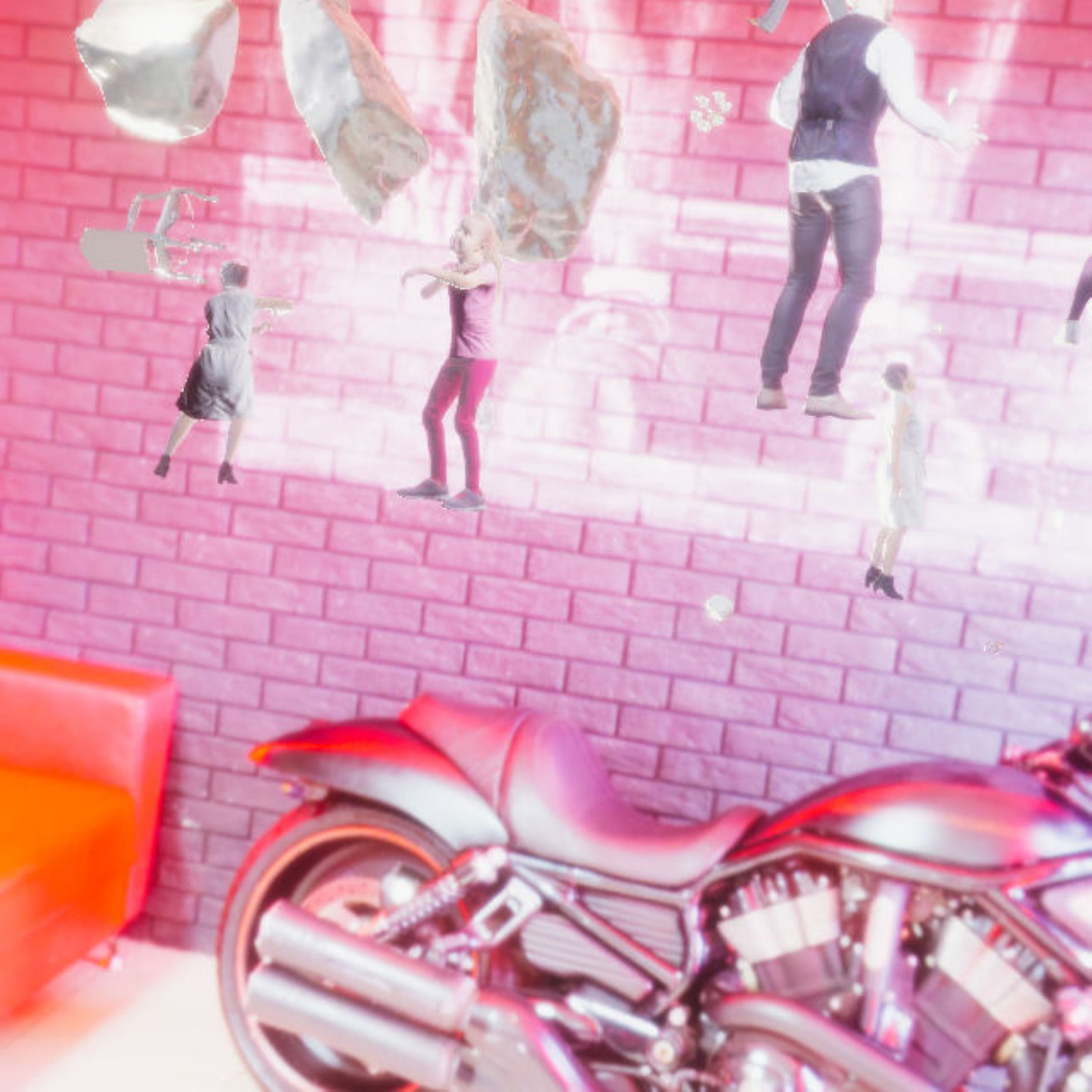}}
    \end{subfigure}
    \\
    \begin{subfigure}[t]{0.132\textwidth}
        {\includegraphics[height=2.3cm]{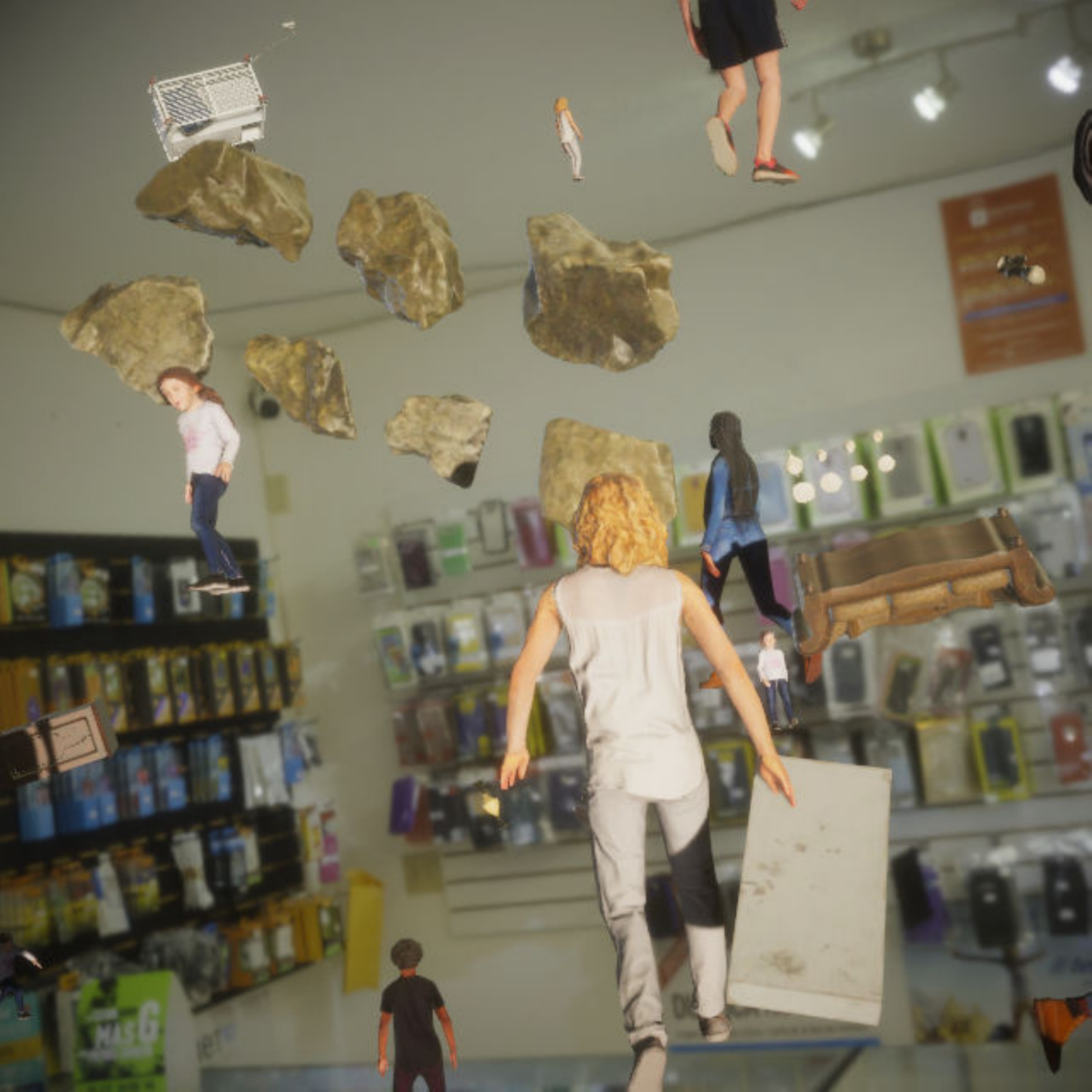}}
    \end{subfigure}
    \begin{subfigure}[t]{0.132\textwidth}
        {\includegraphics[height=2.3cm]{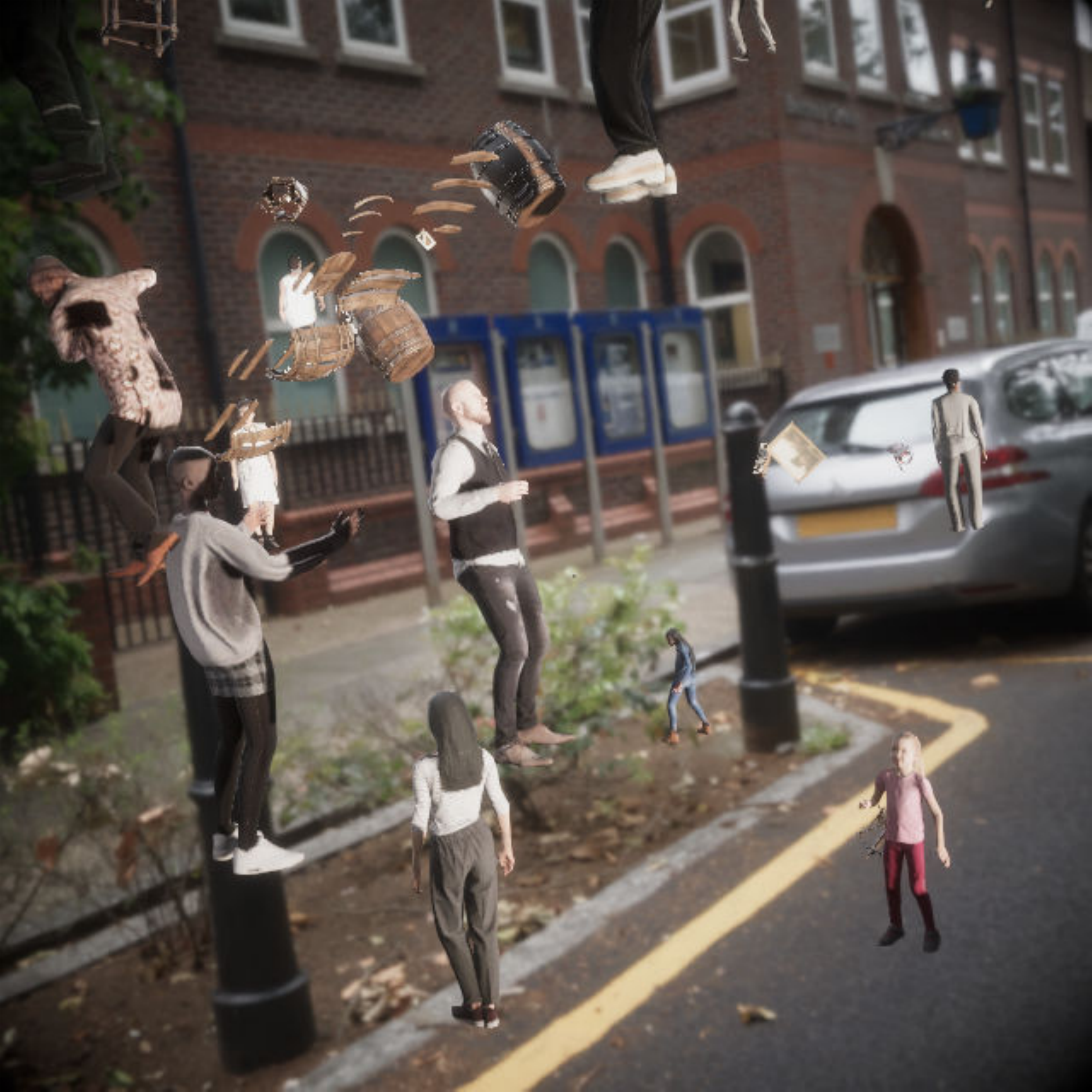}}
    \end{subfigure}
    \begin{subfigure}[t]{0.132\textwidth}
        {\includegraphics[height=2.3cm]{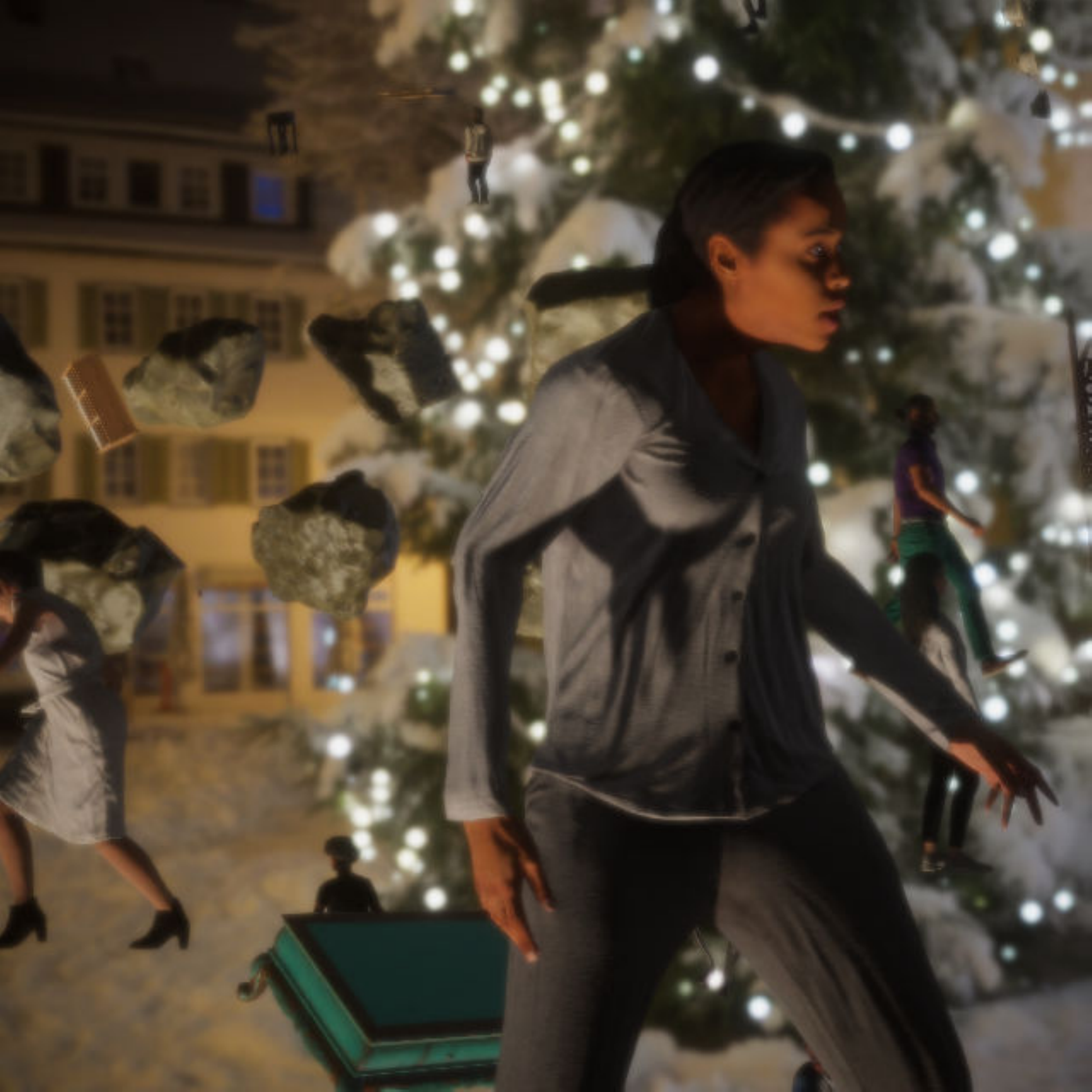}}
    \end{subfigure}
    \begin{subfigure}[t]{0.132\textwidth}
        {\includegraphics[height=2.3cm]{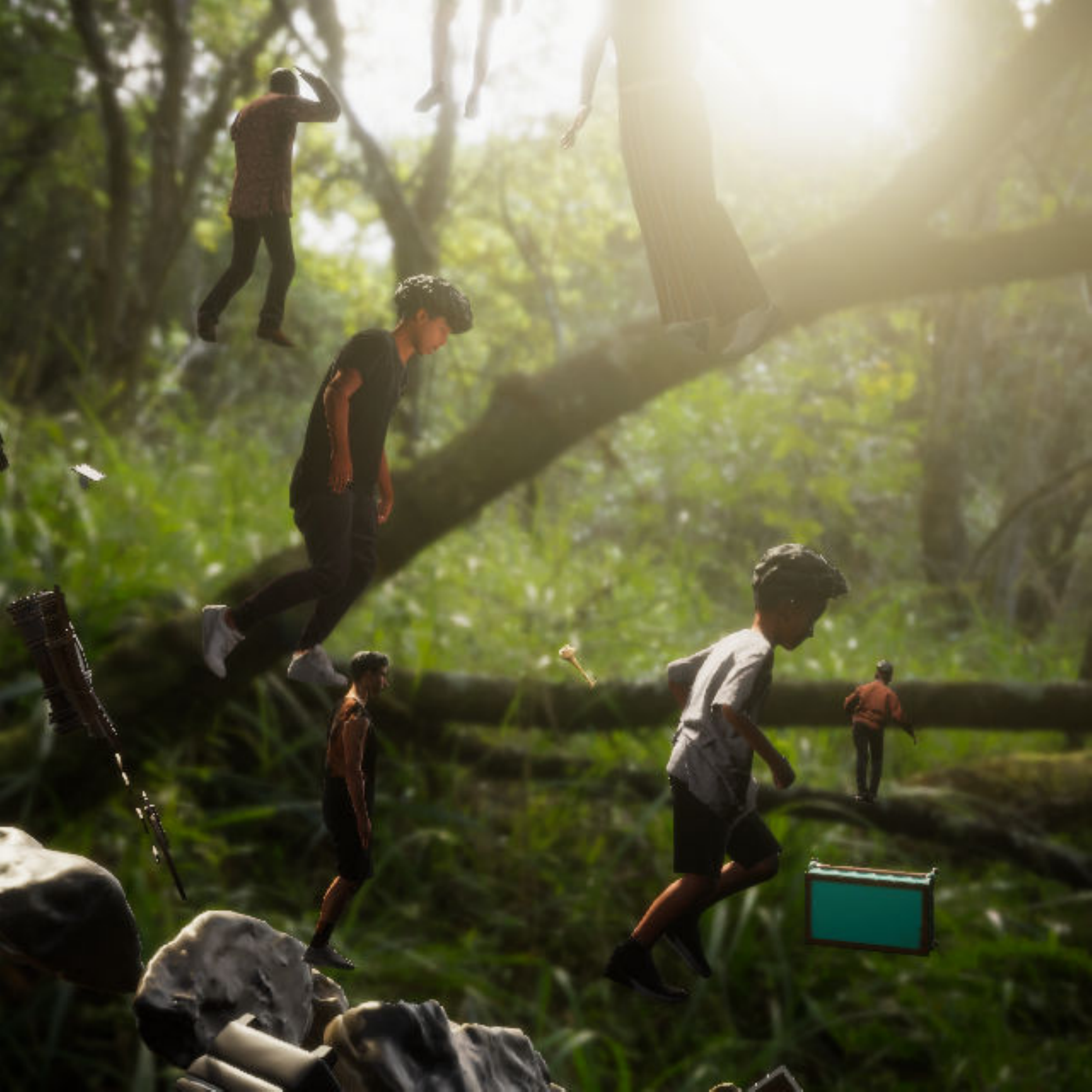}}
    \end{subfigure}
    \begin{subfigure}[t]{0.132\textwidth}
        {\includegraphics[height=2.3cm]{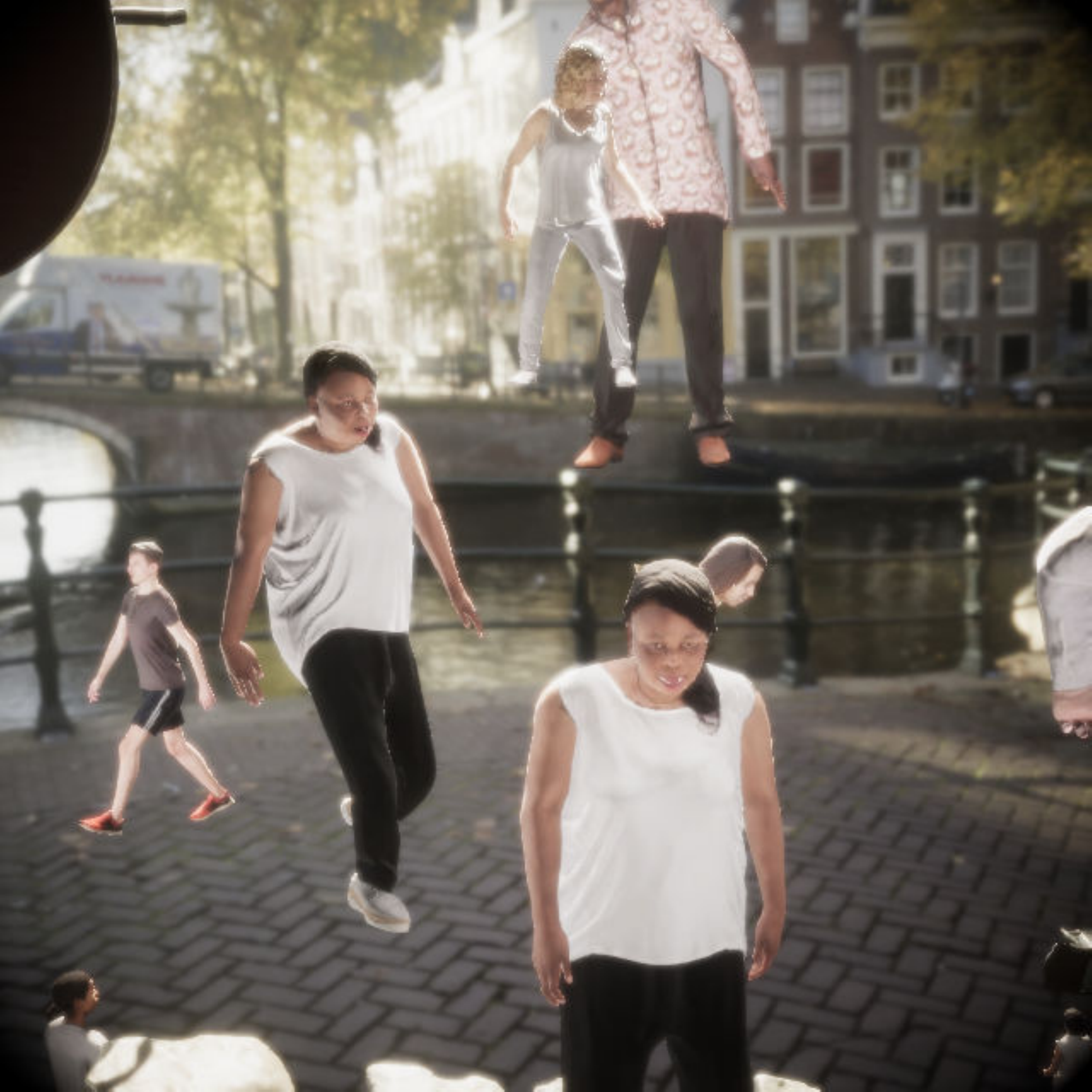}}
    \end{subfigure}
    \begin{subfigure}[t]{0.132\textwidth}
        {\includegraphics[height=2.3cm]{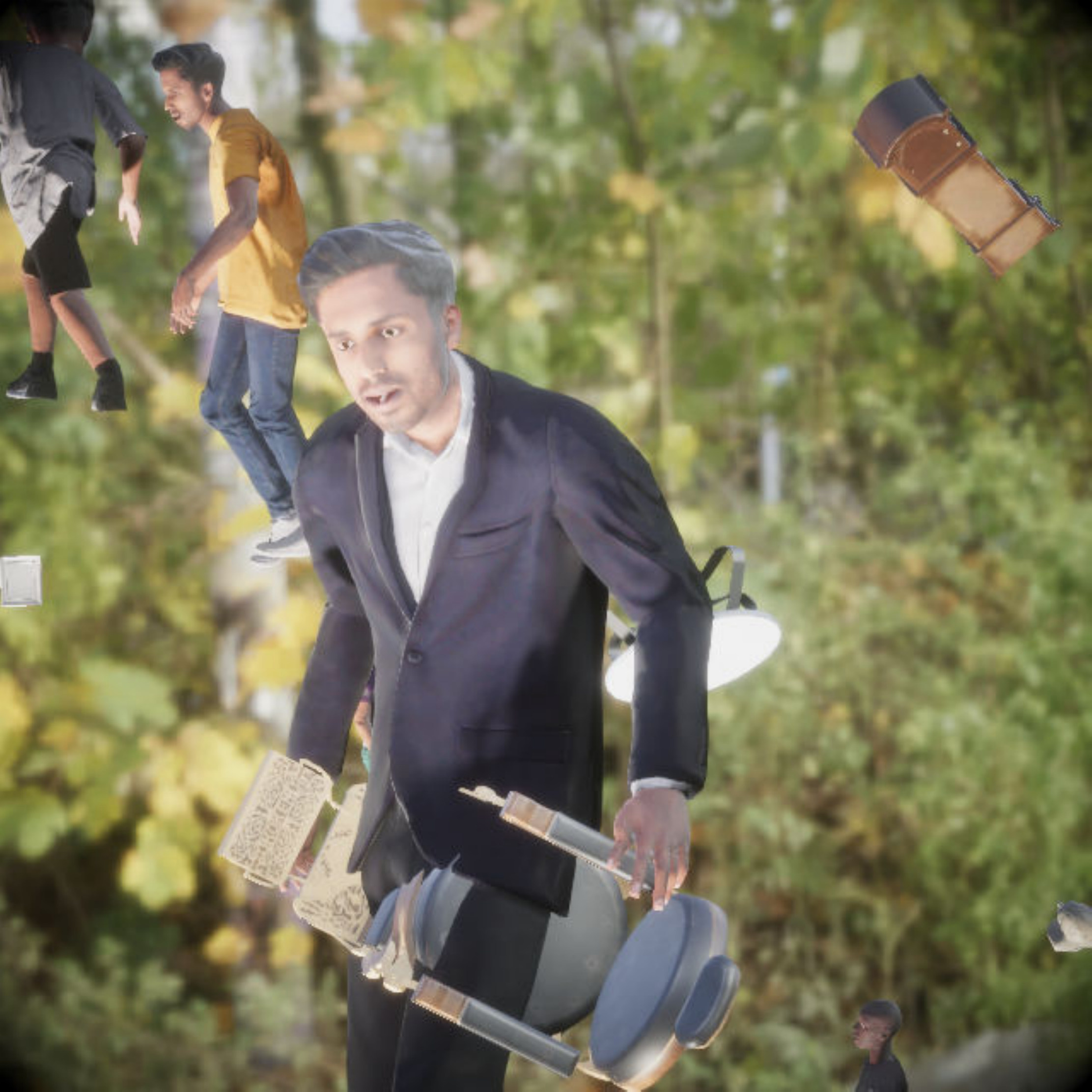}}
    \end{subfigure}
    \\
    \begin{subfigure}[t]{0.132\textwidth}
        {\includegraphics[height=2.3cm]{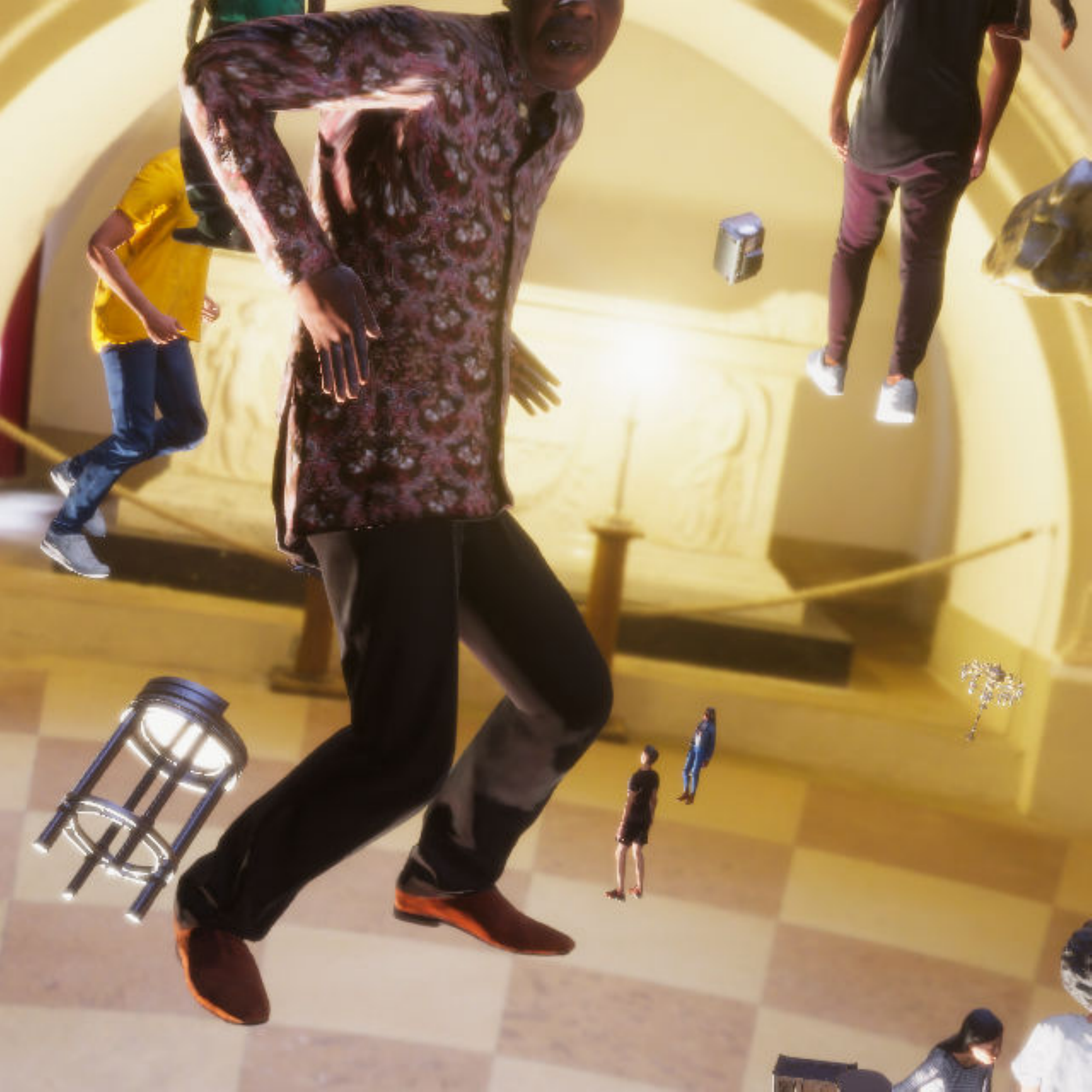}}
    \end{subfigure}
    \begin{subfigure}[t]{0.132\textwidth}
        {\includegraphics[height=2.3cm]{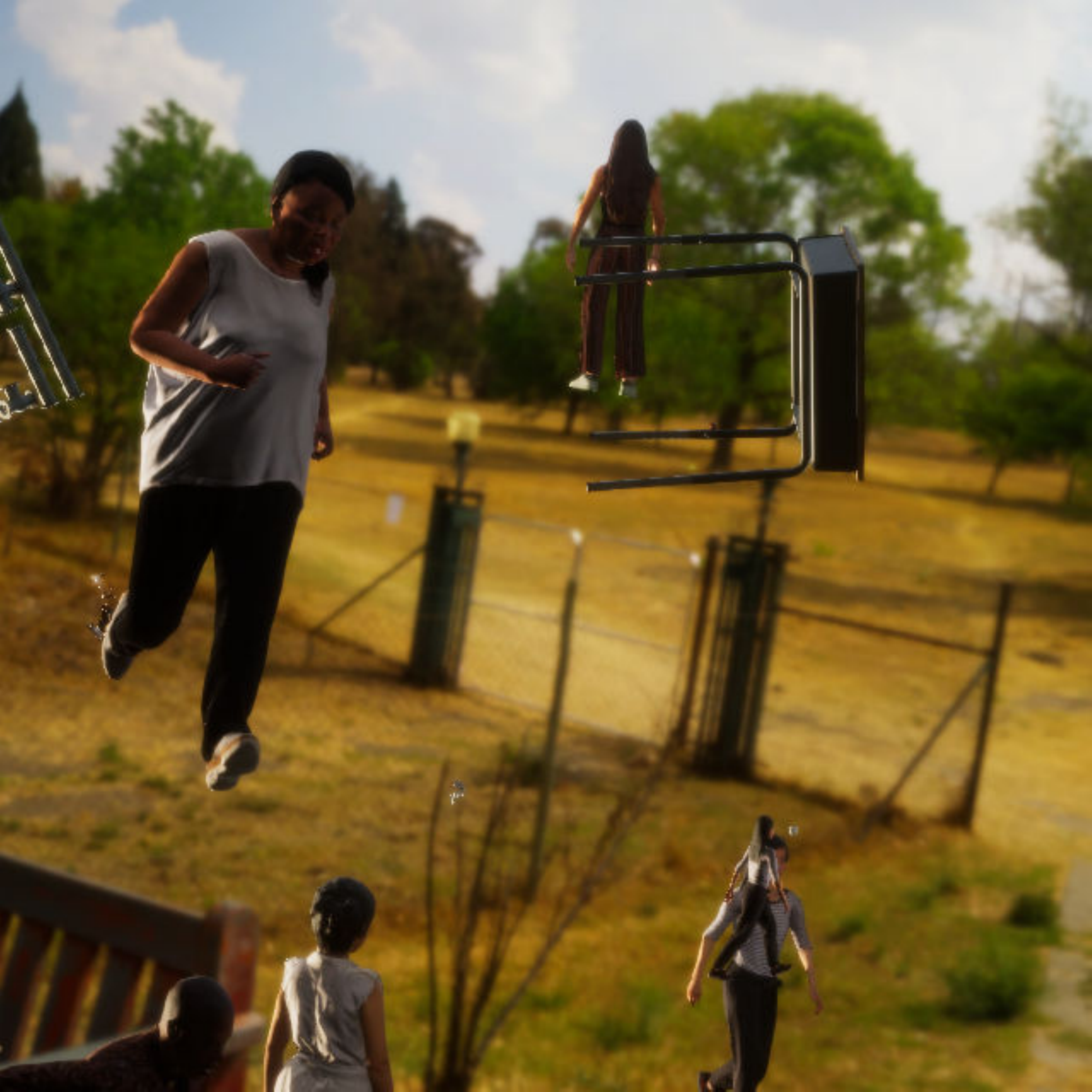}}
    \end{subfigure}
    \begin{subfigure}[t]{0.132\textwidth}
        {\includegraphics[height=2.3cm]{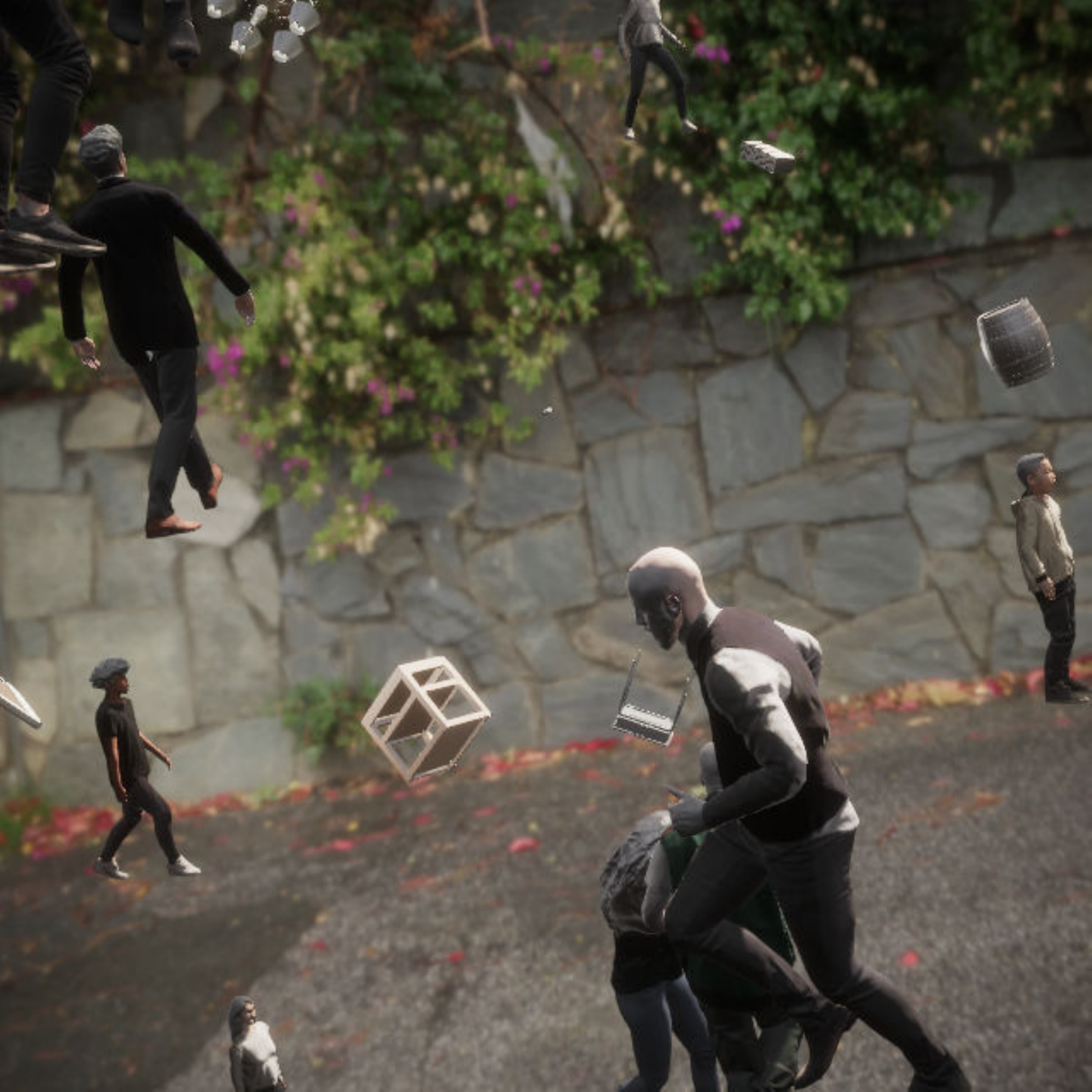}}
    \end{subfigure}
    \begin{subfigure}[t]{0.132\textwidth}
        {\includegraphics[height=2.3cm]{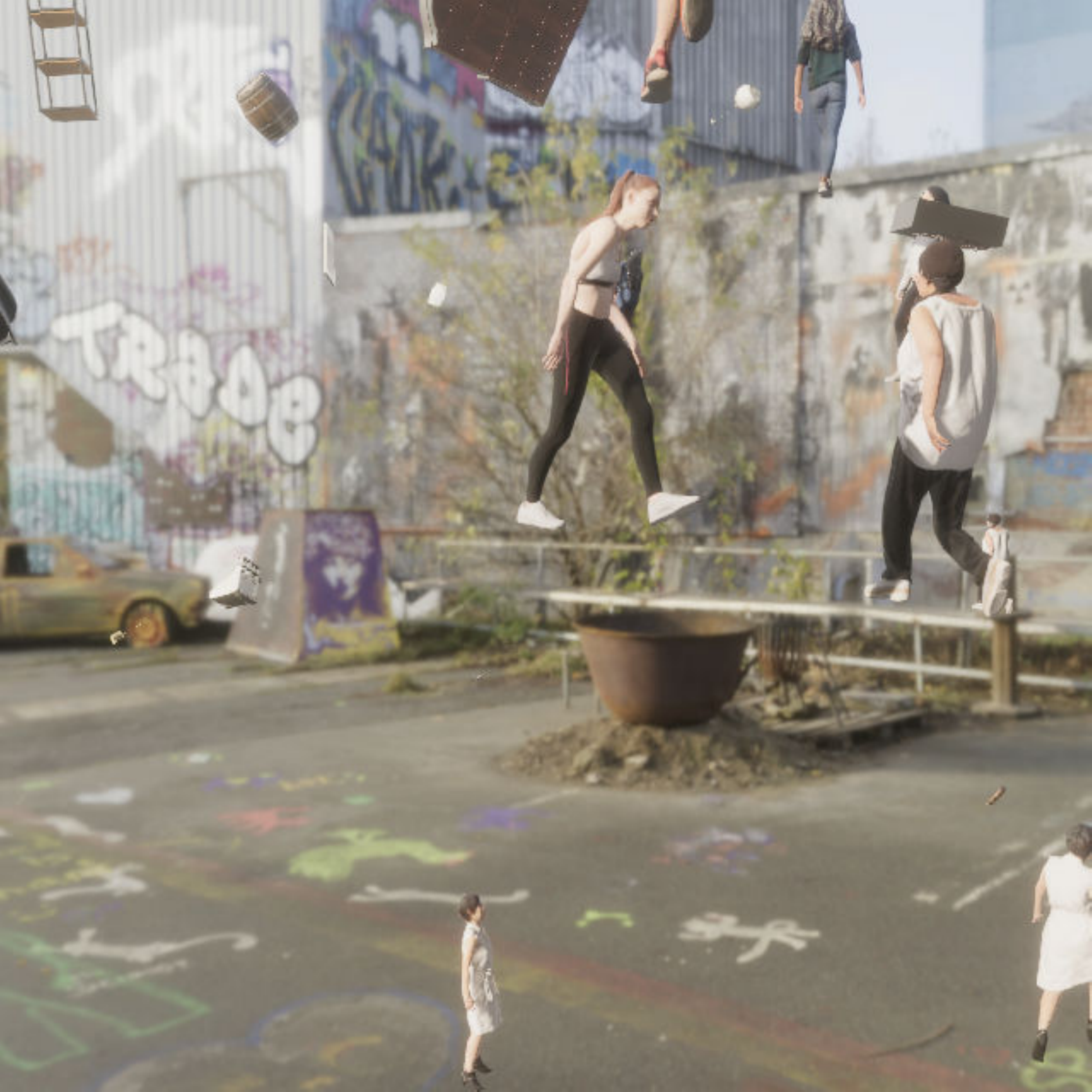}}
    \end{subfigure}
    \begin{subfigure}[t]{0.132\textwidth}
        {\includegraphics[height=2.3cm]{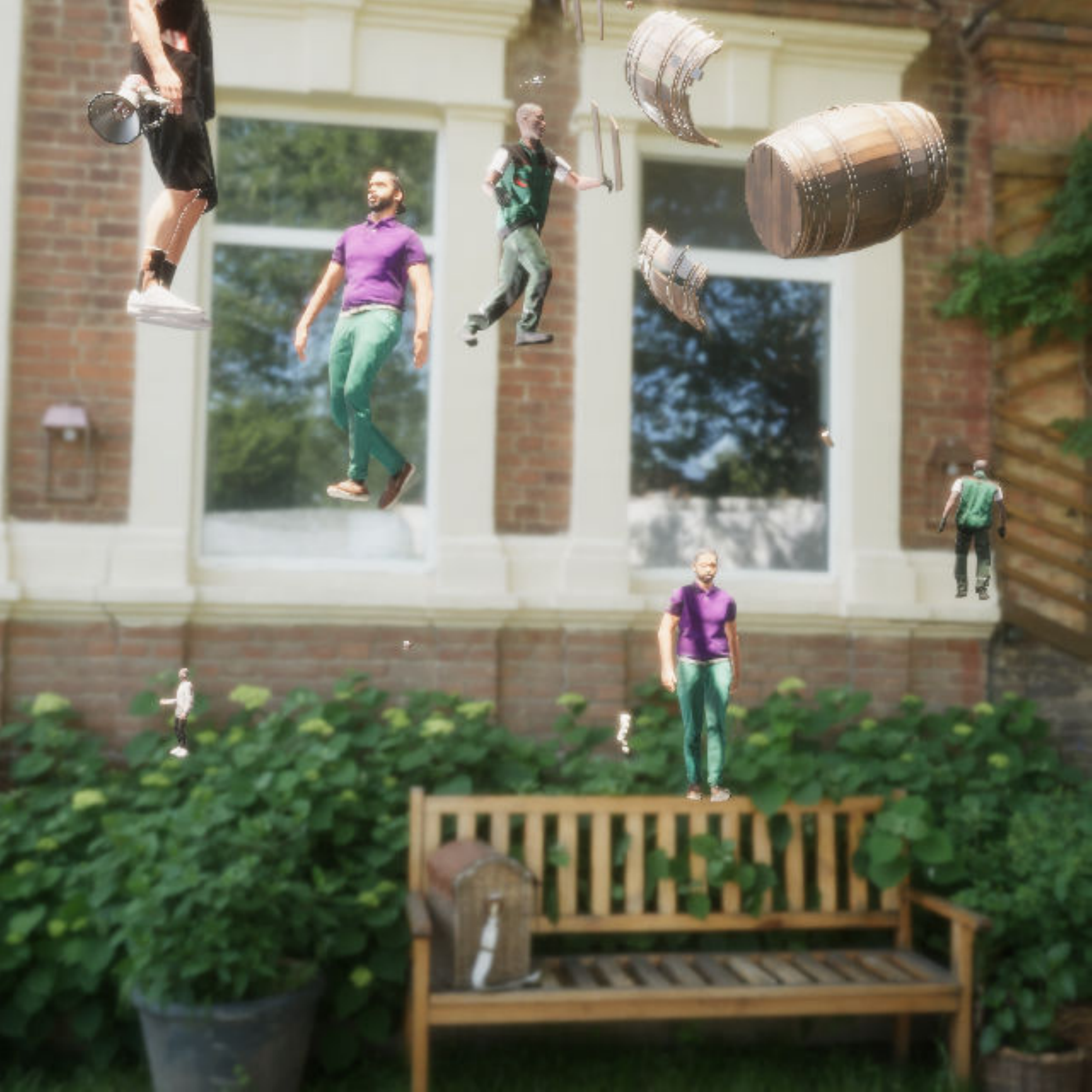}}
    \end{subfigure}
    \begin{subfigure}[t]{0.132\textwidth}
        {\includegraphics[height=2.3cm]{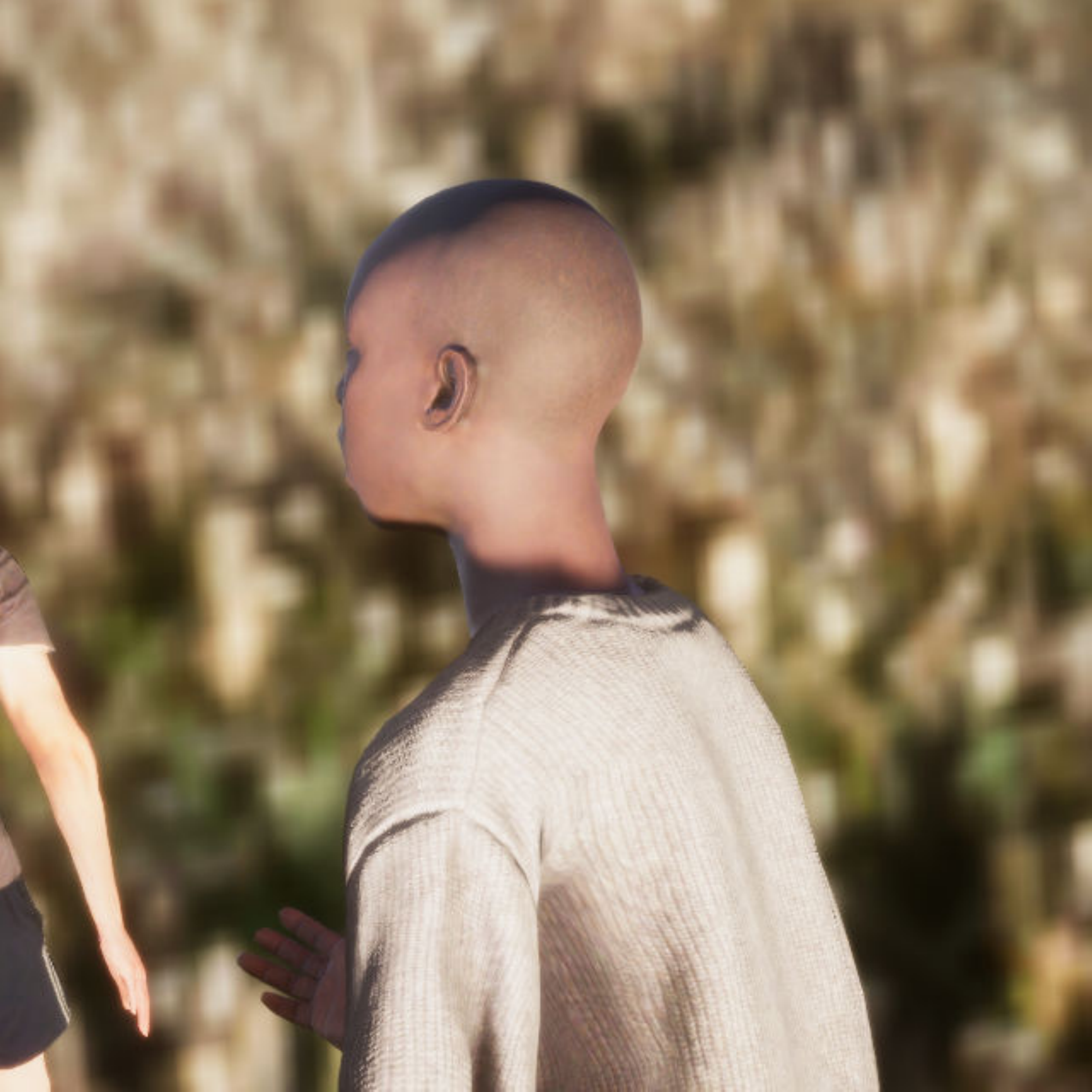}}
    \end{subfigure}
    \\
    \begin{subfigure}[t]{0.132\textwidth}
        {\includegraphics[height=2.3cm]{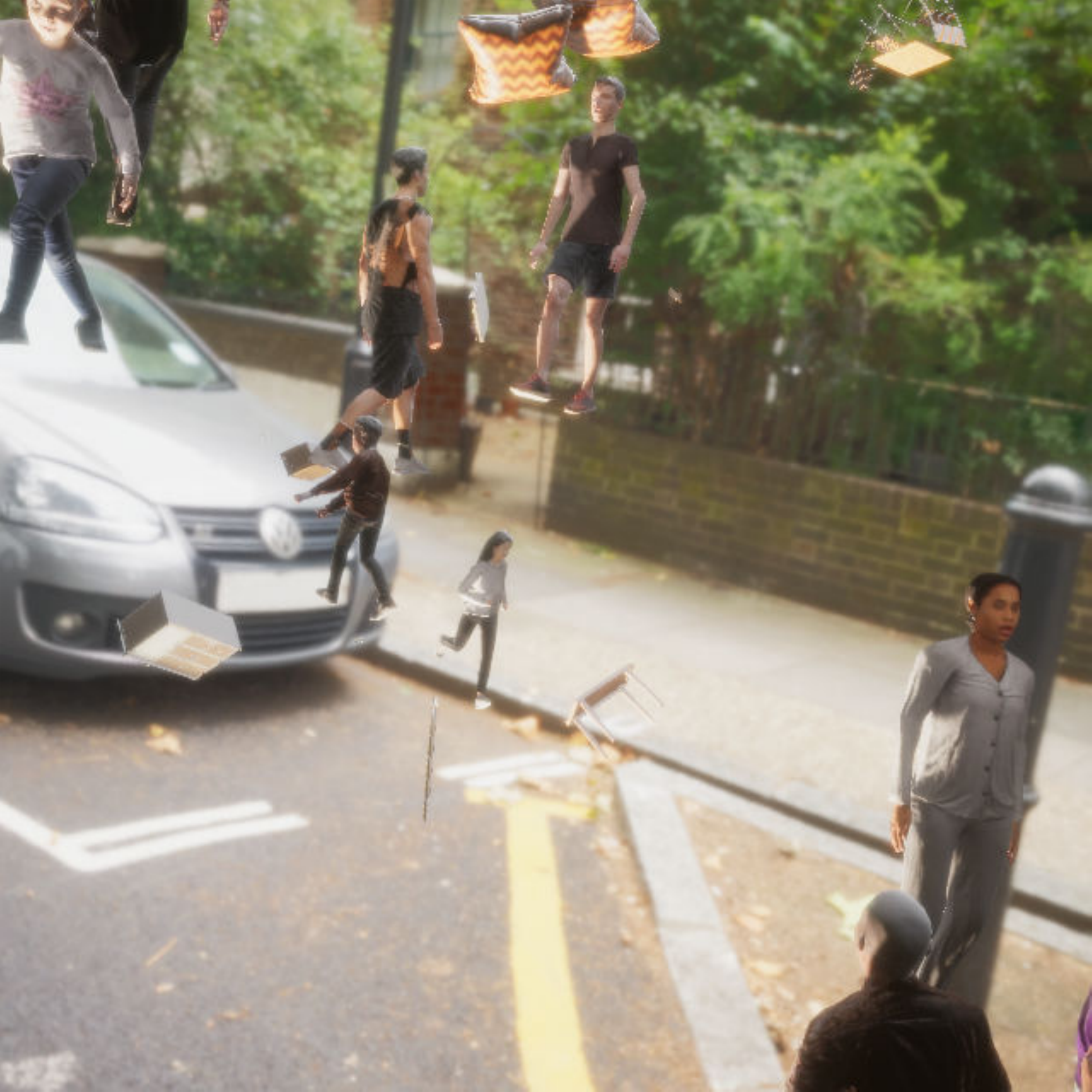}}
    \end{subfigure}
    \begin{subfigure}[t]{0.132\textwidth}
        {\includegraphics[height=2.3cm]{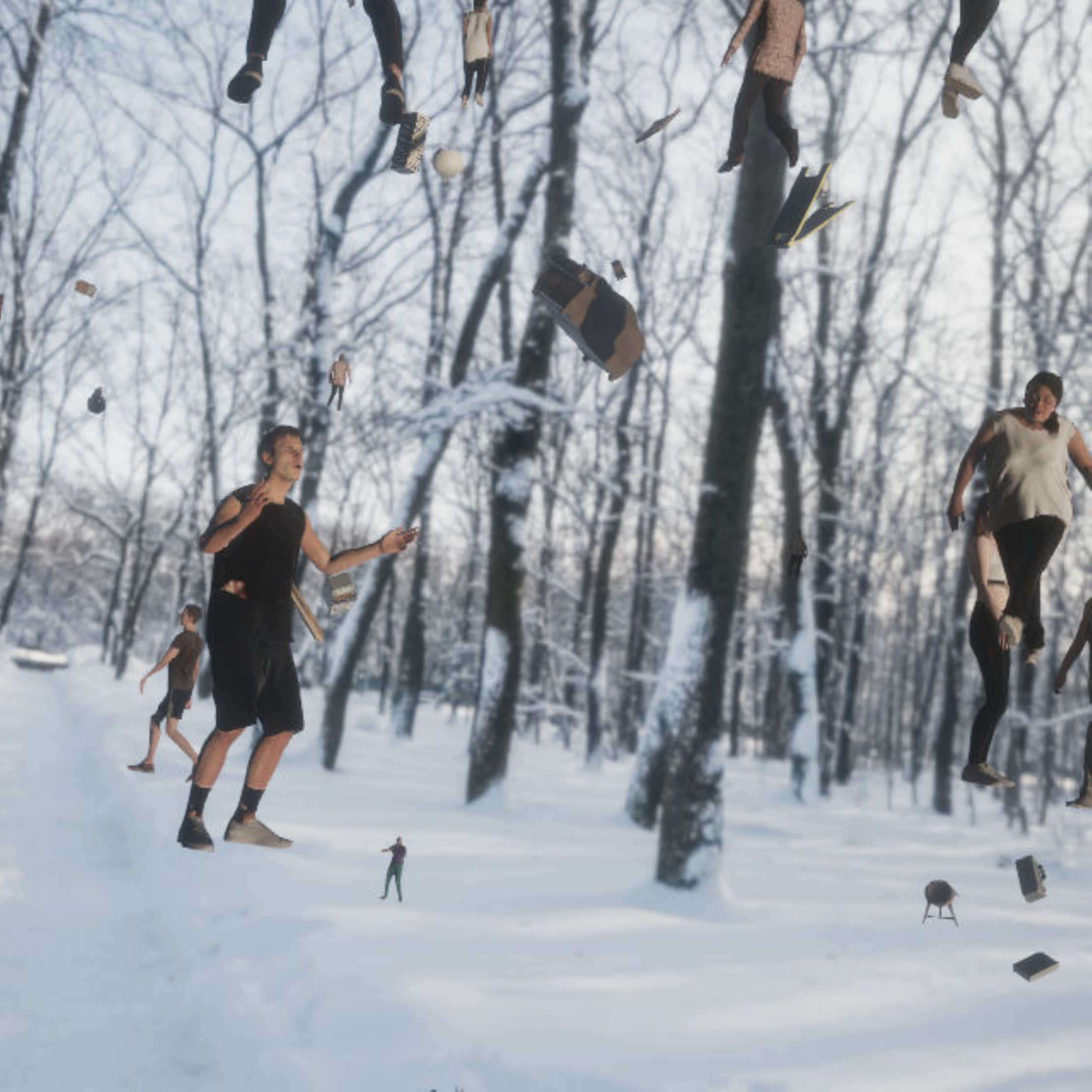}}
    \end{subfigure}
    \begin{subfigure}[t]{0.132\textwidth}
        {\includegraphics[height=2.3cm]{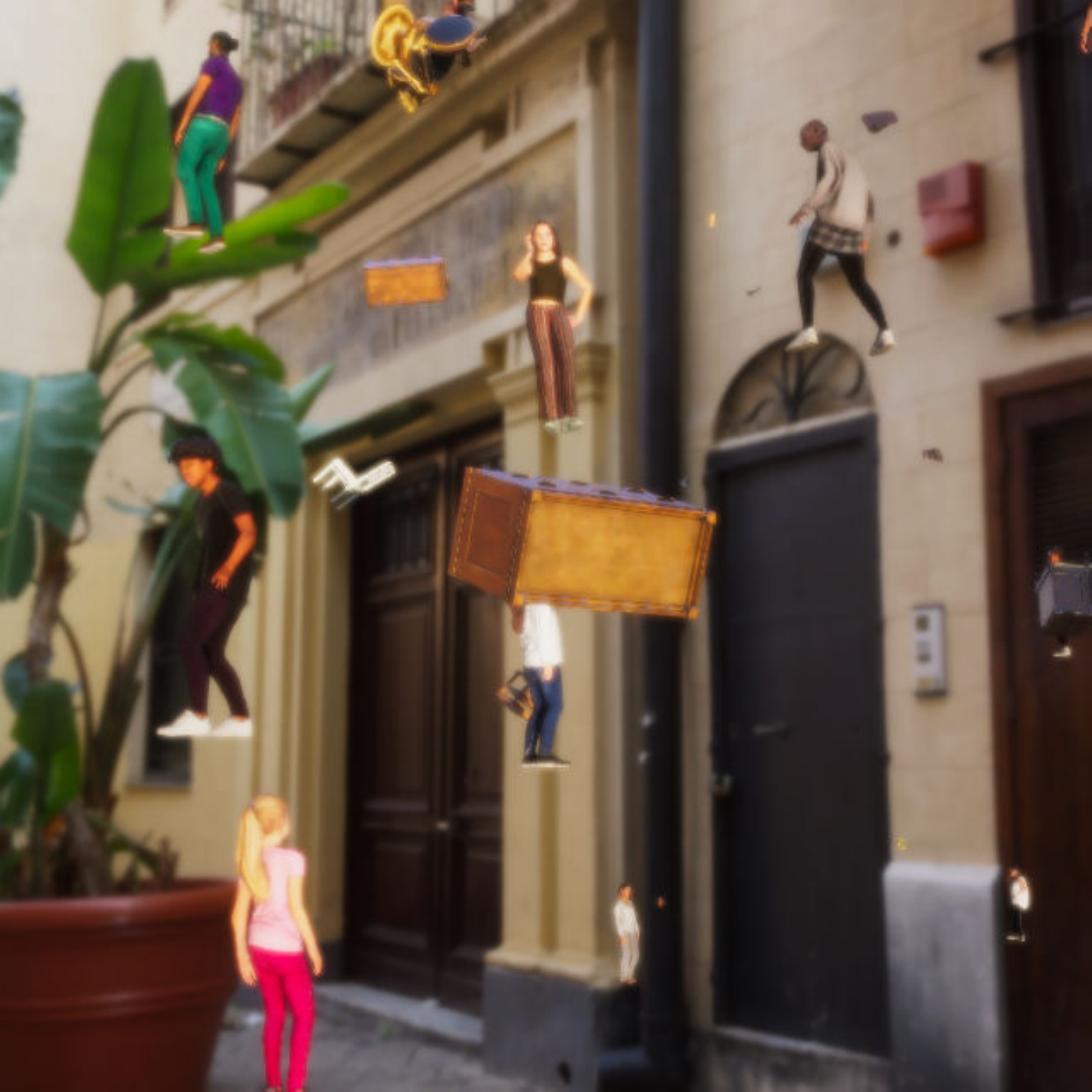}}
    \end{subfigure}
    \begin{subfigure}[t]{0.132\textwidth}
        {\includegraphics[height=2.3cm]{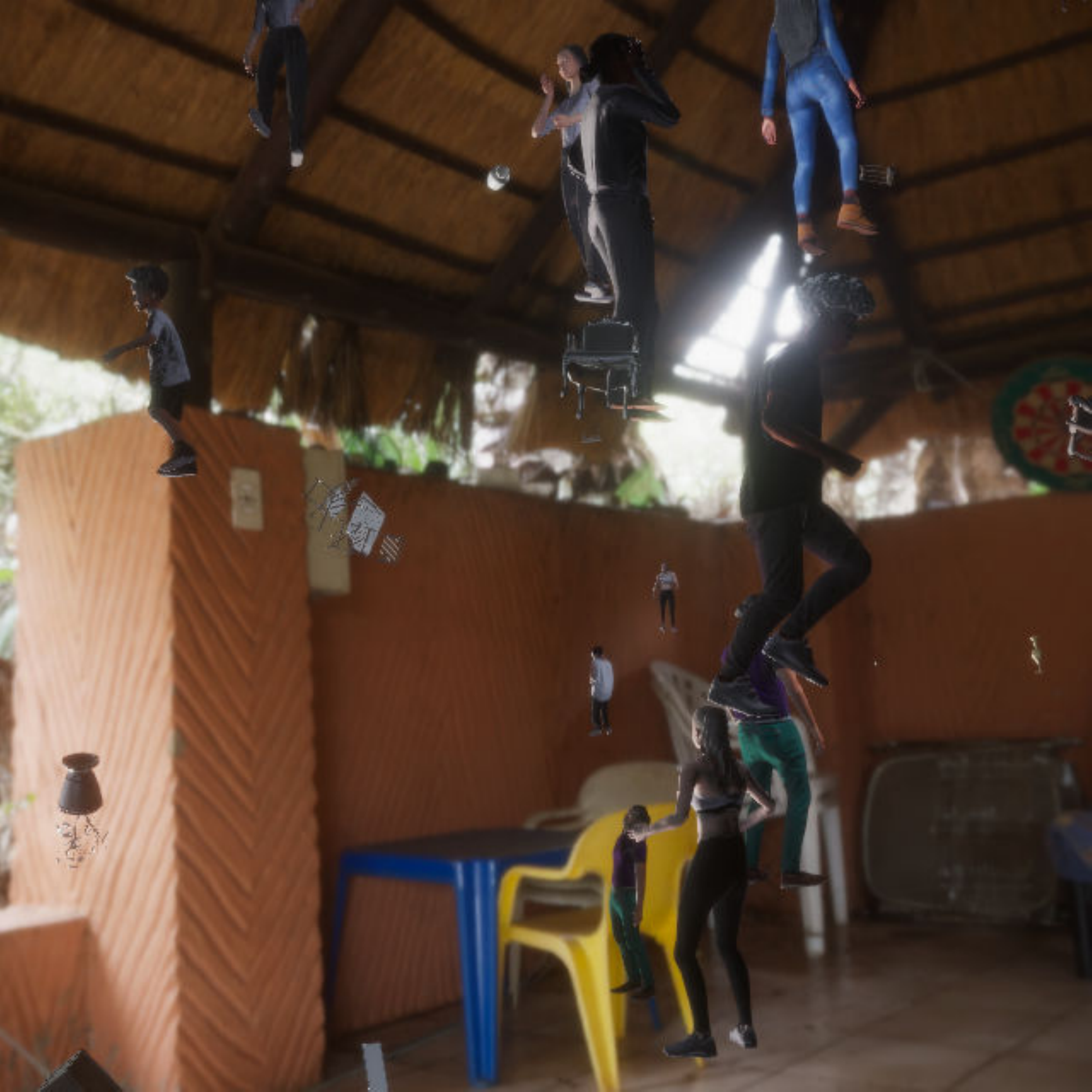}}
    \end{subfigure}
    \begin{subfigure}[t]{0.132\textwidth}
        {\includegraphics[height=2.3cm]{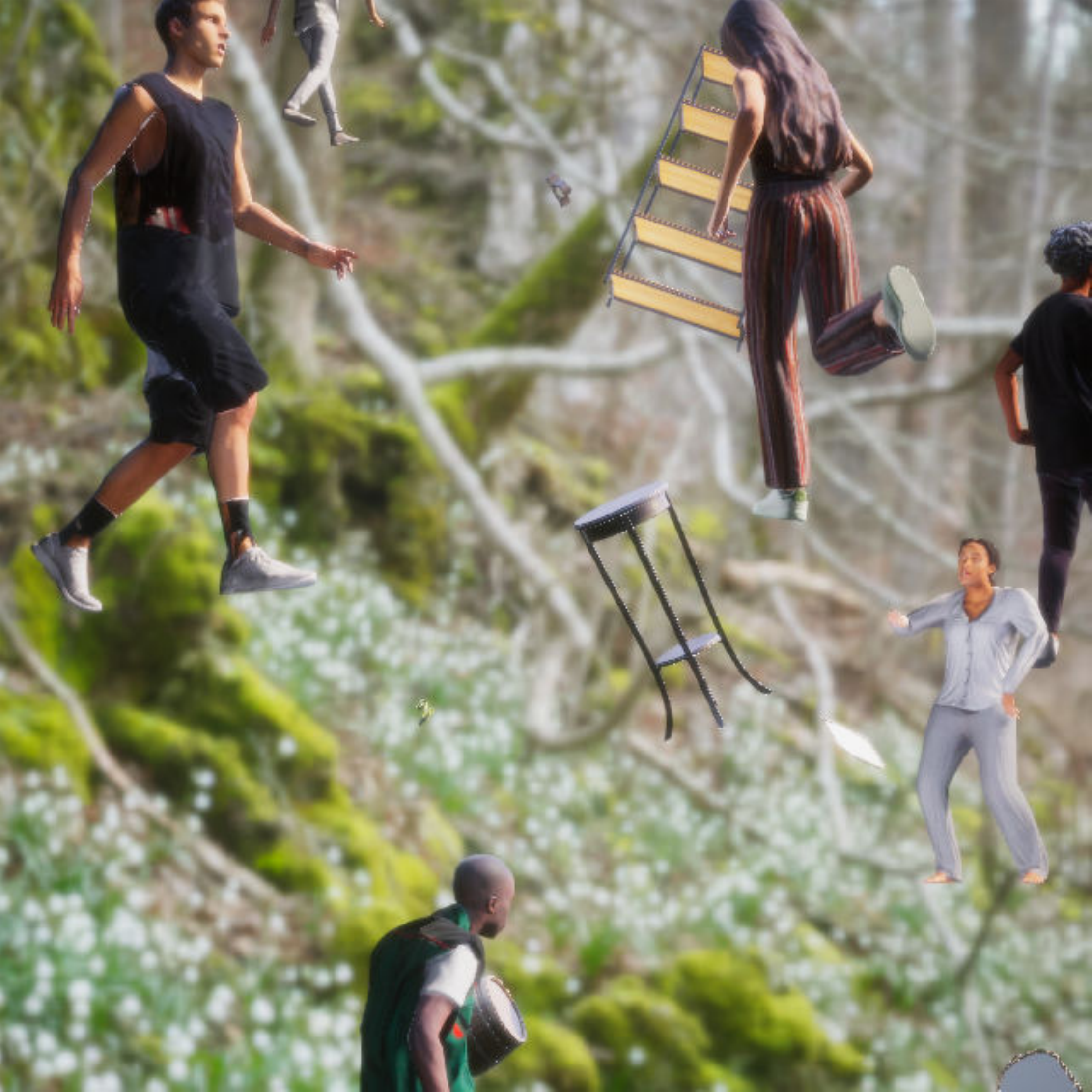}}
    \end{subfigure}
    \begin{subfigure}[t]{0.132\textwidth}
        {\includegraphics[height=2.3cm]{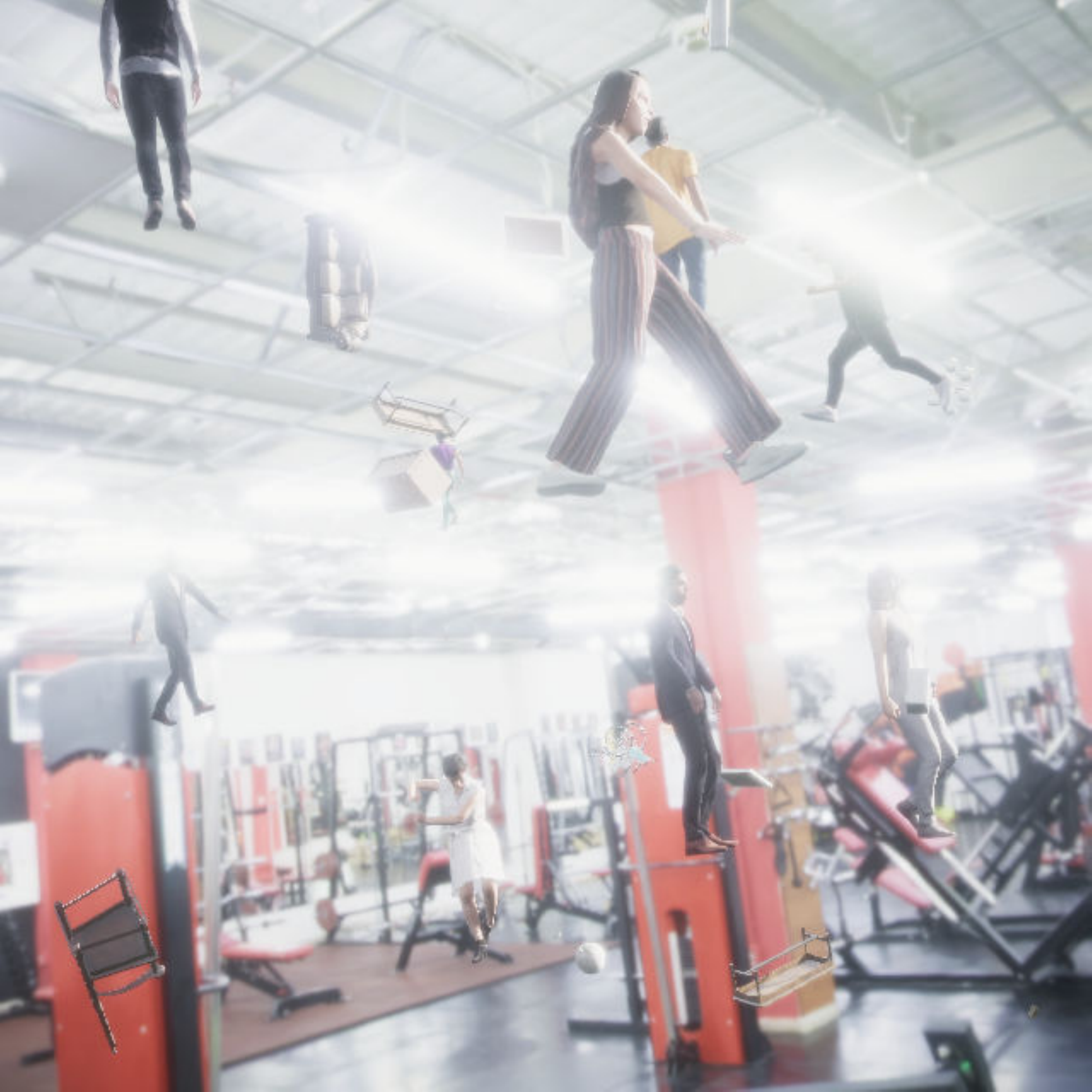}}
    \end{subfigure}
    \caption{\textbf{More examples from PSP-HDRI$+$}. Best viewed on the screen.}
    \label{fig:psp-hdri+_appendix}
\end{figure}

\section{Appendix: Pose Diversity}
\begin{figure}[htb] 
    \centering
    \begin{subfigure}[t]{0.14\textwidth}
        {\includegraphics[height=2.5cm]{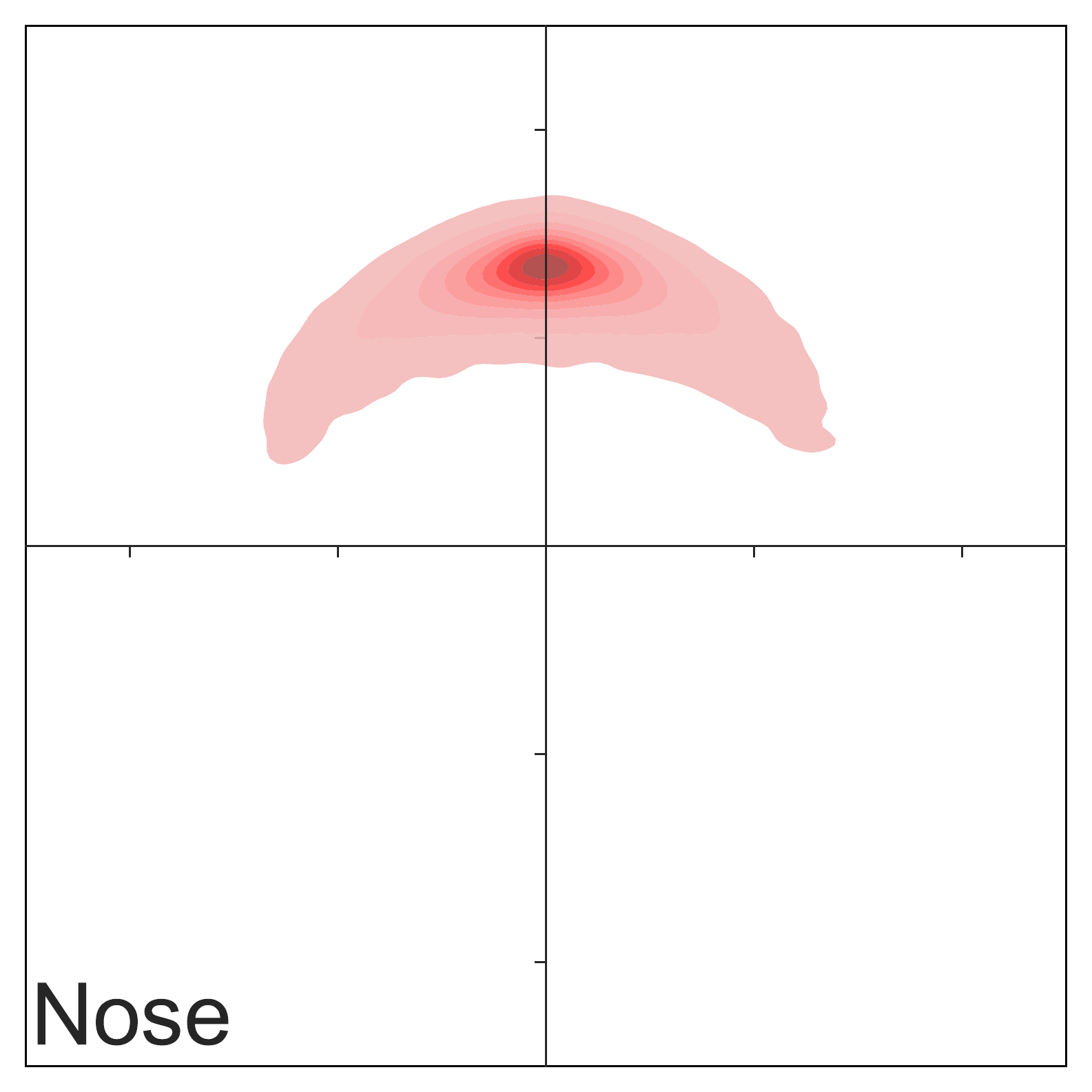}}
    \end{subfigure}
        \begin{subfigure}[t]{0.14\textwidth}
        {\includegraphics[height=2.5cm]{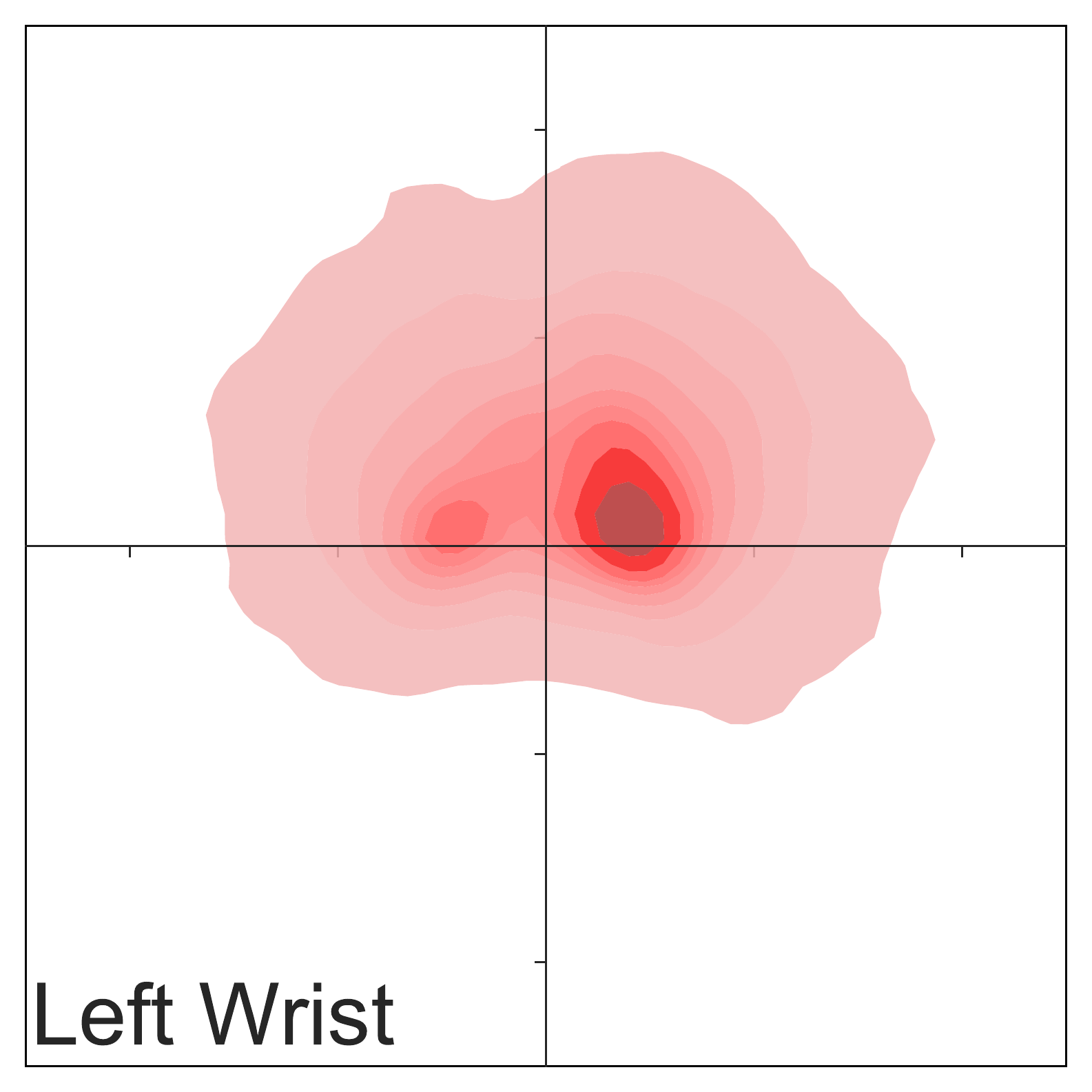}}
    \end{subfigure}
        \begin{subfigure}[t]{0.14\textwidth}
        {\includegraphics[height=2.5cm]{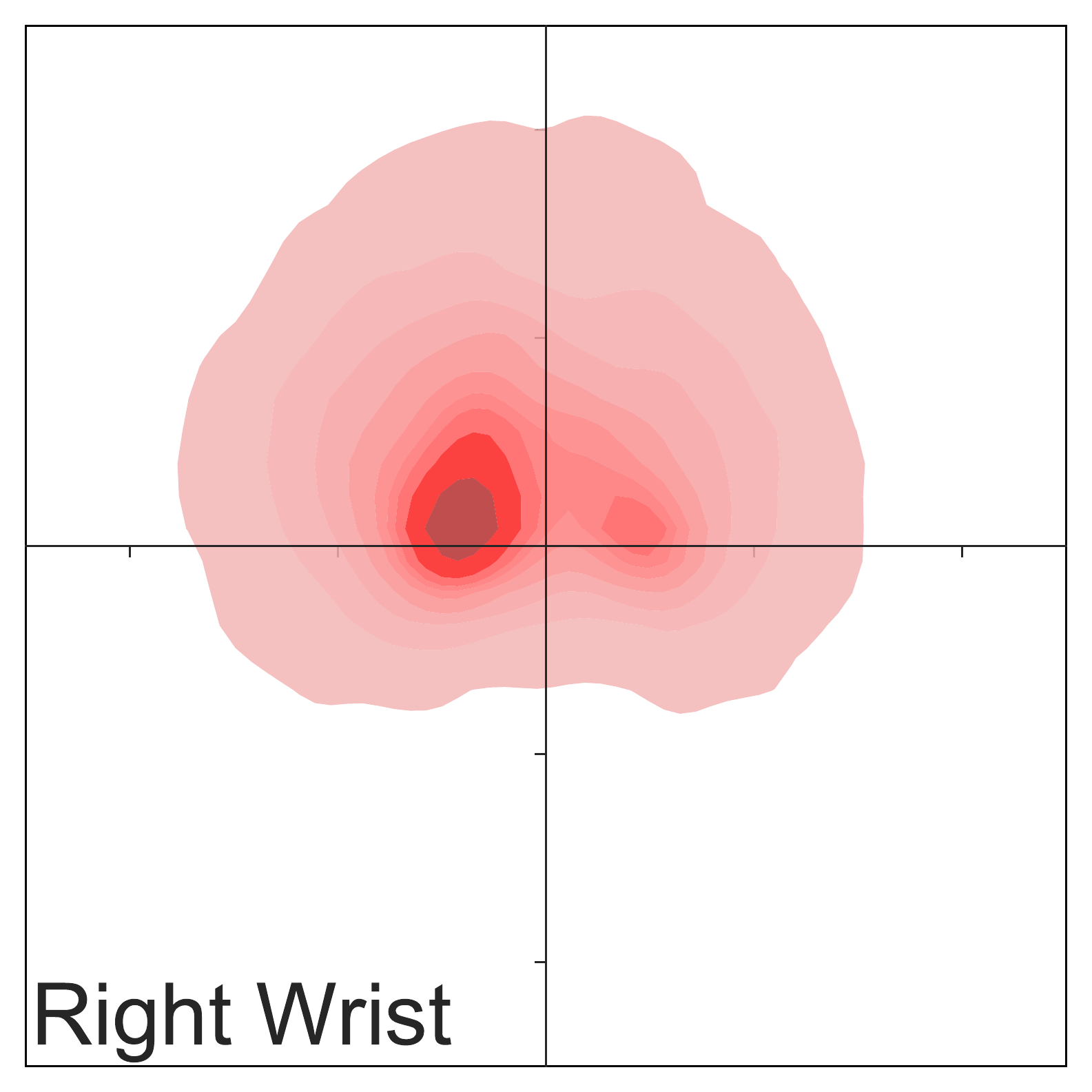}}
    \end{subfigure}
        \begin{subfigure}[t]{0.14\textwidth}
       {\includegraphics[height=2.5cm]{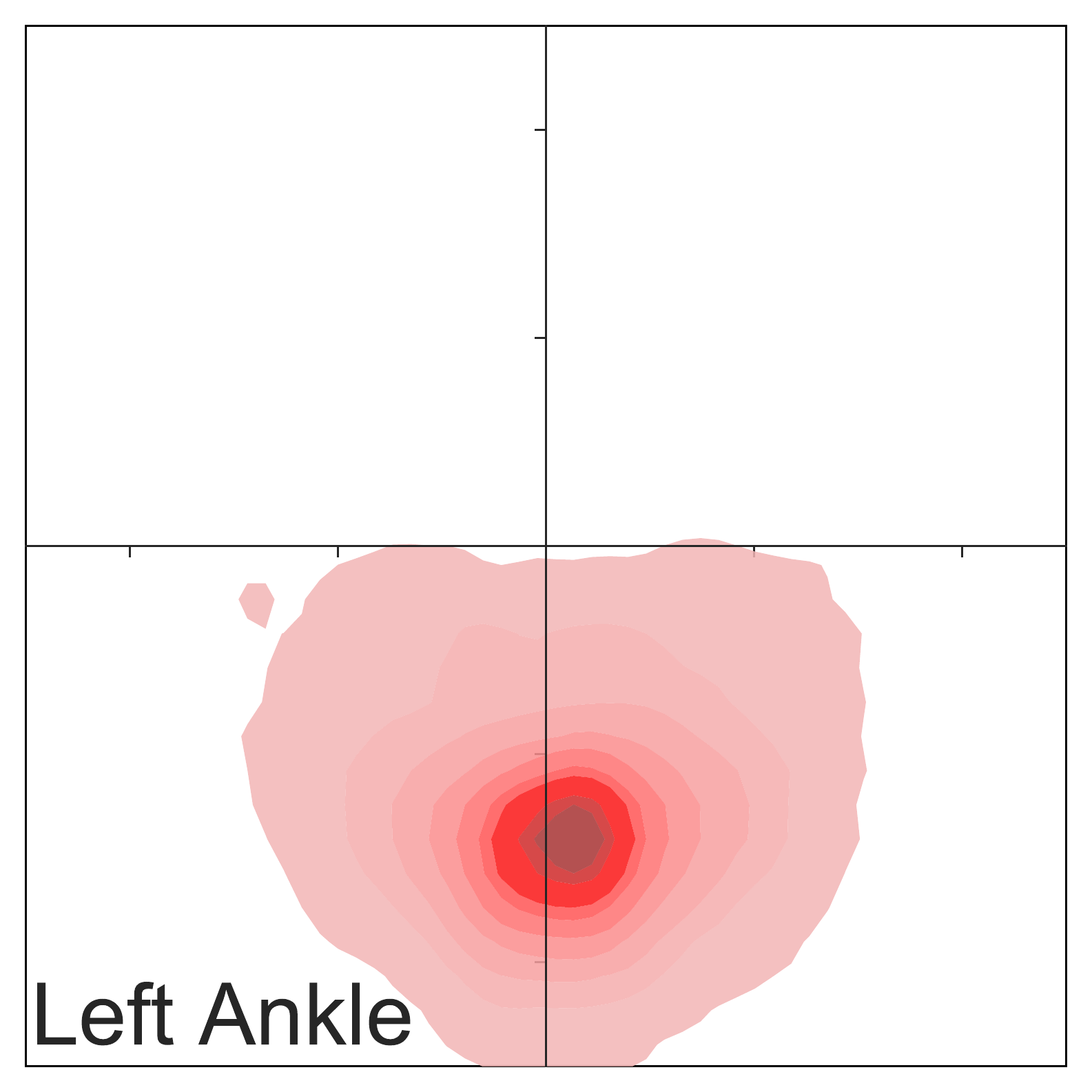}}
    \end{subfigure}
        \begin{subfigure}[t]{0.14\textwidth}
        {\includegraphics[height=2.5cm]{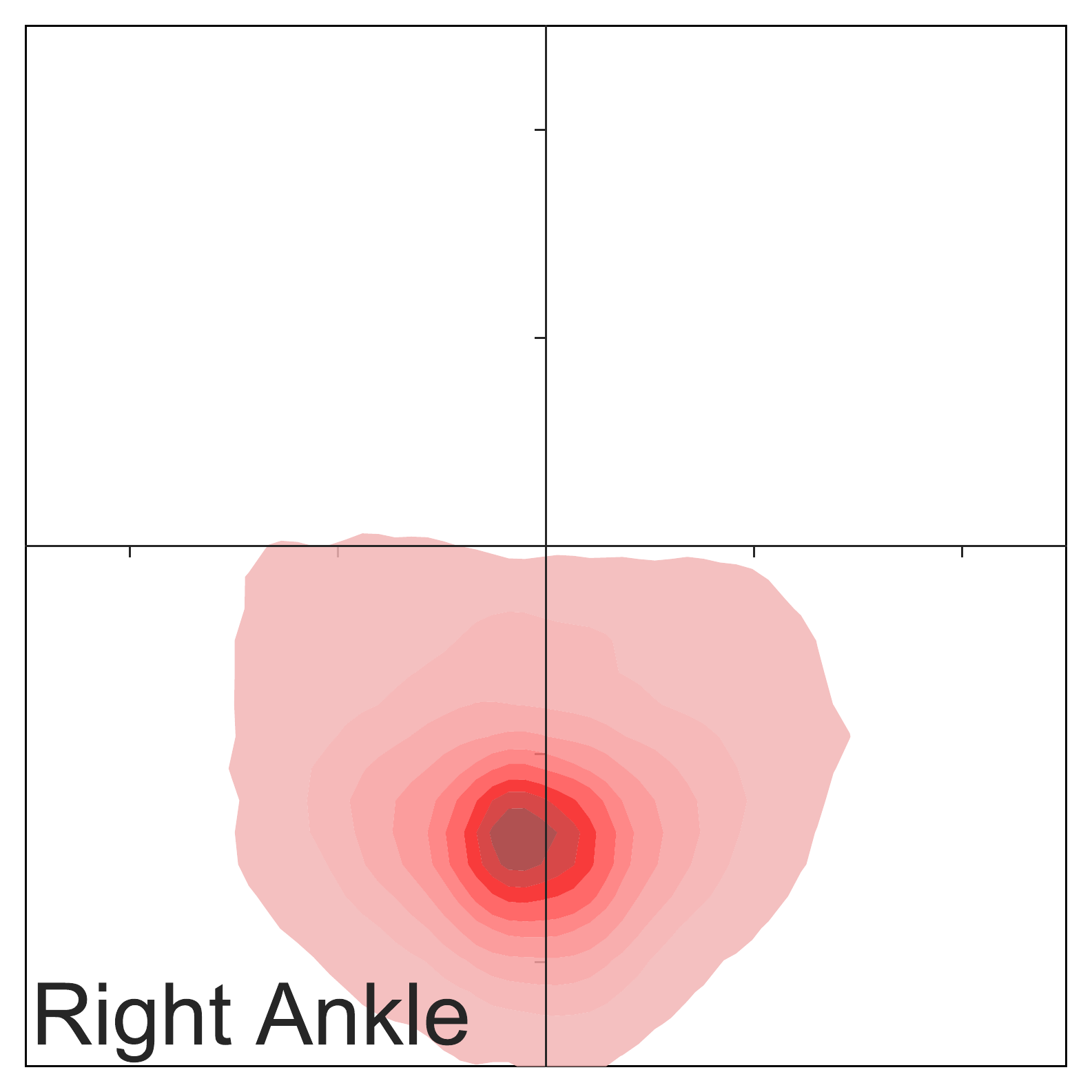}}
    \end{subfigure}
    \begin{subfigure}[t]{0.05\textwidth}
        {\includegraphics[height=2.5cm]{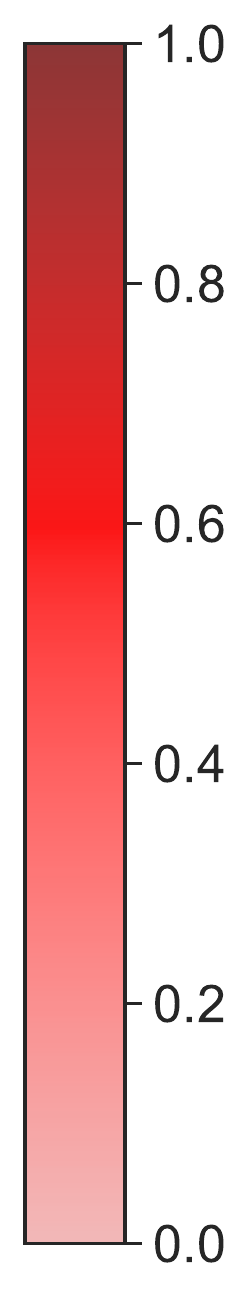}}
    \end{subfigure}
    \\
    \begin{subfigure}[t]{0.14\textwidth}
        {\includegraphics[height=2.5cm]{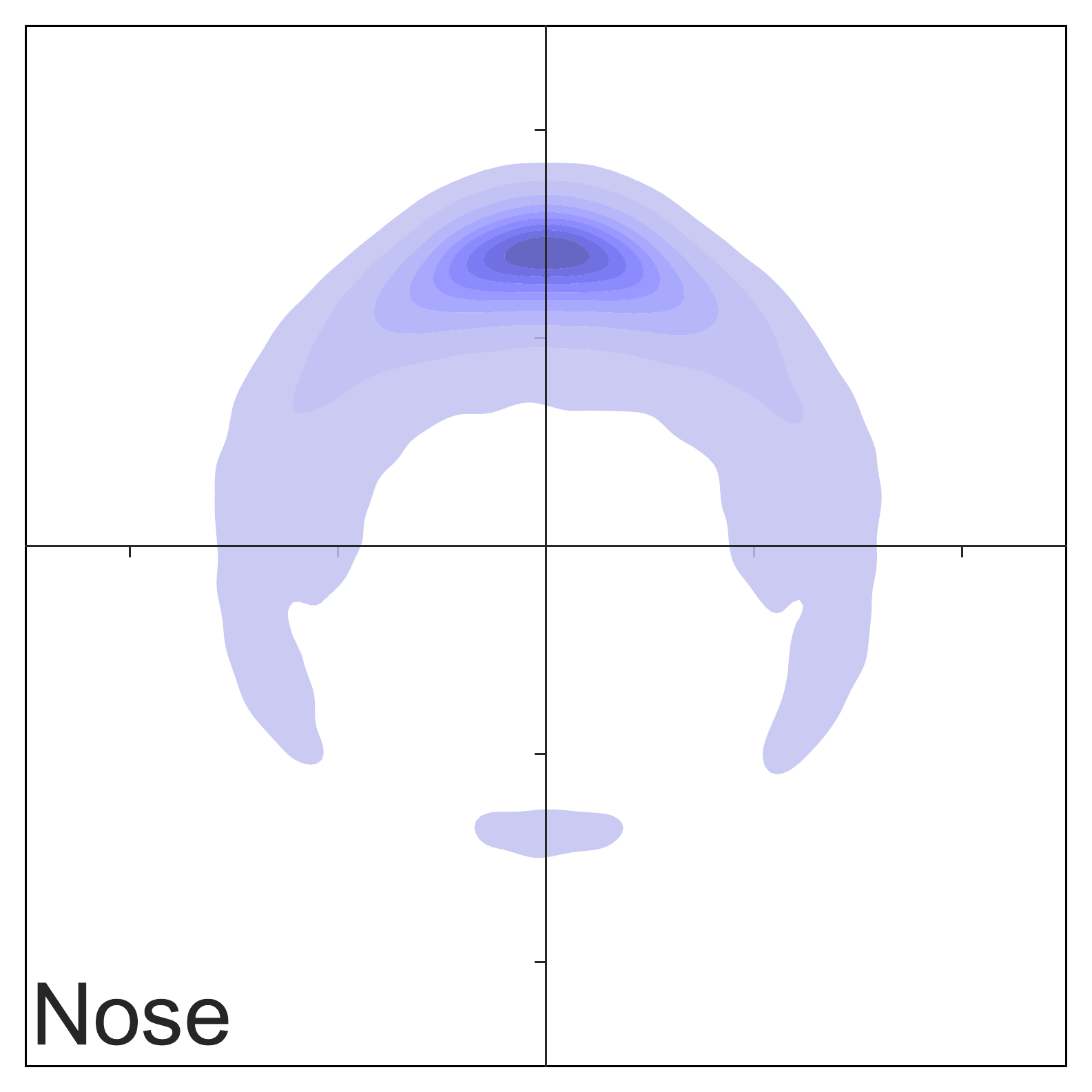}}
    \end{subfigure}
        \begin{subfigure}[t]{0.14\textwidth}
        {\includegraphics[height=2.5cm]{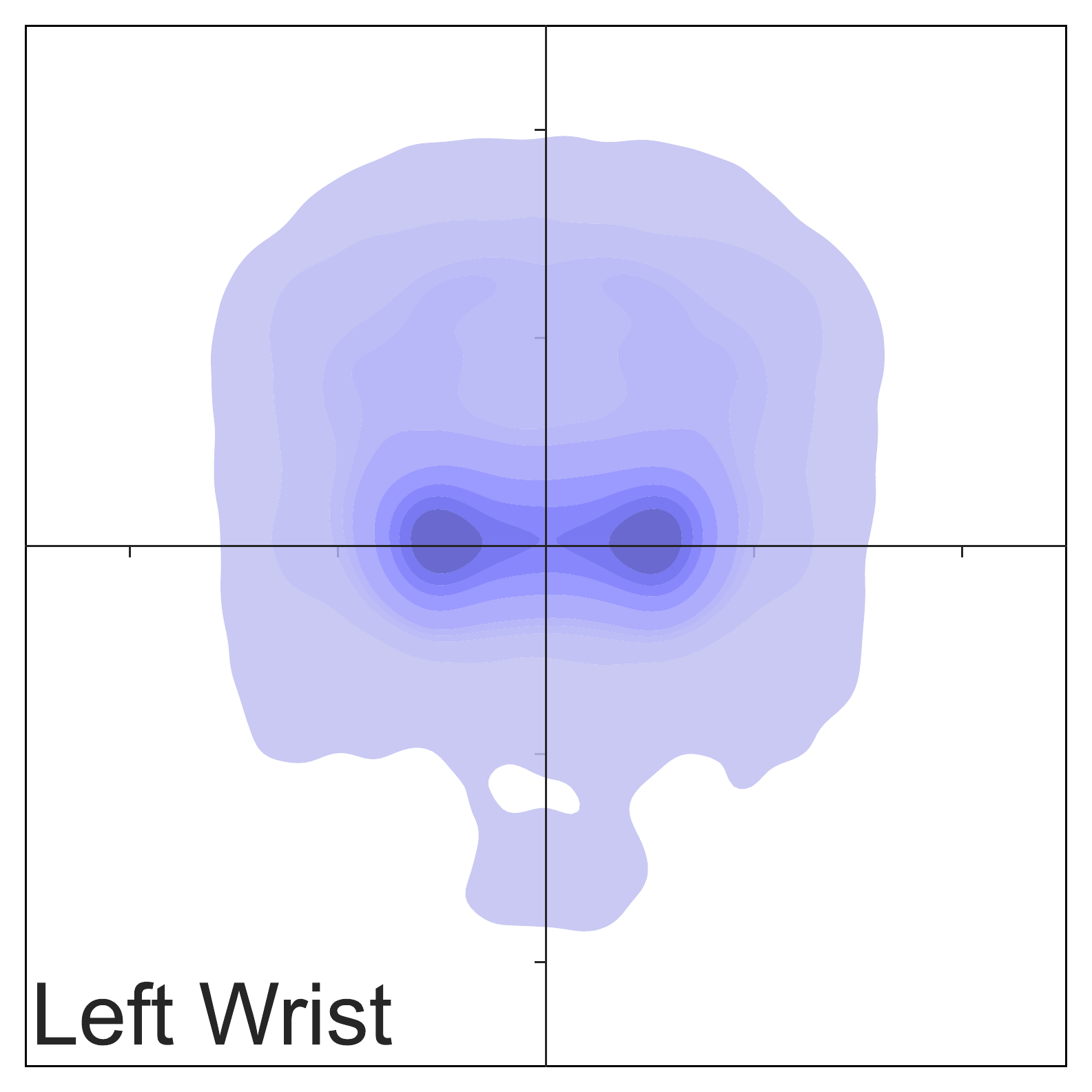}}
    \end{subfigure}
        \begin{subfigure}[t]{0.14\textwidth}
        {\includegraphics[height=2.5cm]{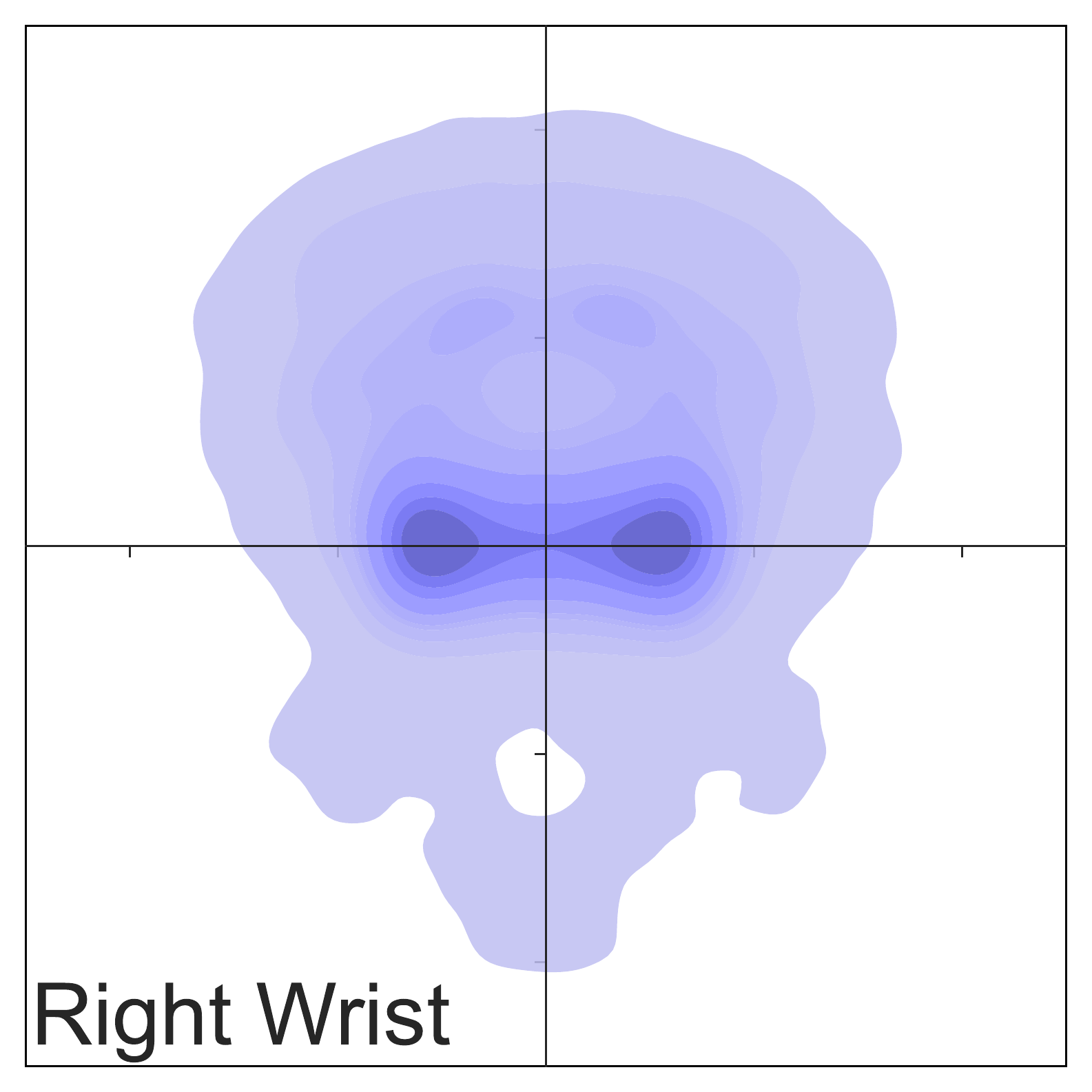}}
    \end{subfigure}
        \begin{subfigure}[t]{0.14\textwidth}
       {\includegraphics[height=2.5cm]{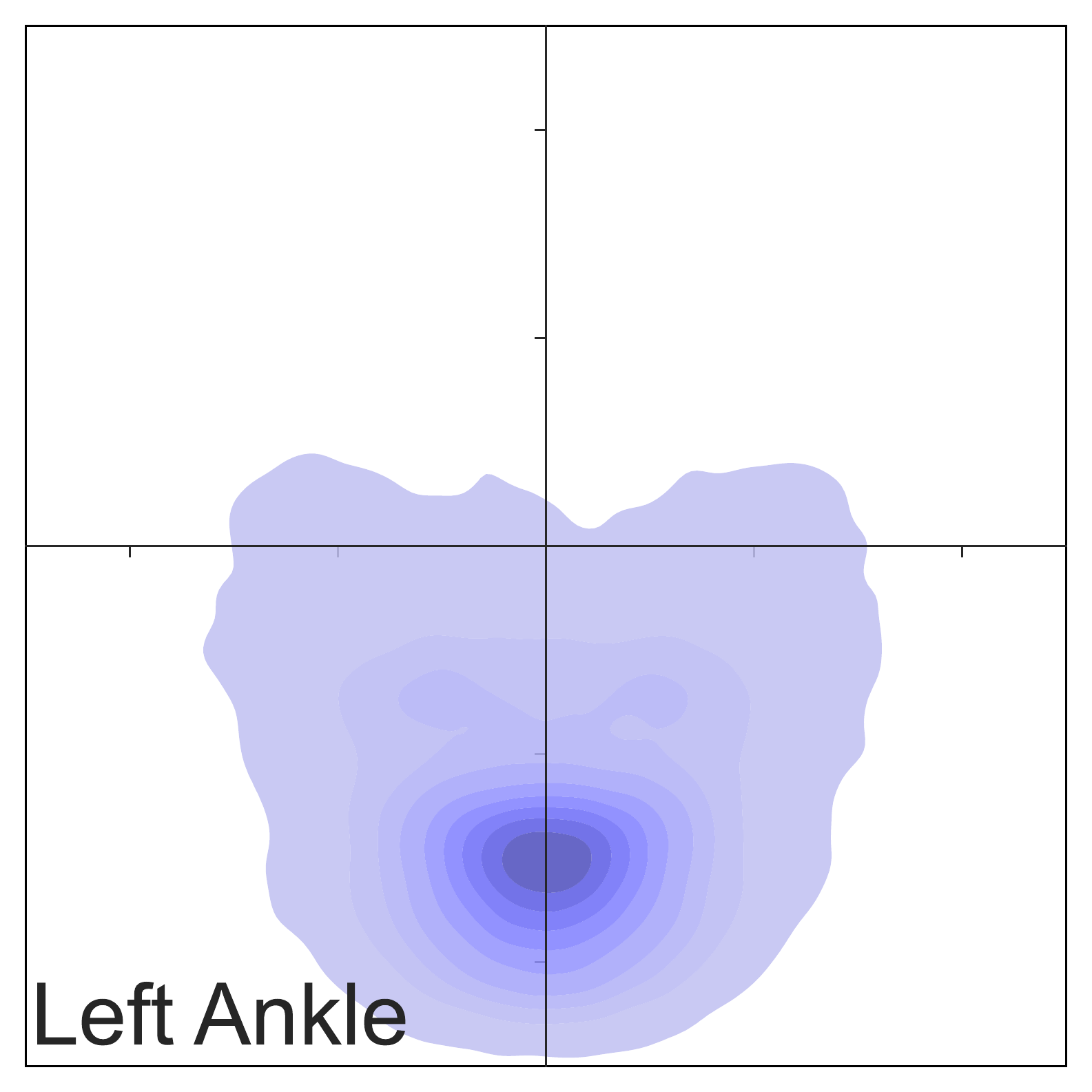}}
    \end{subfigure}
        \begin{subfigure}[t]{0.14\textwidth}
        {\includegraphics[height=2.5cm]{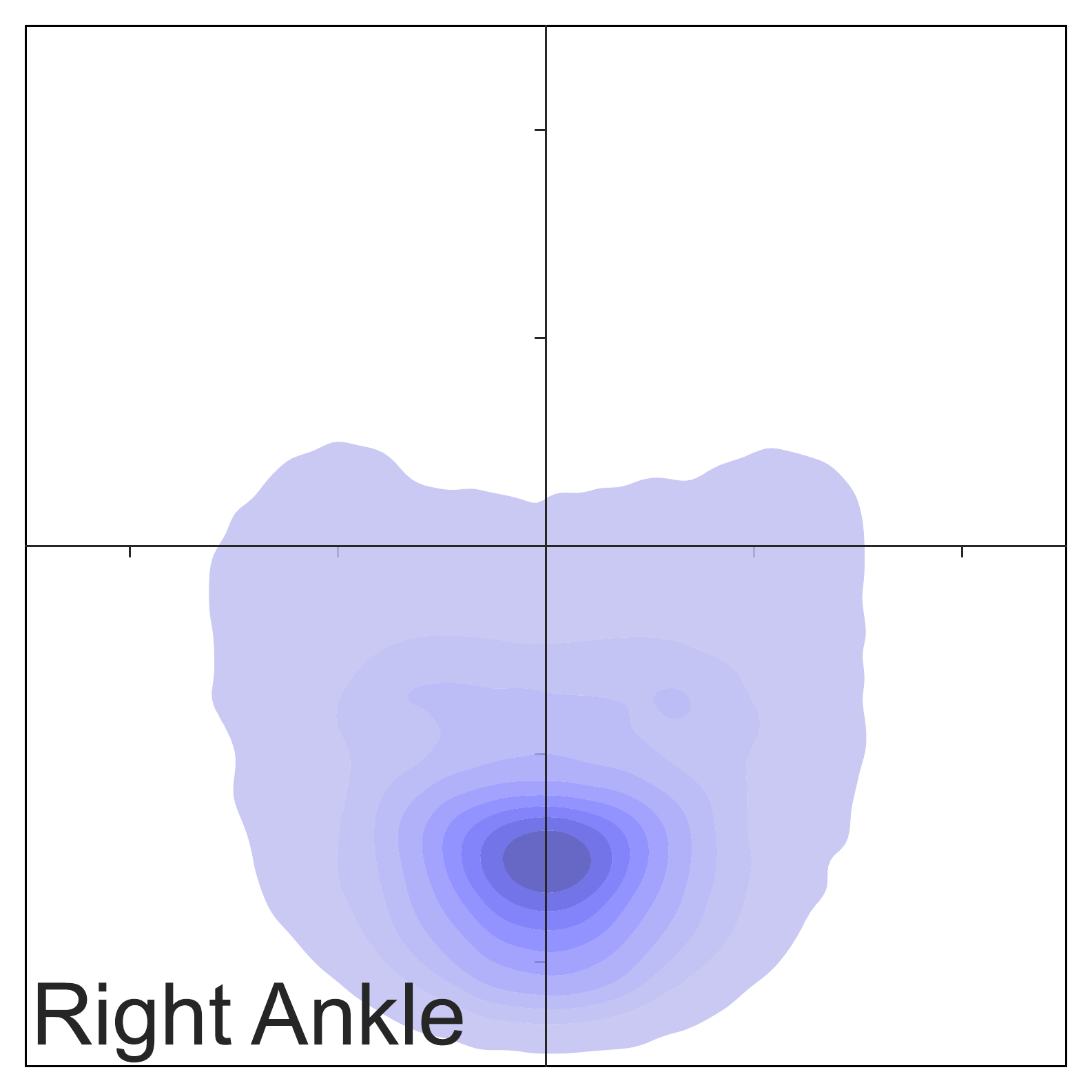}}
    \end{subfigure}
    \begin{subfigure}[t]{0.05\textwidth}
        {\includegraphics[height=2.5cm]{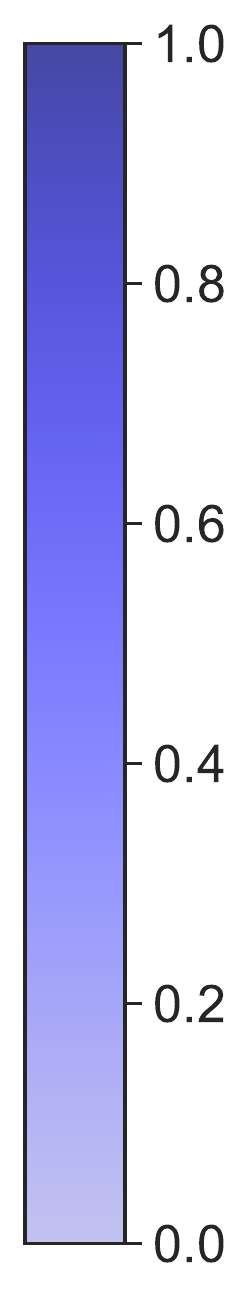}}
    \end{subfigure}
    \\
    \begin{subfigure}[t]{0.14\textwidth}
        {\includegraphics[height=2.5cm]{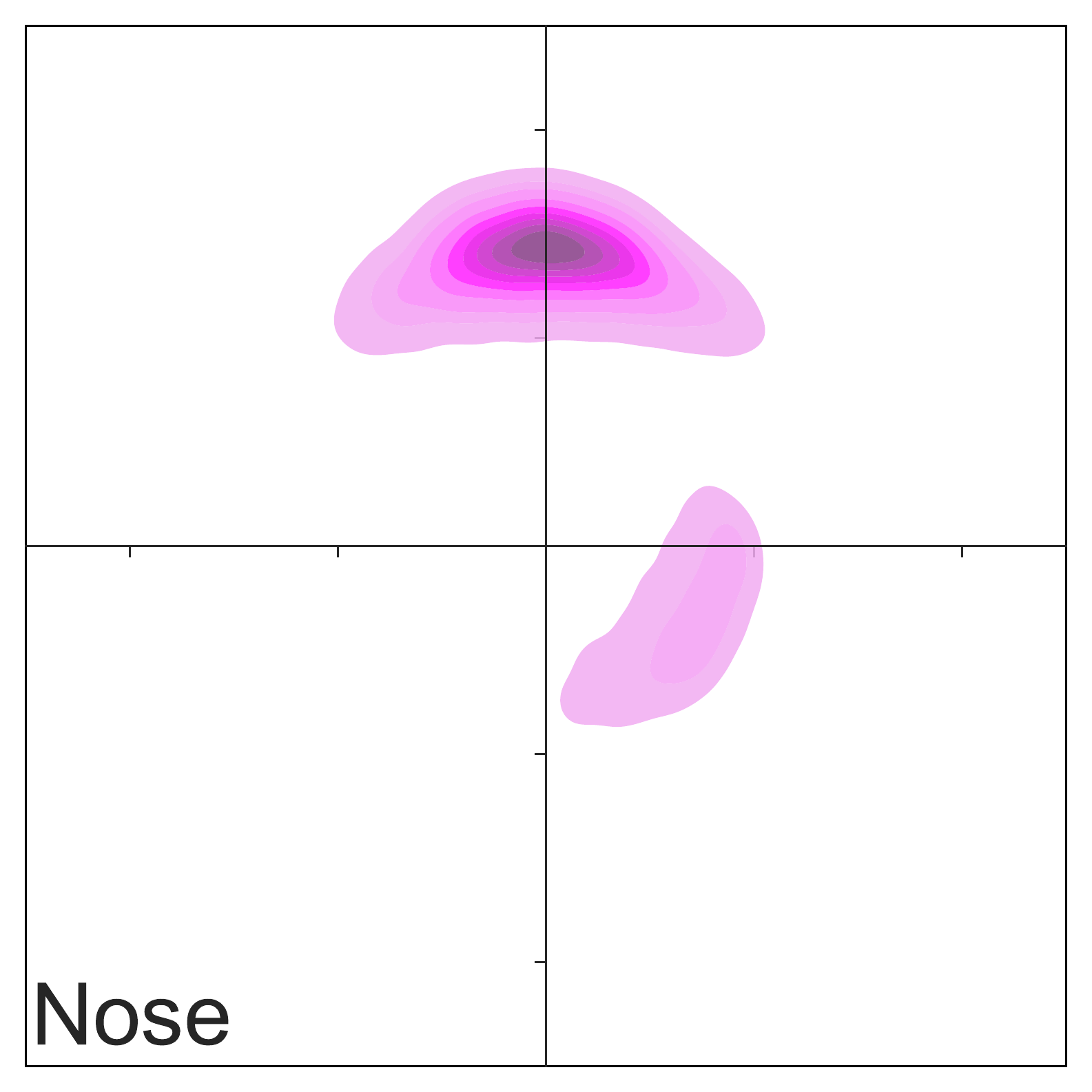}}
    \end{subfigure}
        \begin{subfigure}[t]{0.14\textwidth}
        {\includegraphics[height=2.5cm]{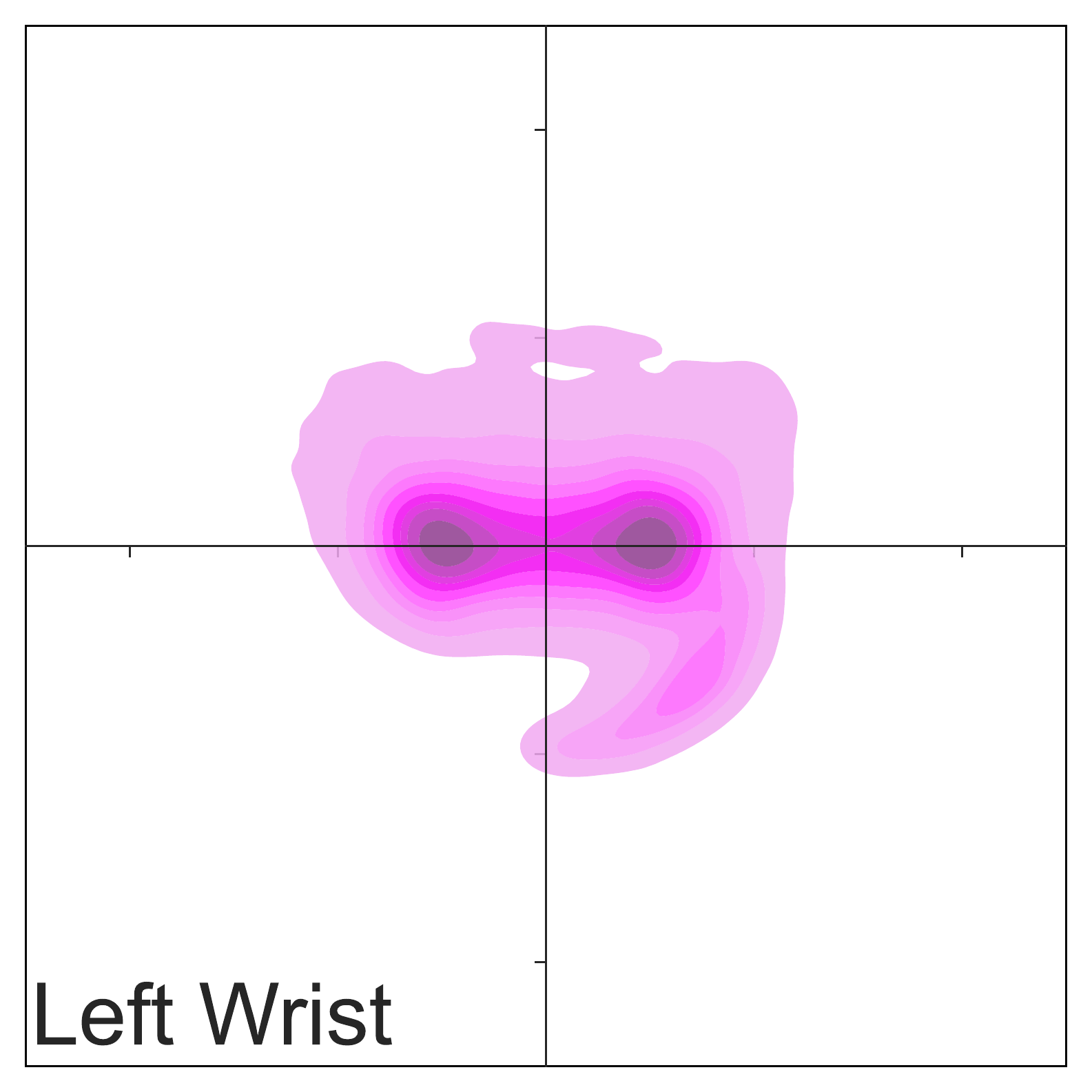}}
    \end{subfigure}
        \begin{subfigure}[t]{0.14\textwidth}
        {\includegraphics[height=2.5cm]{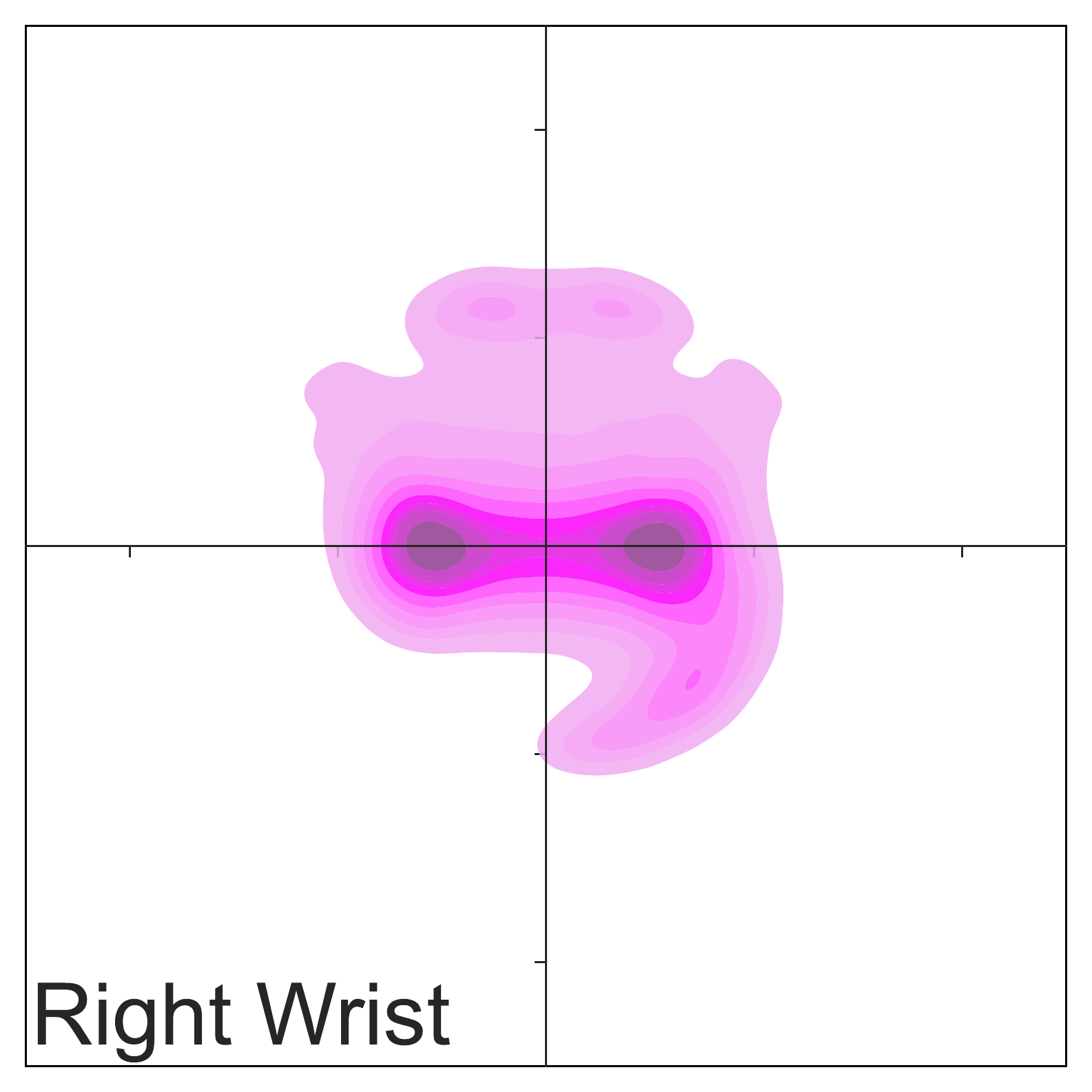}}
    \end{subfigure}
        \begin{subfigure}[t]{0.14\textwidth}
        {\includegraphics[height=2.5cm]{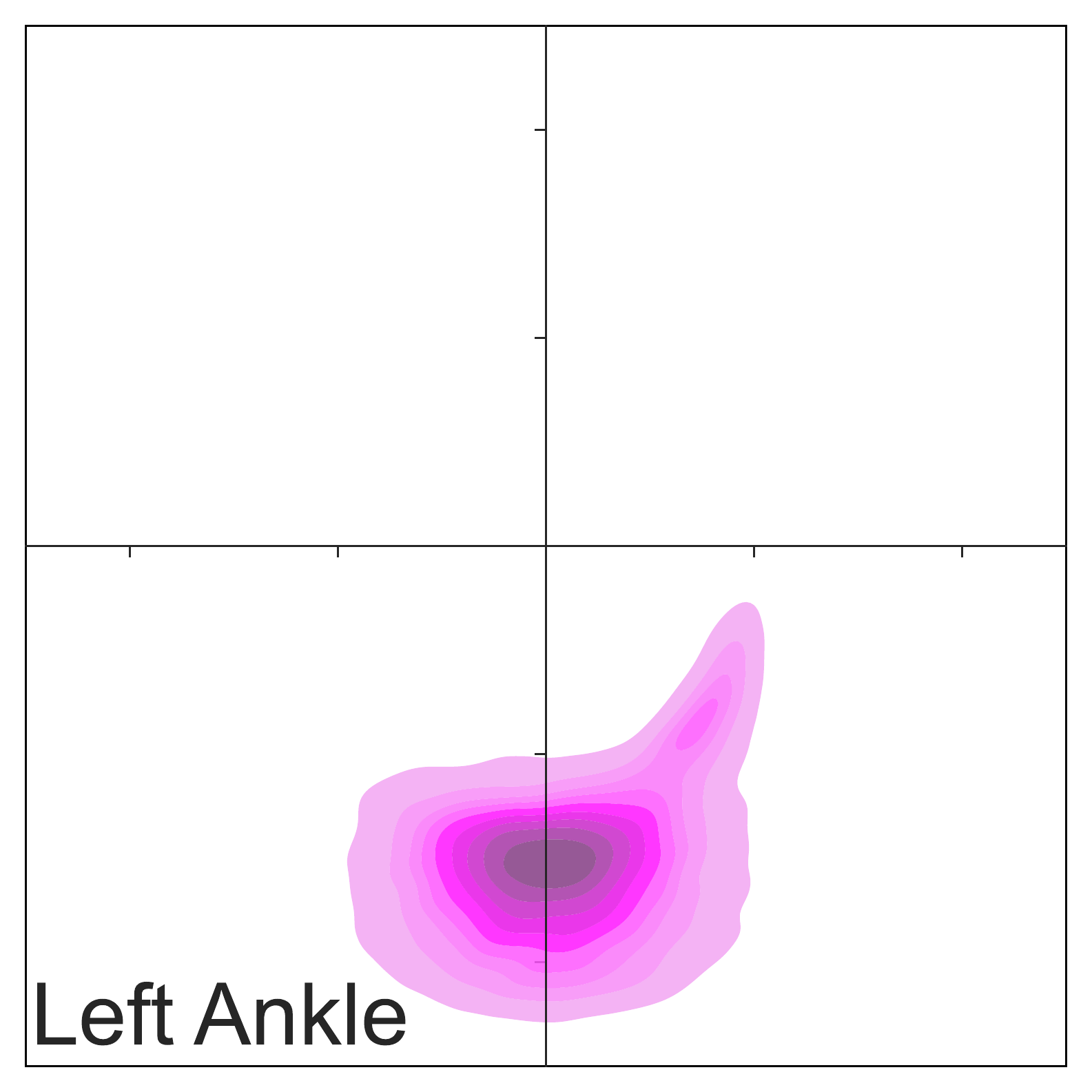}}
    \end{subfigure}
    \begin{subfigure}[t]{0.14\textwidth}
        {\includegraphics[height=2.5cm]{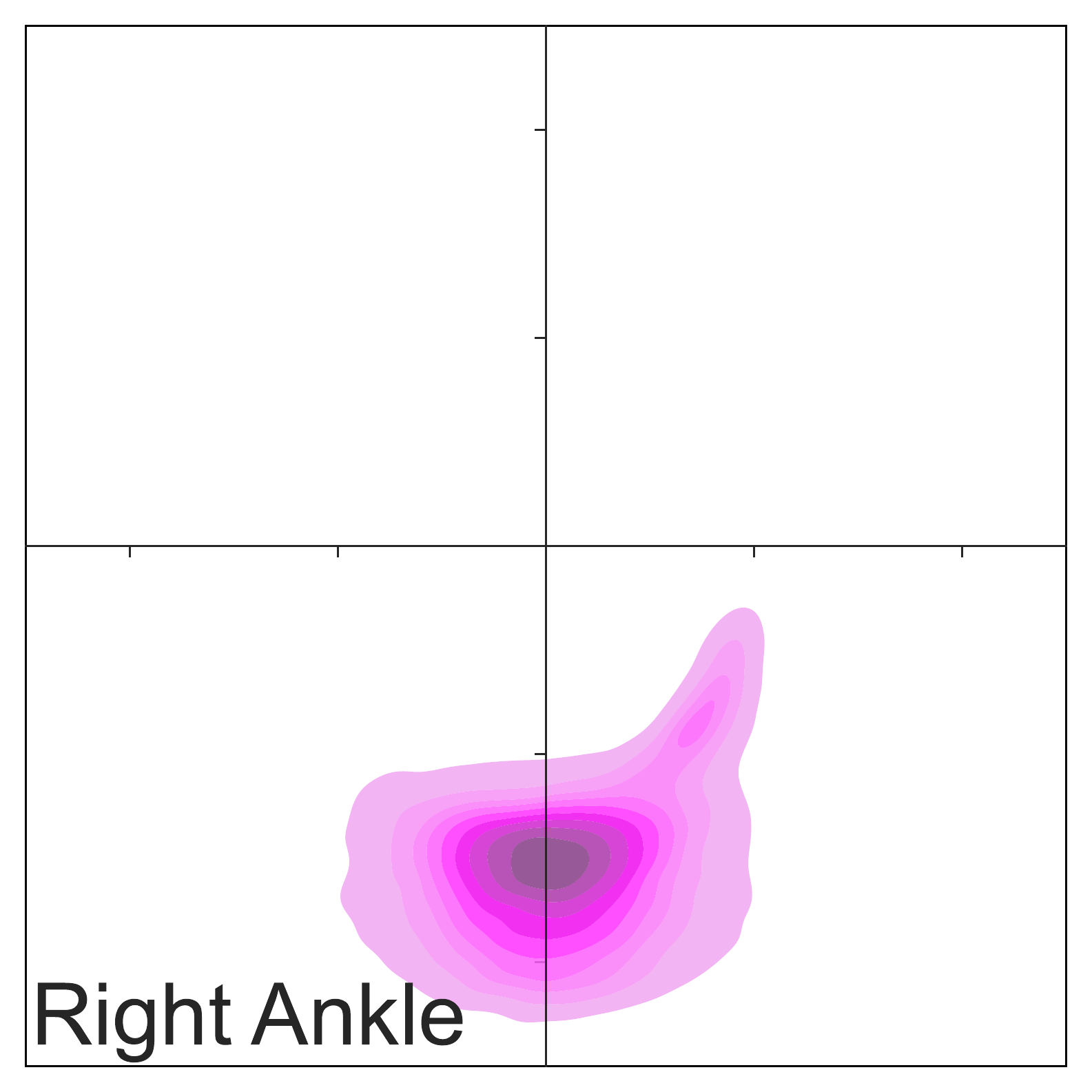}}
    \end{subfigure}
    \begin{subfigure}[t]{0.05\textwidth}
        {\includegraphics[height=2.5cm]{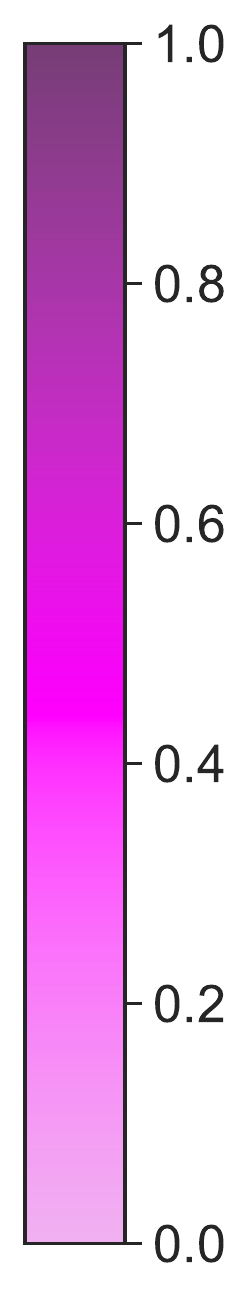}}
    \end{subfigure}
    \\
    \begin{subfigure}[t]{0.14\textwidth}
        {\includegraphics[height=2.5cm]{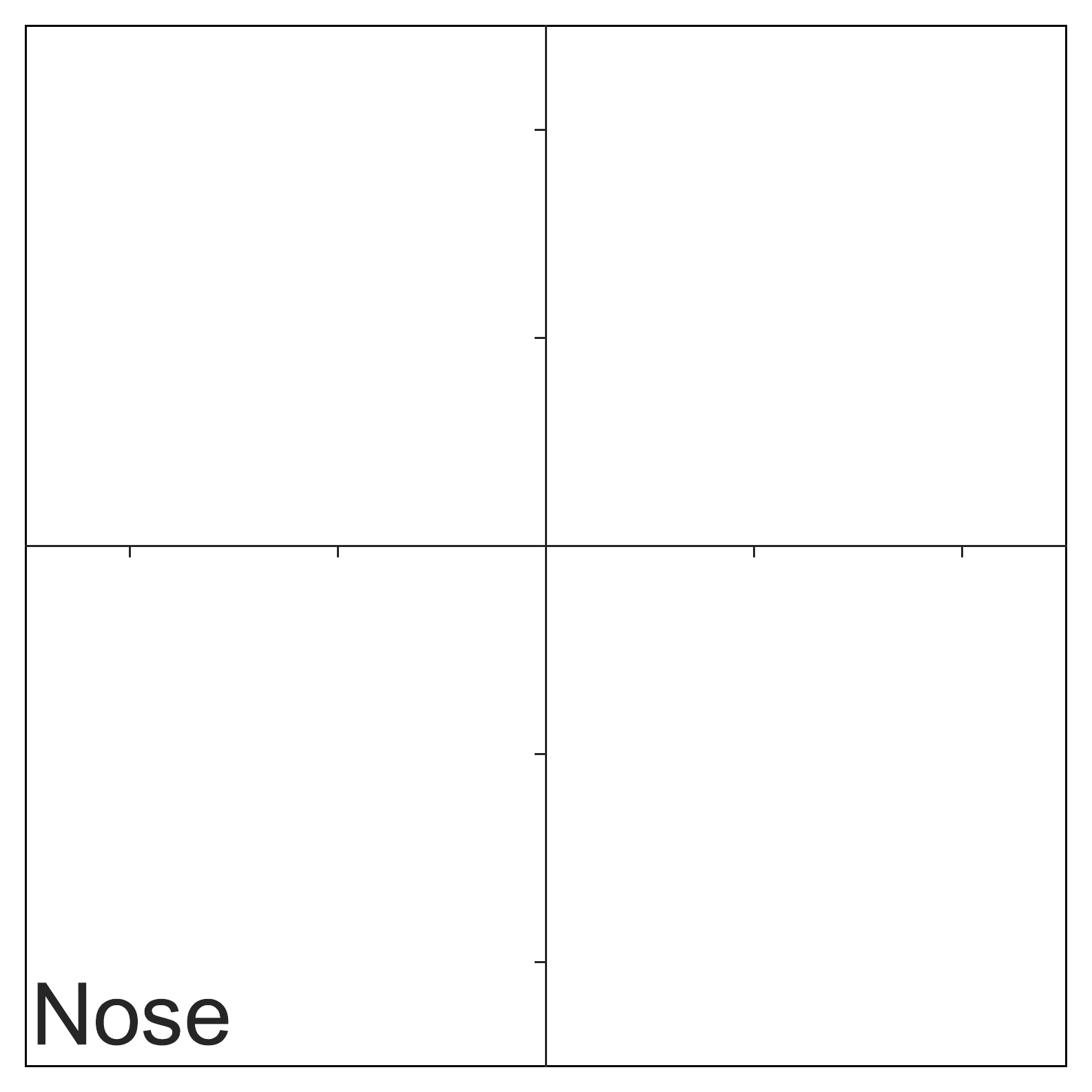}}
    \end{subfigure}
        \begin{subfigure}[t]{0.14\textwidth}
        {\includegraphics[height=2.5cm]{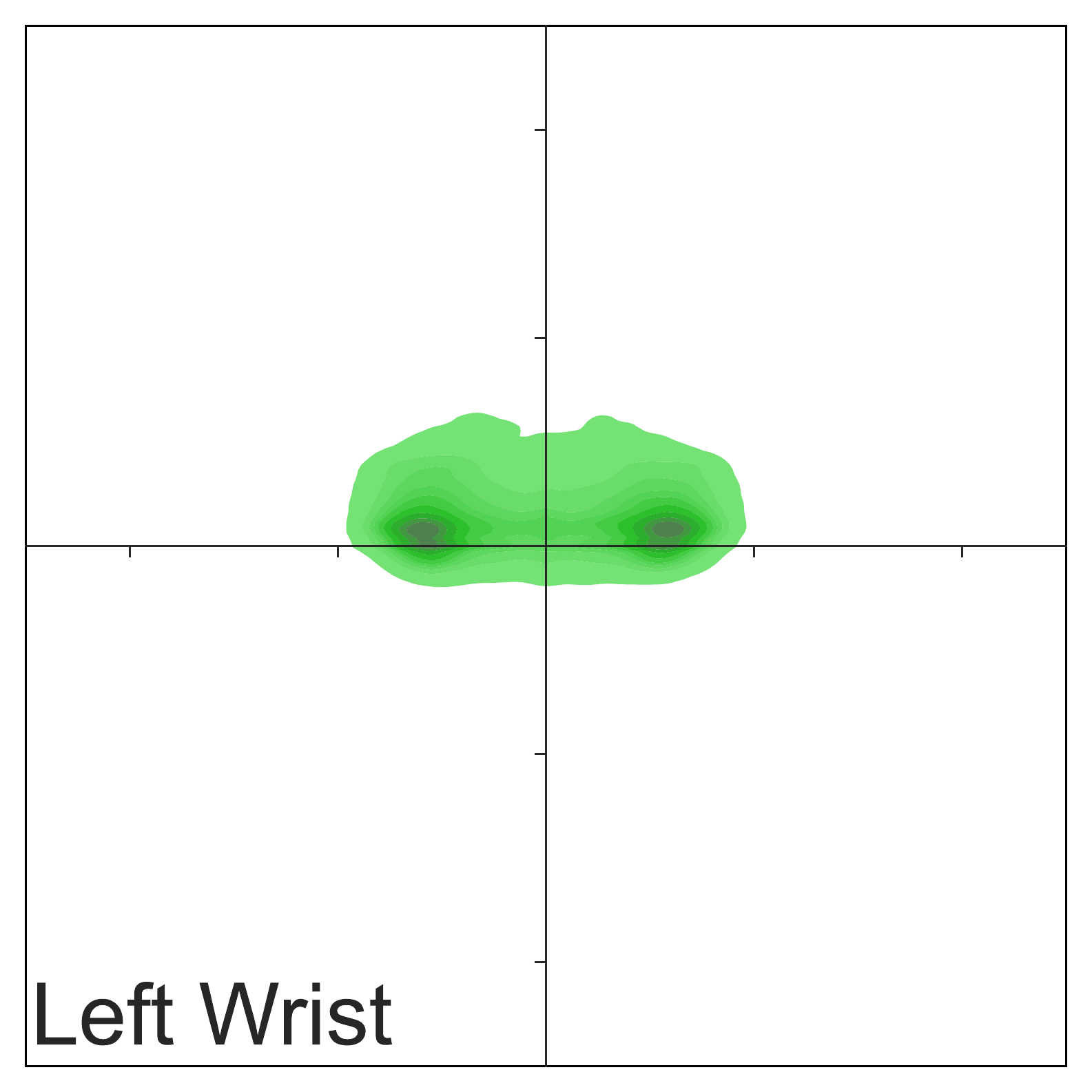}}
    \end{subfigure}
        \begin{subfigure}[t]{0.14\textwidth}
        {\includegraphics[height=2.5cm]{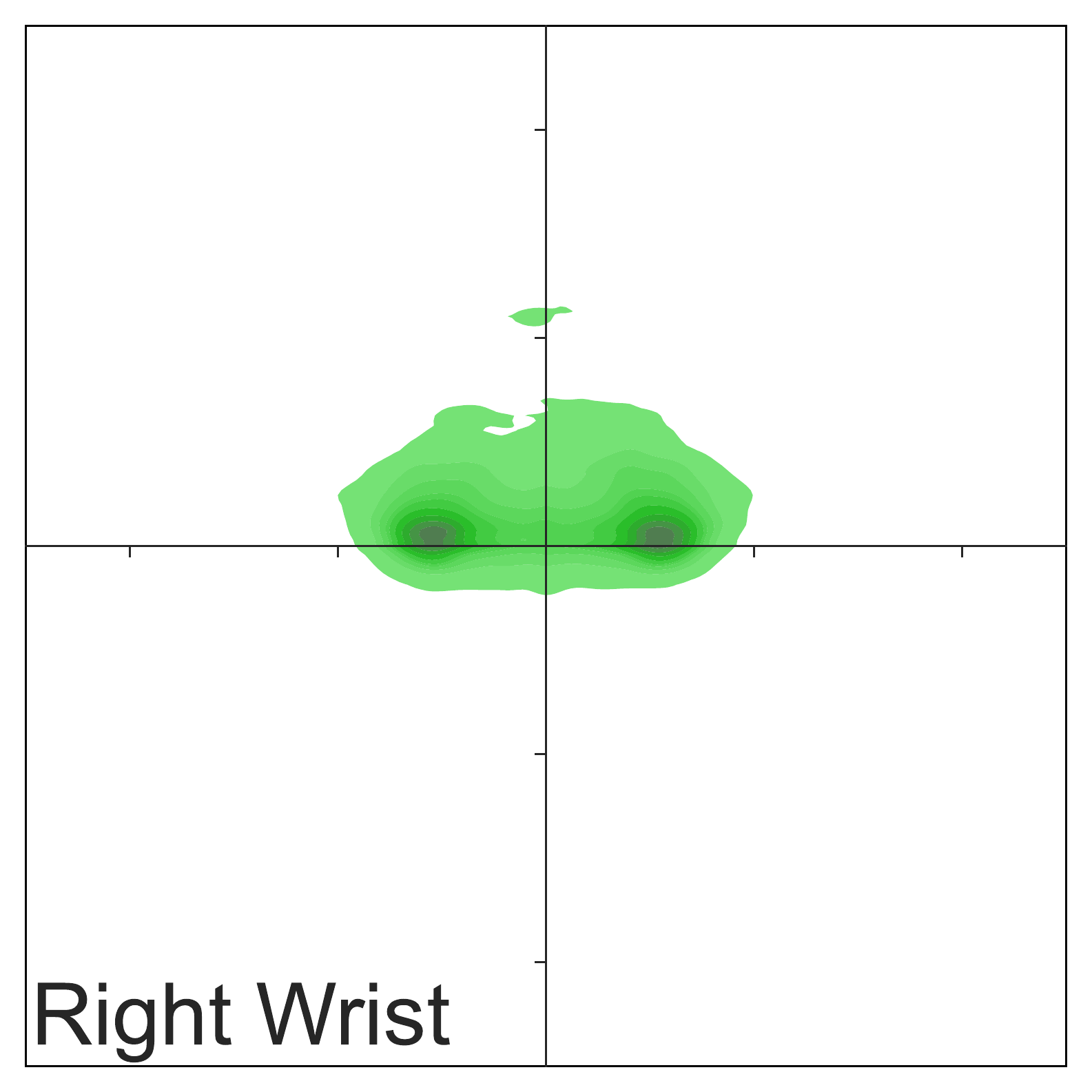}}
    \end{subfigure}
        \begin{subfigure}[t]{0.14\textwidth}
        {\includegraphics[height=2.5cm]{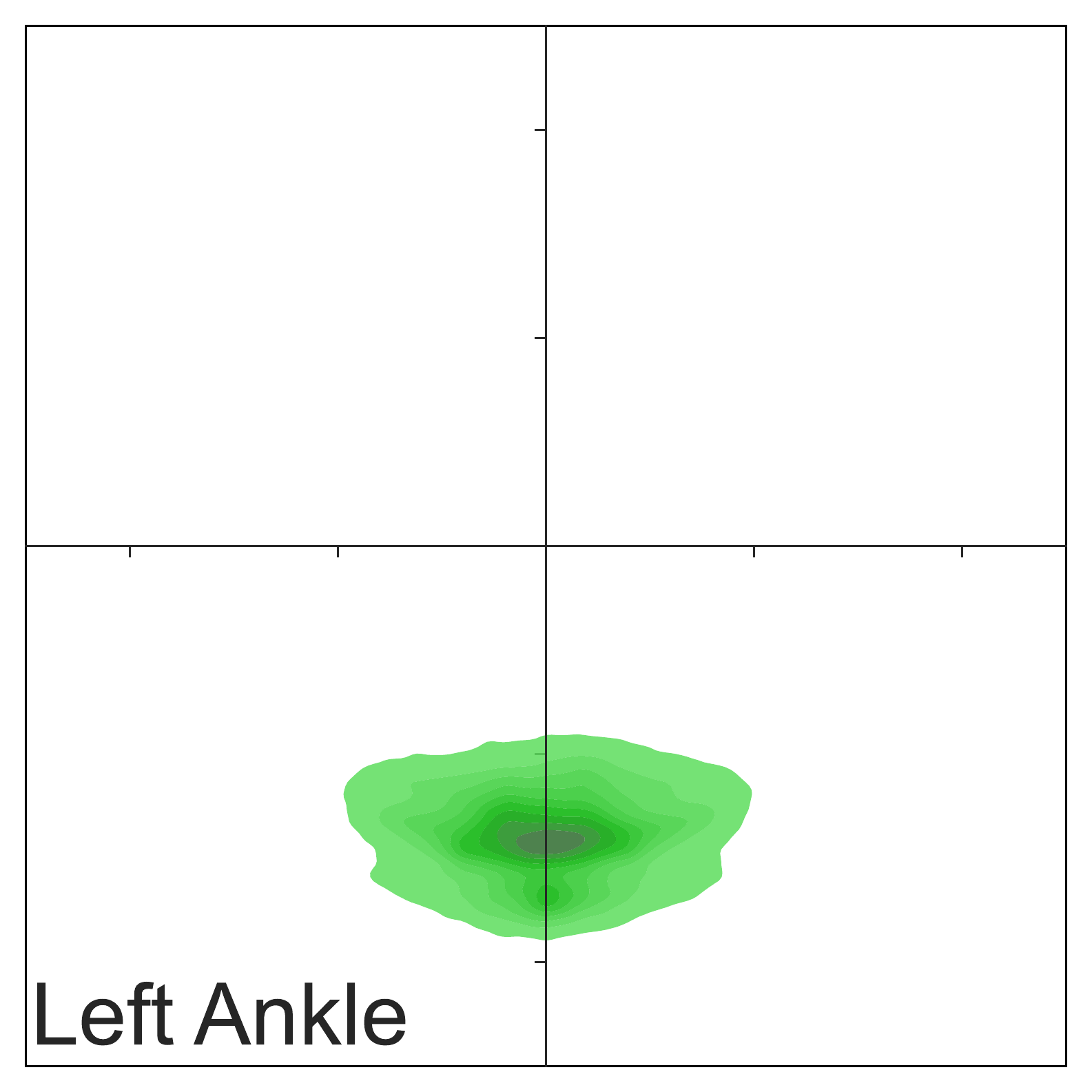}}
    \end{subfigure}
    \begin{subfigure}[t]{0.14\textwidth}
        {\includegraphics[height=2.5cm]{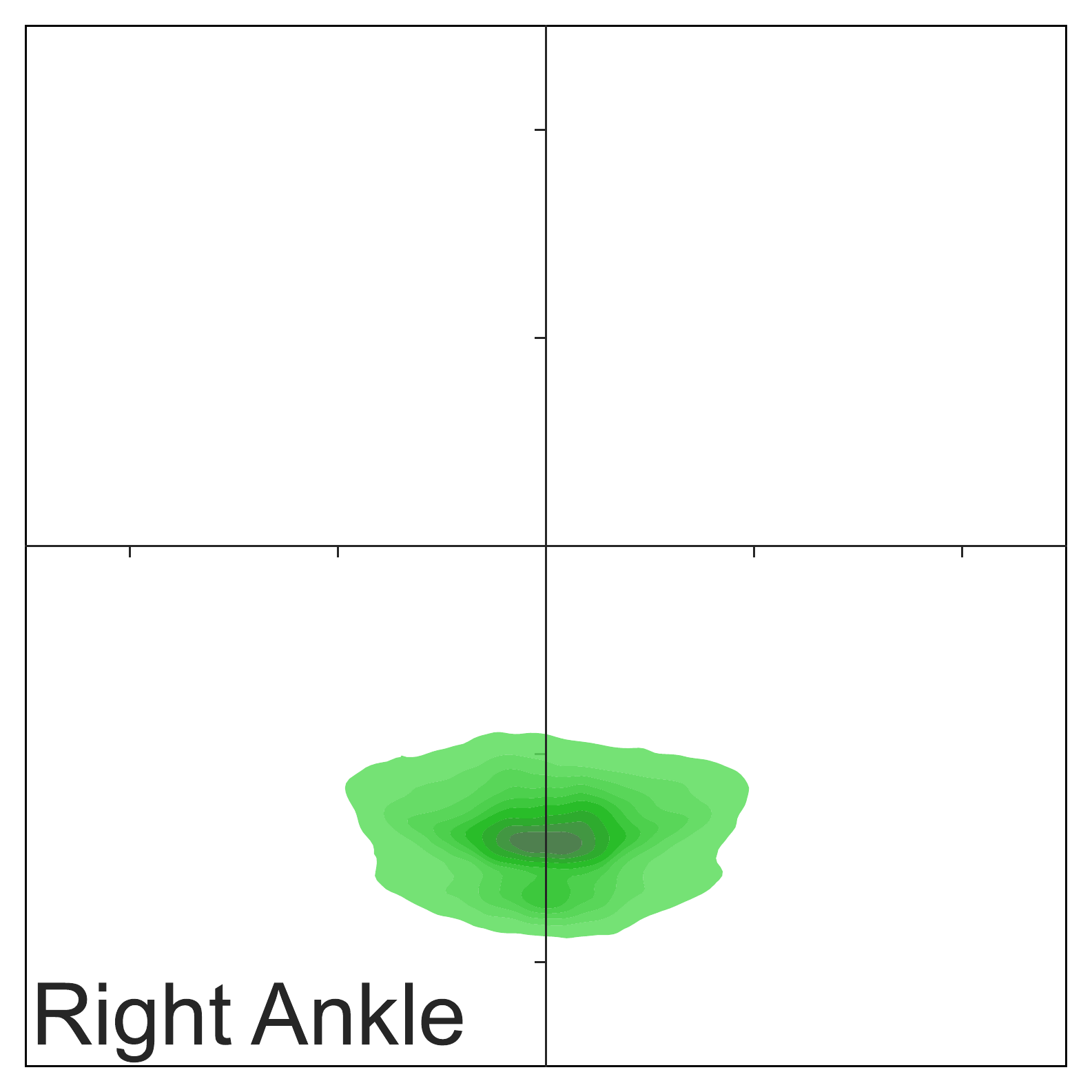}}
    \end{subfigure}
    \begin{subfigure}[t]{0.05\textwidth}
        {\includegraphics[height=2.5cm]{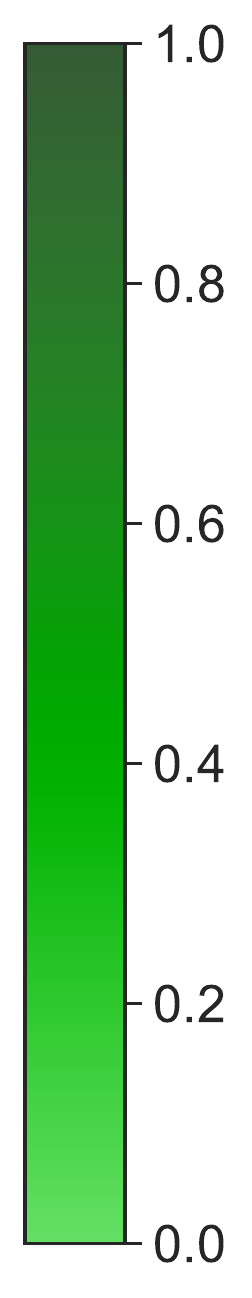}}
    \end{subfigure}
    \caption{\textbf{Pose Diversity Location Heatmaps for Five Representative Keypoints}. From top row to bottom: COCO-person, PSP-HDRI, PSP-HDRI$+$, MOTSynth. We aligned all keypoints according to~\citep{ebadi2021peoplesanspeople} to produce normalized keypoint locations. We use the animation randomization to control the generated human pose diversity.
    }
    \label{fig:posestatselect}
\end{figure}

In our data generator we used a set of animations derived from human motion capture clips to create a reasonably diverse set of poses for our human models. In order to quantify the pose diversity in our generated dataset, we used a technique from~\citep{ebadi2021peoplesanspeople} where keypoint annotations from all the annotated person instances are used, provided that the torso of the character has annotations (hips and shoulders). Then all the keypoints are aligned such that the mid-hip point is at $(0, 0)$ coordinates on a 2D axis. The keypoint distances are scaled according to the length of each torso, in order to make all the skeletons roughly the same size. If we then plot each keypoint individually, we obtain the heatmaps shown in fig.~\ref{fig:posestatselect}. We opted to show only the representative keypoints that belong to the extremities of the human, as those will have the largest dispalcement. Note that PSP-HDRI (in blue) shows a more symmetrical pattern with larger footprint compared with COCO (in red). Most people in COCO are captured from the frontal view, hence the asymmetrical heatmaps. For PSP-HDRI$+$ (in purple) since we only used simple animations, we observe a smaller footprint for each keypoint location variation. The MOTSynth dataset (in green) does not have facial keypoints, hence the nose keypoint has no information. For the rest of the keypoints, we observe a larger footprint for the PSP-HDRI$+$ compared with MOTSynth; meaning that our animations are still more diverse than those of MOTSynth with people walking around in scenes of GTA V game. 

